\documentclass[twoside,11pt]{article}

\usepackage{jmlr2e}
\usepackage{xcolor}
\usepackage{footnote}
\usepackage[utf8]{inputenc} 
\usepackage[T1]{fontenc}    
\usepackage{hyperref}       
\usepackage{url}            
\usepackage{booktabs}       
\usepackage{amsfonts}       
\usepackage{nicefrac}       
\usepackage{microtype}      
\usepackage{algorithm}
\usepackage{algorithmic}
\usepackage{amsmath}
\usepackage{times}
\usepackage{mathtools}
\usepackage{color}
\usepackage{siunitx}
\usepackage{multirow}
\usepackage{mathrsfs}
\usepackage{caption}
\usepackage{xspace}
\usepackage[export]{adjustbox}
\usepackage{float}
\usepackage{enumitem}
\usepackage{listings}
\usepackage{multirow}
\usepackage{bbm}
\usepackage{wrapfig}
\usepackage{subcaption}
\usepackage{wrapfig}

\usepackage{amsmath,amsfonts,bm}

\def\eqref#1{equation~\ref{#1}}

\def\1{\bm{1}}

\def\rv{{\textnormal{v}}}

\def\rz{{\textnormal{z}}}

\def\rvm{{\mathbf{m}}}

\def\rvt{{\mathbf{t}}}

\def\rvx{{\mathbf{x}}}

\def\rvz{{\mathbf{z}}}

\def\vu{{\bm{u}}}
\def\vv{{\bm{v}}}

\def\mA{{\bm{A}}}

\def\mI{{\bm{I}}}

\DeclareMathAlphabet{\mathsfit}{\encodingdefault}{\sfdefault}{m}{sl}
\SetMathAlphabet{\mathsfit}{bold}{\encodingdefault}{\sfdefault}{bx}{n}

\newcommand{\E}{\mathbb{E}}

\newcommand{\R}{\mathbb{R}}

\newcommand{\KL}{D_{\mathrm{KL}}}

\DeclareMathOperator*{\argmax}{arg\,max}

\hypersetup{
    colorlinks,
    anchorcolor={blue!50!black},
    linkcolor={blue!50!black},
    citecolor={blue!50!black},
    urlcolor={blue}
}

\makesavenoteenv{tabular}

\renewcommand{\P}{P}
\newcommand{\Q}{Q}
\newcommand{\ie}{\emph{i.e.}}
\newcommand{\supp}{\operatorname{supp}}
\DeclareRobustCommand{\myhammer}{%
  \begingroup\normalfont
  \includegraphics[height=1.1\fontcharht\font`\B]{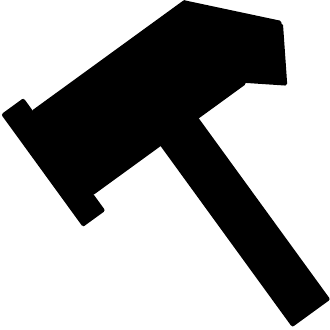}%
  \endgroup
}

\usepackage{lastpage}
\jmlrheading{21}{2020}{1-\pageref{LastPage}}{11/19; Revised
8/20}{9/20}{19-976}{Francesco Locatello, Stefan Bauer, Mario Lucic, Gunnar R\"atsch, Sylvain Gelly, Bernhard Sch\"olkopf and Olivier Bachem}
\ShortHeadings{Unsupervised Learning of Disentangled Representations and their Evaluation}{F. Locatello, S. Bauer, M. Lucic, G. R\"atsch, S. Gelly, B. Sch\"olkopf, and O. Bachem}

\begin{document}

%
%

\title{A Sober Look at the Unsupervised Learning of Disentangled Representations and their Evaluation}

\author{\name Francesco Locatello \email francesco.locatello@inf.ethz.ch \\
       \addr Department of Computer Science\\
       ETH Zurich\\
       Universit\"atstrasse 6, 8092 Z\"urich, Switzerland
       \AND
       \name Stefan Bauer \email stefan.bauer@tuebingen.mpg.de \\
       \addr Department of Empirical Inference\\
       Max Planck Institute for Intelligent Systems\\
       Max-Planck-Ring 4, 72076 T\"ubingen, Germany
       \AND
       \name Mario Lucic \email lucic@google.com \\
       \addr Google Research, Brain Team\\
       Brandschenkestrasse 110, 8002 Z\"urich, Switzerland
       \AND
       \name Gunnar R\"atsch \email ratsch@inf.ethz.ch \\
       \addr Department of Computer Science\\
       ETH Zurich\\
       Universit\"atstrasse 6, 8092 Z\"urich, Switzerland 
       \AND
       \name Sylvain Gelly \email sylvaingelly@google.com \\
       \addr Google Research, Brain Team\\
       Brandschenkestrasse 110, 8002 Z\"urich, Switzerland
       \AND
       \name Bernhard Sch\"olkopf \email bs@tuebingen.mpg.de \\
       \addr Department of Empirical Inference\\
       Max Planck Institute for Intelligent Systems\\
       Max-Planck-Ring 4, 72076 T\"ubingen, Germany
       \AND
       \name Olivier Bachem \email bachem@google.com \\
       \addr Google Research, Brain Team\\
       Brandschenkestrasse 110, 8002 Z\"urich, Switzerland
       }

\editor{Kilian Weinberger}

\maketitle

\begin{abstract}
The idea behind the \emph{unsupervised} learning of \emph{disentangled} representations is that real-world data is generated by a few explanatory factors of variation which can be recovered by unsupervised learning algorithms.
In this paper, we provide a sober look at recent progress in the field and challenge some common assumptions.
We first theoretically show that the unsupervised learning of disentangled representations is fundamentally impossible without inductive biases on both the models and the data.
Then, we train over $\num{14000}$ models covering most prominent methods and evaluation metrics in a reproducible large-scale experimental study on eight data sets.
We observe that while the different methods successfully enforce properties ``encouraged'' by the corresponding losses, well-disentangled models seemingly cannot be identified without supervision.
Furthermore, different evaluation metrics do not always agree on what should be considered ``disentangled'' and exhibit systematic differences in the estimation. 
Finally, increased disentanglement does not seem to necessarily lead to a decreased sample complexity of learning for downstream tasks. 
Our results suggest that future work on disentanglement learning should be explicit about the role of inductive biases and (implicit) supervision, investigate concrete benefits of enforcing disentanglement of the learned representations, and consider a reproducible experimental setup covering several data sets.
\end{abstract}


\section{Introduction}
In representation learning it is often assumed that real-world observations $\rvx$ (such as images or videos) are generated by a two-step generative process.
\looseness=-1First, a multivariate latent random variable $\rvz$ is sampled from a distribution $\P(\rvz)$.
Intuitively, $\rvz$ corresponds to semantically meaningful factors of variation of the observations (such as content and position of objects in an image).
Then, in a second step, the observation $\rvx$ is sampled from the conditional distribution $\P(\rvx|\rvz)$.
The key idea behind this model is that the high-dimensional data $\rvx$ can be explained by the substantially lower dimensional and semantically meaningful latent variable $\rvz$ which is mapped to the higher-dimensional space of observations $\rvx$.
Informally, the goal of representation learning is to find useful transformations $r(\rvx)$ of $\rvx$ that ``make it easier to extract useful information when building classifiers or other predictors'' \citep{bengio2013representation}. 

\looseness=-1A recent line of work has argued that representations that are \emph{disentangled} are an important step towards a better representation learning~\citep{bengio2013representation,peters2017elements,lecun2015deep,bengio2007scaling,schmidhuber1992learning,lake2017building,tschannen2018recent}. They should contain all the information present in $\rvx$ in a compact and interpretable structure~\citep{bengio2013representation,kulkarni2015deep,chen2016infogan} while being independent from the task at hand~\citep{goodfellow2009measuring,lenc2015understanding}. 
They should be useful for (semi-)supervised learning of downstream tasks, transfer and few shot learning~\citep{bengio2013representation,scholkopf2012causal,peters2017elements}. They should enable to integrate out nuisance factors~\citep{kumar2017variational}, to perform interventions, and to answer counterfactual questions~\citep{pearl2009causality,SpiGlySch93,peters2017elements}. 

While there is no single formalized notion of disentanglement (yet) which is widely accepted, the key intuition is that a disentangled representation should separate the distinct, informative \emph{factors of variations} in the data \citep{bengio2013representation}.
A change in a single underlying factor of variation $\rz_i$ should lead to a change in a single factor in the learned representation $r(\rvx)$. This assumption can be extended to groups of factors as, for instance, in the work of \citet{bouchacourt2017multi} or \citet{suter2018interventional}. 
Based on this idea, a variety of disentanglement evaluation protocols have been proposed leveraging the statistical relations between the learned representation and the ground-truth factor of variations. 
Disentanglement is then measured as a particular structural property of these relations~\citep{higgins2016beta,kim2018disentangling,eastwood2018framework,kumar2017variational,chen2018isolating,ridgeway2018learning}. We can group the disentanglement scores in two categories. The scores proposed by~\cite{higgins2016beta},~\citet{kim2018disentangling} and~\citet{suter2018interventional} all require interventions. The first two involve intervening on a factor of variation for each batch and then predicting which factor was intervened on and the third one measures deviations in the latent space after performing the intervention. The scores proposed by ~\citet{eastwood2018framework,kumar2017variational,chen2018isolating,ridgeway2018learning} first construct a matrix of relation between factors of variation and codes (for example pairwise mutual information) and then aggregate this matrix into a single final number. Typically, this step involves computing some normalized gap between the largest and second largest entries either row or column-wise.

State-of-the-art approaches for unsupervised disentanglement learning are largely based on \emph{Variational Autoencoders (VAEs)} \citep{kingma2013auto}:
One assumes a specific prior $\P(\rvz)$ on the latent space and then uses a deep neural network to parameterize the conditional probability $\P(\rvx|\rvz)$. 
Similarly, the distribution $\P(\rvz|\rvx)$ is approximated using a variational distribution  $\Q(\rvz|\rvx)$, again parametrized using a deep neural network.
The model is then trained by minimizing a suitable approximation to the negative log-likelihood. The representation for $r(\rvx)$ is usually taken to be the mean of the approximate posterior distribution $\Q(\rvz|\rvx)$. 
Several variations of VAEs were proposed with the motivation that they lead to better disentanglement~\citep{higgins2016beta,burgess2018understanding,kim2018disentangling,chen2018isolating,kumar2017variational}.
The common theme behind all these approaches is that they try to enforce a factorized aggregated posterior $\int_{\rvx}\Q(\rvz|\rvx)\P(\rvx)d\rvx$, which should encourage disentanglement.

\subsection{Our Contributions}
In this paper, we challenge commonly held assumptions in this field in both theory and practice. 
Our key contributions can be summarized as follows:
\begin{itemize}[itemsep=2pt,topsep=3pt, leftmargin=10pt]
\item We theoretically prove that (perhaps unsurprisingly) the unsupervised learning of disentangled representations is fundamentally impossible without inductive biases both on the considered learning approaches and the data sets.
\item We investigate current approaches and their inductive biases in a reproducible large-scale experimental study\footnote{Reproducing these experiments requires approximately 2.92 GPU years (NVIDIA P100).} with a sound experimental protocol for unsupervised disentanglement learning. 
We implement six recent unsupervised disentanglement learning methods as well as seven disentanglement measures from scratch and train more than $\num{14000}$ models on eight data sets. 
\item We release \texttt{disentanglement\_lib}\footnote{\url{https://github.com/google-research/disentanglement_lib}}, a new library to train and evaluate disentangled representations. As reproducing our results requires substantial computational effort, we also release more than \num{10000} trained models which can be used as baselines for future research.
\item We analyze our experimental results and challenge common beliefs in unsupervised disentanglement learning:
(i) While all considered methods prove effective at ensuring that the individual dimensions of the aggregated posterior (which is sampled) are not correlated, we observe that the dimensions of the representation (which is taken to be the mean) are correlated.
(ii) We do not find any evidence that the considered models can be used to reliably learn disentangled representations in an \emph{unsupervised} manner as random seeds and hyperparameters seem to matter more than the model choice.
Furthermore, good trained models seemingly cannot be identified without access to ground-truth labels even if we are allowed to transfer good hyperparameter values across data sets. 
(iii) We observe systematic differences in the evaluation of disentangled representations. These differences arise both from how disentanglement is ``defined'' and how the relations between factors of variation and the dimensions of the representation are estimated.
(iv) For the considered models and data sets, we cannot validate the assumption that disentanglement is useful for downstream tasks, for example through a decreased sample complexity of learning.
\item Based on these empirical evidence, we suggest three critical areas of further research:
(i) The role of inductive biases and implicit and explicit supervision should be made explicit: unsupervised model selection persists as a key question.
(ii) The concrete practical benefits of enforcing a specific notion of disentanglement of the learned representations should be demonstrated.
(iii) Experiments should be conducted in a reproducible experimental setup on data sets of varying degrees of difficulty and with a clear evaluation protocol.

\end{itemize}

\subsection{Roadmap}
In Section~\ref{sec:other_rel_work} we briefly discuss other related works. In Section~\ref{sec:impossibility}, we present our theoretical result with extensive discussion of its implications. In Section~\ref{sec:experimental_design}, we discuss our experimental design. In Sections~\ref{sec:learning},~\ref{sec:evaluating},  and~\ref{sec:downstream} we present the results of our experimental studies concerning the training, evaluation metrics, and downstream performance respectively. In Section~\ref{sec:conclusion}, we summarize the implications of our findings and highlight directions for future research.

\section{Other Related Work}\label{sec:other_rel_work}
In a similar spirit to disentanglement, (non-)linear independent component analysis~\citep{comon1994independent,bach2002kernel,jutten2003advances, hyvarinen2016unsupervised} studies the problem of recovering independent components of a signal. The underlying assumption is that there is a generative model for the signal composed of the combination of statistically independent non-Gaussian components. While the
identifiability result for linear ICA \citep{comon1994independent} proved to be a milestone for the classical theory of factor analysis, similar results are in general not obtainable for the nonlinear case and the underlying sources generating the data cannot be identified
 \citep{hyvarinen1999nonlinear}. The lack of almost any identifiability result in non-linear ICA has been a main bottleneck for the utility of the approach \citep{hyvarinen2018nonlinear} and partially motivated alternative machine learning approaches~\citep{desjardins2012disentangling,schmidhuber1992learning,cohen2014transformation}.
Given that unsupervised algorithms did not initially perform well on realistic settings most of the other works have considered some more or less explicit form of supervision~\citep{reed2014learning,zhu2014multi,yang2015weakly,kulkarni2015deep,cheung2014discovering,mathieu2016disentangling,narayanaswamy2017learning,suter2018interventional}.~\citep{hinton2011transforming,cohen2014learning} assume some knowledge of the effect of the factors of variations even though they are not observed. One can also exploit known relations between factors in different samples~\citep{karaletsos2015bayesian,goroshin2015learning,whitney2016understanding,fraccaro2017disentangled,denton2017unsupervised,hsu2017unsupervised,yingzhen2018disentangled,locatello2018clustering}. This is not a limiting assumption especially in sequential data like for videos. There is for example a rich literature in disentangling pose from content in 3D objects and content from motion in videos or time series in general~\citep{yang2015weakly,li2018disentangled,hsieh2018learning,fortuin2018deep,deng2017factorized,goroshin2015learning}.
Similarly, the non-linear ICA community recently shifted to non-iid data types exploiting time dependent or grouped observations~\citep{hyvarinen2016unsupervised,hyvarinen2018nonlinear,gresele2019incomplete}

We focus our study on the setting where factors of variations are not observable at all, that is, we only observe samples from $\P(\rvx)$. 

\clearpage
\section{Impossibility Result}\label{sec:impossibility}
The first question that we investigate is whether unsupervised disentanglement learning is even possible for arbitrary generative models.
Theorem~\ref{thm:impossibility} essentially shows that without inductive biases both on models and data sets the task is fundamentally impossible.
The proof is provided in Appendix~\ref{sec:proof}. 
\begin{theorem}
\label{thm:impossibility}
For $d>1$, let $\rvz\sim\P$ denote any distribution which admits a density $p(\rvz)=\prod_{i=1}^dp(\rz_i)$.
Then, there exists an infinite family of bijective functions $f:\supp(\rvz)\to\supp(\rvz)$ such that $\frac{\partial f_i(\vu)}{\partial u_j} \neq 0$ almost everywhere for all $i$ and $j$ (implying that $\rvz$ and $f(\rvz)$ are completely entangled) and $\P(\rvz \leq \vu) = \P(f(\rvz) \leq \vu)$ for all $\vu\in\supp(\rvz)$ (they have the same marginal distribution). 
\end{theorem}

Consider the commonly used ``intuitive'' notion of disentanglement which advocates that a change in a single ground-truth factor should lead to a single change in the representation.
In that setting, Theorem~\ref{thm:impossibility} implies that unsupervised disentanglement learning is \emph{impossible} for arbitrary generative models with a factorized prior\footnote{Theorem~\ref{thm:impossibility} only applies to factorized priors; however, we expect that a similar result can be extended to non-factorizing priors.} in the following sense:
Assume we have $p(\rvz)$ and some $P(\rvx|\rvz)$ defining a generative model. Consider any unsupervised disentanglement method and assume that it finds a representation $r(\rvx)$ that is perfectly disentangled with respect to $\rvz$ in the generative model.
Then, Theorem~\ref{thm:impossibility} implies that there is an equivalent generative model with the latent variable $\hat{\rvz}=f(\rvz)$ where $\hat{\rvz}$ is completely \emph{entangled} with respect to $\rvz$ and thus also $r(\rvx)$: as all the entries in the Jacobian of $f$ are non-zero, a change in a single dimension of $\rvz$ implies that all dimensions of $\hat{\rvz}$ change.
Furthermore, since $f$ is deterministic and $p(\rvz)=p(\hat{\rvz})$ almost everywhere, both generative models have the same marginal distribution of the observations $\rvx$ by construction, that is, $P(\rvx) = \int p(\rvx| \rvz)p(\rvz) d\rvz = \int p(\rvx|\hat{\rvz})p(\hat{\rvz}) d\hat{\rvz}$.
Since the (unsupervised) disentanglement method only has access to observations $\rvx$, it hence cannot distinguish between the two equivalent generative models and thus has to be entangled to at least one of them.

\looseness=-1This may not be surprising to readers familiar with the causality and ICA literature as it is consistent with the following argument: 
After observing $\rvx$, we can construct infinitely many generative models which have the same marginal distribution of $\rvx$.
Any one of these models could be the true causal generative model for the data, and the right model cannot be identified given only the distribution of $\rvx$ \citep{peters2017elements}.
Similar results have been obtained in the context of non-linear ICA~\citep{hyvarinen1999nonlinear}. 
The main novelty of Theorem~\ref{thm:impossibility} is that it allows the explicit construction of latent spaces $\rvz$ and $\hat{\rvz}$ that are completely \textit{entangled} with each other in the sense of~\citep{bengio2013representation}.
We note that while this result is very intuitive for multivariate Gaussians it also holds for distributions which are not invariant to rotation, for example multivariate uniform distributions.

While Theorem~\ref{thm:impossibility} shows that unsupervised disentanglement learning is fundamentally impossible for arbitrary generative models, this does not necessarily mean it is an impossible endeavour in practice.
After all, real world generative models may have a certain structure that could be exploited through suitably chosen inductive biases.
However, Theorem~\ref{thm:impossibility} clearly shows that inductive biases are required both for the models (so that we find a specific set of solutions) and for the data sets (such that these solutions match the true generative model).
We hence argue that the role of inductive biases should be made explicit and investigated further as done in the following experimental study.

\section{Experimental Design}\label{sec:experimental_design}
In this section, we discuss the methods, evaluation metrics, data sets and overall experimental conditions of our study.
\subsection{Considered Methods} 
All the considered methods augment the VAE loss with a regularizer: The $\beta$-VAE~\citep{higgins2016beta}, introduces a hyperparameter in front of the KL regularizer of vanilla VAEs to constrain the capacity of the VAE bottleneck.
The AnnealedVAE~\citep{burgess2018understanding} progressively increase the bottleneck capacity so that the encoder can focus on learning one factor of variation at the time (the one that most contribute to a small reconstruction error). The FactorVAE~\citep{kim2018disentangling} and the $\beta$-TCVAE~\citep{chen2018isolating} penalize the total correlation~\citep{watanabe1960information} with adversarial training~\citep{nguyen2010estimating,sugiyama2012density} or with a tractable but biased Monte-Carlo estimator respectively.
The DIP-VAE-I and the DIP-VAE-II~\citep{kumar2017variational} both penalize the mismatch between the aggregated posterior and a factorized prior.
Implementation details can be found in Appendix~\ref{app:hyperparameters}.
\subsubsection{Unsupervised Learning of Disentangled Representations with VAEs}~\label{app:methods}
Variants of variational autoencoders~\cite{kingma2013auto} are considered the state-of-the-art for unsupervised disentanglement learning. 
One assumes a specific prior $\P(\rvz)$ on the latent space and then parameterizes the conditional probability $\P(\rvx|\rvz)$ with a deep neural network. 
Similarly, the distribution $\P(\rvz|\rvx)$ is approximated using a variational distribution  $\Q(\rvz|\rvx)$, again parametrized using a deep neural network. One can then derive the following approximation to the maximum likelihood objective, 
\begin{align}
\max_{\phi, \theta}\quad \E_{p(\rvx)} [\E_{q_\phi(\rvz|\rvx)}[\log p_\theta(\rvx|\rvz)] -  \KL(q_\phi(\rvz|\rvx) \| p(\rvz))]\label{eq:elbo}
\end{align}
which is also know as the evidence lower bound (ELBO). By carefully considering the KL term, one can encourage various properties of the resulting presentation. We will briefly review the main approaches. We now briefly categorize the different approaches.

\subsubsection{Bottleneck Capacity}
 \citet{higgins2016beta} propose the $\beta$-VAE, introducing a hyperparameter in front of the KL regularizer of vanilla VAEs. They maximize the following expression:
\begin{align*}
\E_{p(\rvx)} [\E_{q_\phi(\rvz|\rvx)}[\log p_\theta(\rvx|\rvz)] - \beta \KL(q_\phi(\rvz|\rvx) \| p(\rvz))]
\end{align*} 
By setting $\beta > 1$, the encoder distribution will be forced to better match the factorized unit Gaussian prior. This procedure introduces additional constraints on the capacity of the latent bottleneck, encouraging the encoder to learn a disentangled representation for the data. 
~\citet{burgess2018understanding} argue that when the bottleneck has limited capacity, the network will be forced to specialize on the factor of variation that most contributes to a small reconstruction error. Therefore, they propose to progressively increase the bottleneck capacity, so that the encoder can focus on learning one factor of variation at the time:
\begin{align*}
\E_{p(\rvx)} [\E_{q_\phi(\rvz|\rvx)}[\log p_\theta(\rvx|\rvz)] - \gamma | \KL(q_\phi(\rvz|\rvx) \| p(\rvz)) - C |]
\end{align*}
where C is annealed from zero to some value which is large enough to produce good reconstruction. 
In the following, we refer to this model as AnnealedVAE.

\subsubsection{Penalizing the Total Correlation}
Let $I(\rvx; \rvz)$ denote the mutual information between $\rvx$ and $\rvz$ and note that the second term in~\eqref{eq:elbo} can be rewritten as 
\begin{align*}
\E_{p(\rvx)} [\KL(q_\phi(\rvz|\rvx) \| p(\rvz))] = I(\rvx;\rvz) + \KL(q(\rvz) \| p(\rvz)).
\end{align*}
Therefore, when $\beta>1$, $\beta$-VAE penalizes the mutual information between the latent representation and the data, thus constraining the capacity of the latent space. Furthermore, it pushes $q(\rvz)$, the so called \textit{aggregated posterior}, to match the prior and therefore to factorize, given a factorized prior. 
\citet{kim2018disentangling} argues that penalizing $I(\rvx;\rvz)$ is neither necessary nor desirable for disentanglement. The FactorVAE~\citep{kim2018disentangling} and the $\beta$-TCVAE~\citep{chen2018isolating} augment the VAE objective with an additional regularizer that specifically penalizes dependencies between the dimensions of the representation:
\begin{align*}
\E_{p(\rvx)} [\E_{q_\phi(\rvz|\rvx)}[\log p_\theta(\rvx|\rvz)] -  \KL(q_\phi(\rvz|\rvx) \| p(\rvz))] - \gamma \KL(q(\rvz)\| \prod_{j=1}^d q(\rz_j)).
\end{align*}
This last term is also known as \textit{total correlation}~\citep{watanabe1960information}. The total correlation is intractable and vanilla Monte Carlo approximations require marginalization over the training set. \citep{kim2018disentangling} propose an estimate using the density ratio trick~\citep{nguyen2010estimating,sugiyama2012density} (FactorVAE). Samples from $\prod_{j=1}^d q(\rz_j)$ can be obtained shuffling samples from $q(\rvz)$~\citep{arcones1992bootstrap}. Concurrently, \citet{chen2018isolating} propose a tractable biased Monte-Carlo estimate for the total correlation ($\beta$-TCVAE).

\subsubsection{Disentangled Priors}
\citet{kumar2017variational} argue that a disentangled generative model requires a disentangled prior. This approach is related to the total correlation penalty, but now the aggregated posterior is pushed to match a factorized prior. Therefore
\begin{align*}
&\E_{p(\rvx)} [ \E_{q_\phi(\rvz|\rvx)}[\log p_\theta(\rvx|\rvz)] - \KL(q_\phi(\rvz|\rvx) \| p(\rvz))] - \lambda D(q(\rvz)\|p(\rvz)),
\end{align*}
where $D$ is some (arbitrary) divergence. Since this term is intractable when $D$ is the KL divergence, they propose to match the moments of these distribution. In particular, they regularize the deviation of either $\mathrm{Cov}_{p(\rvx)}[\mu_\phi(\rvx)]$ or $\mathrm{Cov}_{q_\phi}[\rvz]$ from the identity matrix in the two variants of the DIP-VAE. This results in maximizing either the DIP-VAE-I objective
\begin{align*}
\E_{p(\rvx)} [ \E_{q_\phi(\rvz|\rvx)}[\log p_\theta(\rvx|\rvz)] &- \KL(q_\phi(\rvz|\rvx) \| p(\rvz))] - \lambda_{od}\sum_{i\neq j} \left[\mathrm{Cov}_{p(\rvx)}[\mu_\phi(\rvx)]\right]_{ij}^2\\&- \lambda_{d}\sum_{i} \left(\left[\mathrm{Cov}_{p(\rvx)}[\mu_\phi(\rvx)]\right]_{ii} - 1\right)^2
\end{align*}
or the DIP-VAE-II objective
\begin{align*}
\E_{p(\rvx)} [ \E_{q_\phi(\rvz|\rvx)}[\log p_\theta(\rvx|\rvz)] &- \KL(q_\phi(\rvz|\rvx) \| p(\rvz))] - \lambda_{od}\sum_{i\neq j} \left[\mathrm{Cov}_{q_\phi}[\rvz]\right]_{ij}^2\\ &-\lambda_{d}\sum_{i} \left(\left[\mathrm{Cov}_{q_\phi}[\rvz]\right]_{ii} - 1\right)^2.
\end{align*}

\subsection{Considered Metrics}
\looseness=-1The \emph{BetaVAE} metric~\citep{higgins2016beta} measures disentanglement as the accuracy of a linear classifier that predicts the index of a fixed factor of variation.
\citet{kim2018disentangling} address several issues with this metric in their \emph{FactorVAE} metric by using a majority vote classifier on a different feature vector which accounts for a corner case in the BetaVAE metric.
The \emph{Mutual Information Gap (MIG)}~\citep{chen2018isolating} measures for each factor of variation the normalized gap in mutual information between the highest and second highest coordinate in $r(\rvx)$. Instead, the \emph{Modularity}~\citep{ridgeway2018learning} measures if each dimension of $r(\rvx)$ depends on at most a factor of variation using their mutual information.
The metrics of~\citet{eastwood2018framework}  compute the entropy of the distribution obtained by normalizing the importance of each dimension of the learned representation for predicting the value of a factor of variation. Their disentanglement score (which we call \emph{DCI Disentanglement} for clarity) penalizes multiple factors of variation being captured by the same code and their completeness score (which we call \emph{DCI Completeness}) penalizes a factor of variation being captured by multiple codes.
The \emph{SAP score}~\citep{kumar2017variational} is the average difference of the prediction error of the two most predictive latent dimensions for each factor. 
The \emph{Interventional Robustness Score (IRS)}~\citep{suter2018interventional} measures whether the representation is robustly disentangled by performing interventions on the factors of variations and measuring deviations in the latent space.
Finally, we note that MIG, DCI Disentanglement, Modularity and SAP scores all involves the estimation of a matrix relating the factors of variation to the latent codes. Then, this matrix is aggregated into a score following some different disentanglement notion. In order to understand the role of each of these two steps we separate them and consider blends of these scores. For example, we compute the mutual information matrix as in the MIG or Modularity but compute the score using the DCI Disentanglement aggregation. We call this score \emph{MIG-DCI Disentanglement}. In our experiments, we consider all possible pairs of matrix and aggregation.

All our metrics consider the expected representation of training samples (except total correlation for which we also consider the sampled representation as described in Section~\ref{sec:learning}).

\subsubsection{BetaVAE Metric} \looseness=-1\citet{higgins2016beta} suggest to fix a random factor of variation in the underlying generative model and to sample two mini batches of observations $\rvx$. 
Disentanglement is then measured as the accuracy of a linear classifier that predicts the index of the fixed factor based on the coordinate-wise sum of absolute differences between the representation vectors in the two mini batches.
We sample two batches of 64 points with a random factor fixed to a randomly sampled value across the two batches and the others varying randomly. 
We compute the mean representations for these points and take the absolute difference between pairs from the two batches. 
We then average these 64 values to form the features of a training (or testing) point. 
We train a Scikit-learn logistic regression with default parameters on $\num{10000}$ points. 
We test on $\num{5000}$ points.

\subsubsection{FactorVAE Metric}
\looseness=-1\citet{kim2018disentangling} address several issues with this metric by using a majority vote classifier that predicts the index of the fixed ground-truth factor based on the index of the representation vector with the least variance.
First, we estimate the variance of each latent dimension by embedding $\num{10000}$ random samples from the data set and we exclude collapsed dimensions with variance smaller than 0.05. 
Second, we generate the votes for the majority vote classifier by sampling a batch of 64 points, all with a factor fixed to the same random value. 
Third, we compute the variance of each dimension of their latent representation and divide by the variance of that dimension we computed on the data without interventions. 
The training point for the majority vote classifier consists of the index of the dimension with the smallest normalized variance. 
We train on $\num{10000}$ points and evaluate on $\num{5000}$ points.

\subsubsection{Mutual Information Gap} 
\citet{chen2018isolating} argue that the BetaVAE metric and the FactorVAE metric are neither general nor unbiased as they depend on some hyperparameters. 
They compute the mutual information between each ground truth factor and each dimension in the computed representation $r(\rvx)$.
For each ground-truth factor $\rz_k$, they then consider the two dimensions in $r(\rvx)$ that have the highest and second highest mutual information with $\rz_k$.
The \emph{Mutual Information Gap (MIG)} is then defined as the average, normalized difference between the highest and second highest mutual information of each factor with the dimensions of the representation.
The original metric was proposed evaluating the sampled representation. 
Instead, we consider the mean representation, in order to be consistent with the other metrics. 
We estimate the discrete mutual information by binning each dimension of the representations obtained from $\num{10000}$ points into 20 bins. 
Then, the score is computed as follows:
\begin{align*}
\frac{1}{K}\sum_{k=1}^K \frac{1}{H_{\rz_k}}\left(I(\rv_{j_k}, \rz_k) - \max_{j\neq j_k} I(\rv_j, \rz_k)\right),
\end{align*}
where $\rz_k$ is a factor of variation, $\rv_j$ is a dimension of the latent representation, $H_{\rz_k}$ is the entropy of $\rz_k$ (using again 20 bins), and $j_k = \argmax_j I(\rv_j,\rz_k)$.

\subsubsection{Modularity}
\citet{ridgeway2018learning} argue that two different properties of representations should be considered: \emph{Modularity} and \emph{Explicitness}.
In a modular representation each dimension of $r(\rvx)$ depends on at most a single factor of variation. 
In an explicit representation, the value of a factor of variation is easily predictable (for example with a linear model) from $r(\rvx)$. 
They propose to measure the Modularity as the average normalized squared difference of the mutual information of the factor of variations with the highest and second-highest mutual information with a dimension of $r(\rvx)$.
They measure Explicitness as the ROC-AUC of a one-versus-rest logistic regression classifier trained to predict the factors of variation.
In this study, we focus on Modularity as it is the property that corresponds to disentanglement.
For the modularity score, we sample $\num{10000}$ points for which we obtain the latent representations. 
We discretize these points into 20 bins and compute the mutual information between representations and the values of the factors of variation. 
These values are stored in a matrix $\rvm$. 
For each dimension of the representation $i$, we compute a vector $\rvt_i$ as:
\begin{align*}
t_{i,f} = \begin{cases} \theta_i & \mbox{if } f=\argmax_g m_{i,g}\\ 0 & \mbox{otherwise}\end{cases}
\end{align*}
 where $\theta_i = \max_g m_{ig}$. The modularity score is the average over the dimensions of the representation of $1-\delta_i$ where:
 \begin{align*}
 \delta_i = \frac{\sum_f (m_{if}-t_{if})^2}{\theta_i^2(N-1)}
 \end{align*}
 and N is the number of factors. 
 
 \subsubsection{DCI Disentanglement}
\looseness=-1\citet{eastwood2018framework} consider three properties of representations: \emph{Disentanglement}, \emph{Completeness} and \emph{Informativeness}.
First,~\citet{eastwood2018framework} compute the importance of each dimension of the learned representation for predicting a factor of variation. 
The predictive importance of the dimensions of $r(\rvx)$ can be computed with a Lasso or a Random Forest classifier.
Disentanglement is the average of the difference from one of the entropy of the probability that a dimension of the learned representation is useful for predicting a factor weighted by the relative importance of each dimension. 
Completeness, is the average of the difference from one of the entropy of the probability that a factor of variation is captured by a dimension of the learned representation. 
Finally, the Informativeness can be computed as the prediction error of predicting the factors of variations.
 We sample $\num{10000}$ and $\num{5000}$ training and test points respectively. 
 For each factor, we fit gradient boosted trees from Scikit-learn with the default setting. 
 From this model, we extract the importance weights for the feature dimensions. 
 We take the absolute value of these weights and use them to form the importance matrix $R$, whose rows correspond to factors and columns to the representation. 
 To compute the disentanglement score, we first subtract from 1 the entropy of each column of this matrix (we treat the columns as a distribution by normalizing them). 
 This gives a vector of length equal to the dimensionality of the latent space. 
 Then, we compute the relative importance of each dimension by $\rho_i = \sum_j R_{ij}/\sum_{ij} R_{ij}$ and the disentanglement score as $\sum_i \rho_i (1-H(R_i))$. 
 
 \subsubsection{SAP Score}
 \citet{kumar2017variational} propose to compute the $R^2$ score of the linear regression predicting the factor values from each dimension of the learned representation. 
For discrete factors, they propose to train a classifier. 
The \emph{Separated Attribute Predictability (SAP)} score is the average difference of the prediction error of the two most predictive latent dimensions for each factor.
 We sample $\num{10000}$ points for training and $\num{5000}$ for testing.
 We then compute a score matrix containing the prediction error on the test set for a linear SVM with $C=0.01$ predicting the value of a factor from a single latent dimension. 
 The SAP score is computed as the average across factors of the difference between the top two most predictive latent dimensions.
 
 \subsubsection{Interventional Robustness Score}
 \citet{suter2018interventional} introduce a causality perspective and measure the robustness of a representation after interventions on the factors of variation: the \emph{Interventional Robustness Score (IRS)}. For two factors of variation $\rvz_i$ and $\rvz_j$, they define the \textit{post interventional disagreement} as the distance between the representation with an intervention on $\rvz_i$ and on both $\rvz_i$ and $\rvz_j$. Then, they take the supremum of this distance with respect to the values of $\rvz_j$ and average with respect to the distribution of $\rvz_i$. This value is normalized by the maximum post interventional disagreement with no fixed $\rvz_i$ and subtracted from 1. This score measure essentially how well $\rvz_i$ is robustly disentangled from $\rvz_j$. The disentanglement of $\rvz_i$ can be computed by taking its maximum disagreement with all other factors of variation for each dimension dimension of the representation.
 
 \subsubsection{Downstream Task} 
 We sample training sets of different sizes: $\num{10}$, $\num{100}$, $\num{1000}$ and $\num{10000}$ points. We always evaluate on $\num{5000}$ samples. We consider as a downstream task the prediction of the values of each factor from $r(\rvx)$. For each factor we fit a different model and report then report the average test accuracy across factors. We consider two different models. First, we train a cross validated logistic regression from Scikit-learn with 10 different values for the regularization strength ($Cs = 10$) and $5$ folds. Finally, we train a gradient boosting classifier from Scikit-learn with default parameters.
 
 \subsubsection{Total Correlation Based on Fitted Gaussian}
 We sample $\num{10000}$ points and obtain their latent representation $r(\rvx)$ by either sampling from the encoder distribution of by taking its mean. We then compute the mean $\mu_{r(\rvx)}$ and covariance matrix $\Sigma_{r(\rvx)}$ of these points and compute the total correlation of a Gaussian with mean $\mu_{r(\rvx)}$ and covariance matrix $\Sigma_{r(\rvx)}$:
 \begin{align*}
 \KL\left(\mathcal{N}(\mu_{r(\rvx)},\Sigma_{r(\rvx)})\Big\|\prod_{j}\mathcal{N}(\mu_{r(\rvx)_j},\Sigma_{r(\rvx)_{jj}})\right),
 \end{align*}
 where $j$ indexes the dimensions in the latent space.
 We choose this approach for the following reasons.
 In this study, we compute statistics of $r(\rvx)$ which can be either sampled from the probabilistic encoder or taken to be its mean. We argue that estimating the total correlation as in~\citep{kim2018disentangling} is not suitable for this comparison as it consistently underestimates the true value (see Figure 7 in ~\citep{kim2018disentangling}) and depends on a non-convex optimization procedure (for fitting the discriminator). The estimate of~\citep{chen2018isolating} is also not suitable as the mean representation is a deterministic function for the data, therefore we cannot use the encoder distribution for the estimate. 
 Furthermore, we argue that the total correlation based on the fitted Gaussian provides a simple and robust way to detect if a representation is not factorizing based on the first two moments.
 In particular, if it is high, it is a strong signal that the representation is not factorizing (while a low score may not imply the opposite). 
We note that this procedure is similar to the penalty of DIP-VAE-I.

\subsection{Data Sets} We consider five data sets in which $\rvx$ is obtained as a deterministic function of $\rvz$: \textit{dSprites}~\citep{higgins2016beta}, \textit{Cars3D}~\citep{reed2015deep}, \textit{SmallNORB}~\citep{lecun2004learning}, \textit{Shapes3D}~\citep{kim2018disentangling} and \textit{MPI3D}~\citep{gondal2019transfer}.
We also introduce three data sets where the observations $\rvx$ are stochastic given the factor of variations $\rvz$: \textit{Color-dSprites}, \textit{Noisy-dSprites} and \textit{Scream-dSprites}. 
In \textit{Color-dSprites}, the shapes are colored with a random color. 
In \textit{Noisy-dSprites}, we consider white-colored shapes on a noisy background. 
Finally, in \textit{Scream-dSprites} the background is replaced with a random patch in a random color shade extracted from the famous \textit{The Scream} painting~\citep{munch_1893}. 
The dSprites shape is embedded into the image by inverting the color of its pixels. Further details on the preprocessing of the data can be found in Appendix~\ref{app:datasets}.

\subsection{Inductive Biases} 
To fairly evaluate the different approaches, we separate the effect of regularization (in the form of model choice and regularization strength) from the other inductive biases (for example, the choice of the neural architecture). 
Each method uses the same convolutional architecture, optimizer, hyperparameters of the optimizer and batch size. 
All methods use a Gaussian encoder where the mean and the log variance of each latent factor is parametrized by the deep neural network, a Bernoulli decoder and latent dimension fixed to 10. 
We note that these are all standard choices in prior work \citep{higgins2016beta,kim2018disentangling}.

We choose six different regularization strengths, that is, hyperparameter values, for each of the considered methods.
The key idea was to take a wide enough set to ensure that there are useful hyperparameters for different settings for each method and not to focus on specific values known to work for specific data sets.
However, the values are partially based on the ranges that are prescribed in the literature (including the hyperparameters suggested by the authors). 

\looseness=-1We fix our experimental setup in advance and we run all the considered methods on each data set for 50 different random seeds and evaluate them on the considered metrics. 
The full details on the experimental setup are provided in the Appendix~\ref{app:hyperparameters}.
Our experimental setup, the limitations of this study, and the differences with previous implementations  are extensively discussed in Appendices~\ref{app:guiding_principles}-\ref{app:differences}.

\section{Can We Learn Disentangled Representations Without Supervision?}\label{sec:learning}
In this section, we provide a sober look at the performances of state-of-the-art approaches and investigate how effectively we can learn disentangled representations without looking at the labels. We focus our analysis on key questions for practitioners interested in learning disentangled representations reliably and without supervision.

\begin{figure}[p!]
  \centering
  \begin{subfigure}{0.5\textwidth}
    {\adjincludegraphics[scale=0.4, trim={0 0 {.5\width} {.5\height}}, clip]{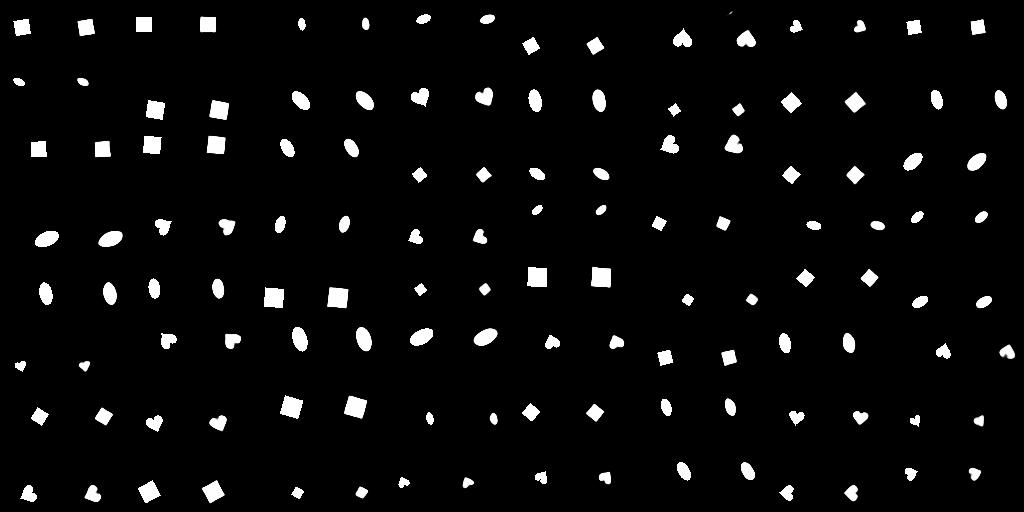}}\hspace{2mm}
    \caption{DIP-VAE-I trained on dSprites.}
  \end{subfigure}%
  \begin{subfigure}{0.5\textwidth}
    {\adjincludegraphics[scale=0.4, trim={0 0 {.5\width} {.5\height}}, clip]{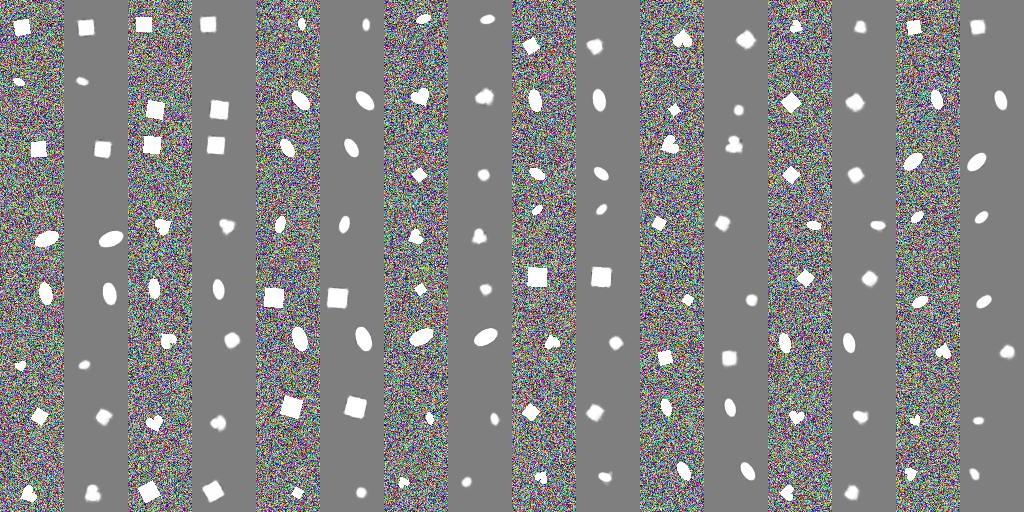}}
    \caption{$\beta$-VAE trained on Noisy-dSprites.}
  \end{subfigure}
  \begin{subfigure}{0.5\textwidth}
    {\adjincludegraphics[scale=0.4, trim={0 0 {.5\width} {.5\height}}, clip]{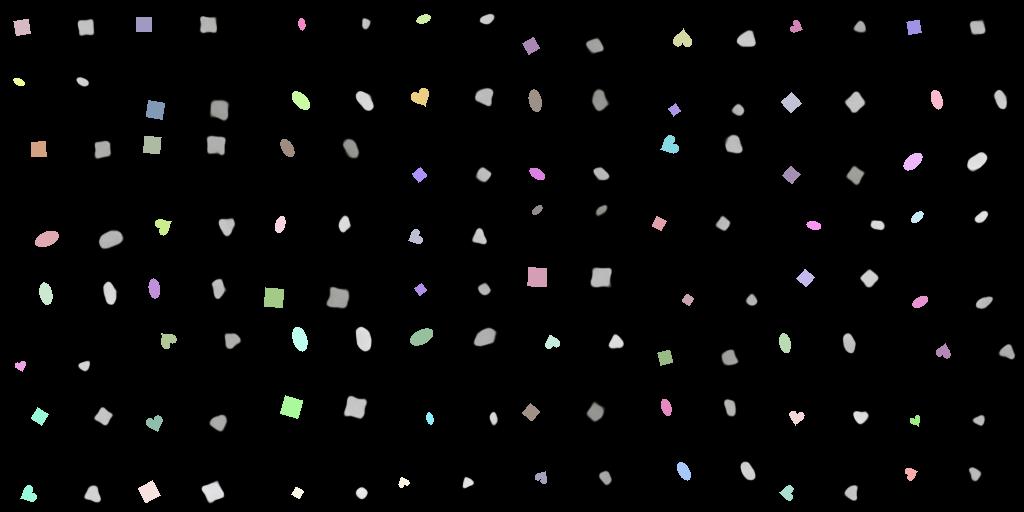}}\hspace{2mm}
    \caption{FactorVAE trained on Color-dSprites.}
  \end{subfigure}%
  \begin{subfigure}{0.5\textwidth}
    {\adjincludegraphics[scale=0.4, trim={0 0 {.5\width} {.5\height}}, clip]{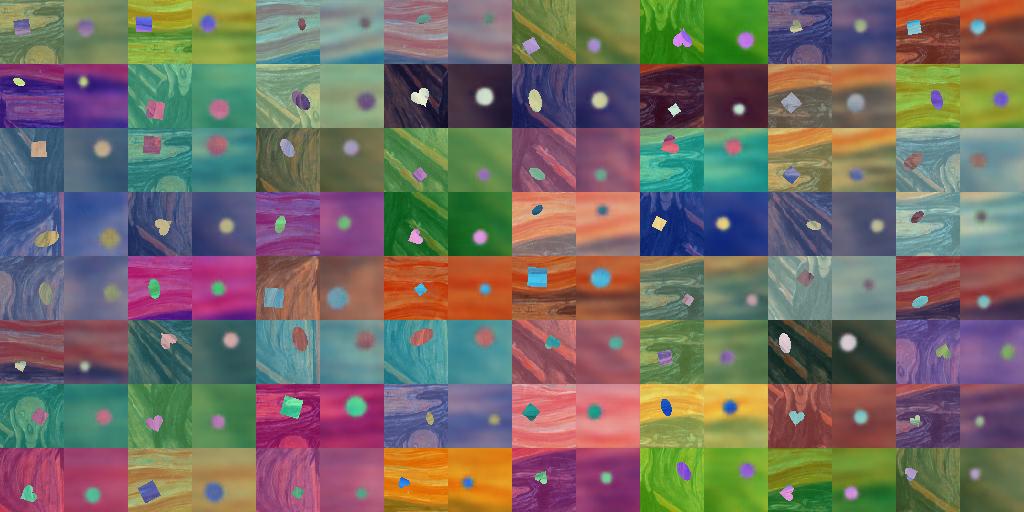}}
    \caption{FactorVAE trained on Scream-dSprites.}
  \end{subfigure}
  \begin{subfigure}{0.5\textwidth}
    {\adjincludegraphics[scale=0.4, trim={0 0 {.5\width} {.5\height}}, clip]{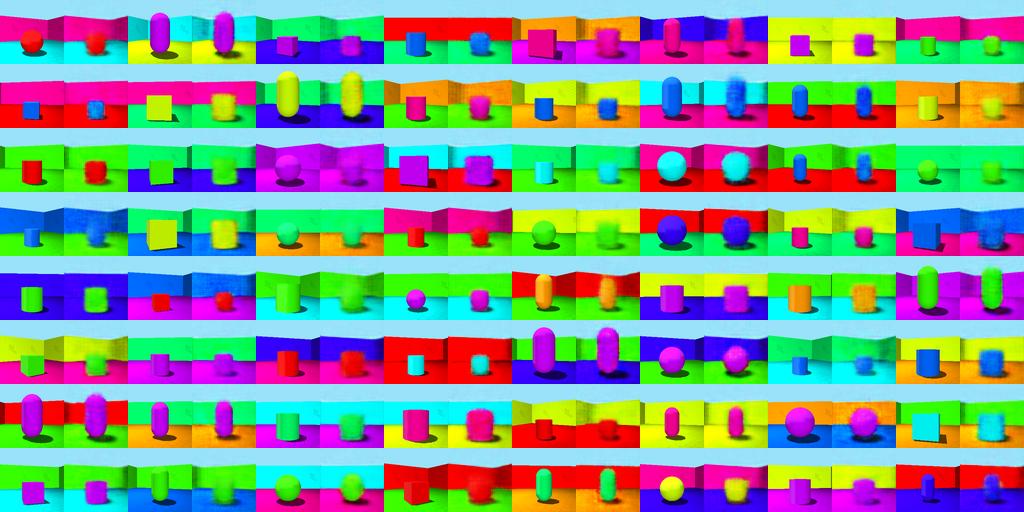}}\hspace{2mm}
    \caption{AnneaeledVAE trained on Shapes3D.}
  \end{subfigure}%
  \begin{subfigure}{0.5\textwidth}
    {\adjincludegraphics[scale=0.4, trim={0 0 {.5\width} {.5\height}}, clip]{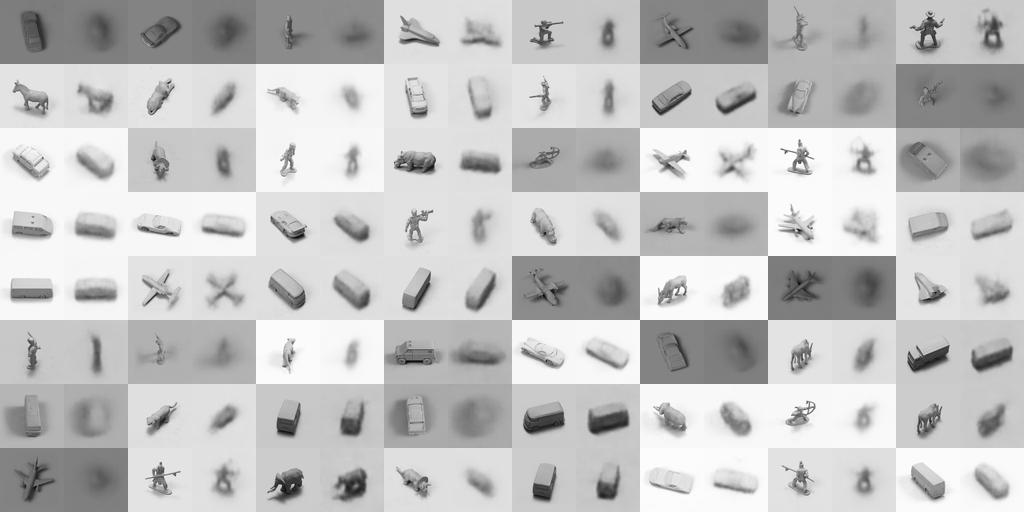}}
    \caption{$\beta$-TCVAE trained on SmallNORB.}
  \end{subfigure}
  \begin{subfigure}{0.5\textwidth}
    {\adjincludegraphics[scale=0.4, trim={0 0 {.5\width} {.5\height}}, clip]{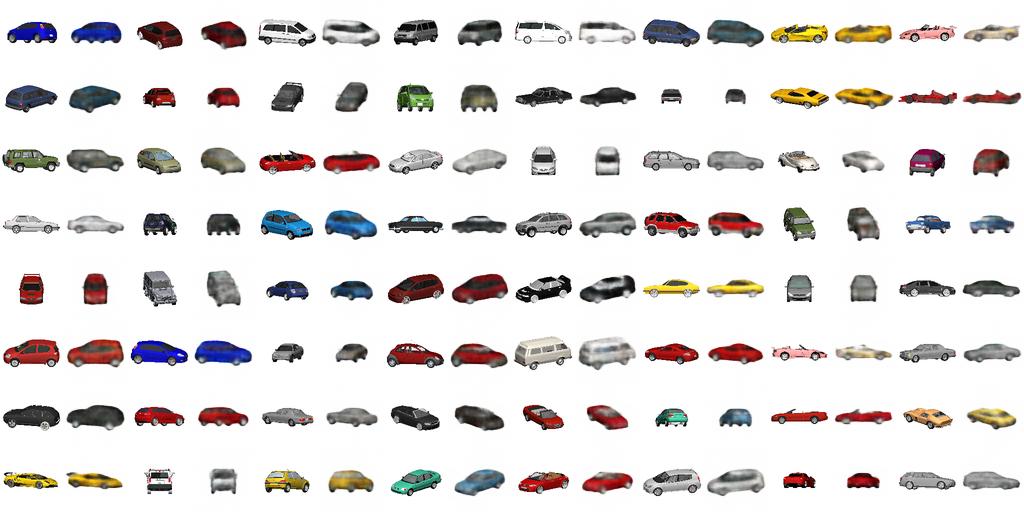}}
    \caption{DIP-VAE-II trained on Cars3D.}
  \end{subfigure}%
  \begin{subfigure}{0.5\textwidth}
    {\adjincludegraphics[scale=0.4, trim={0 0 {.5\width} {.5\height}}, clip]{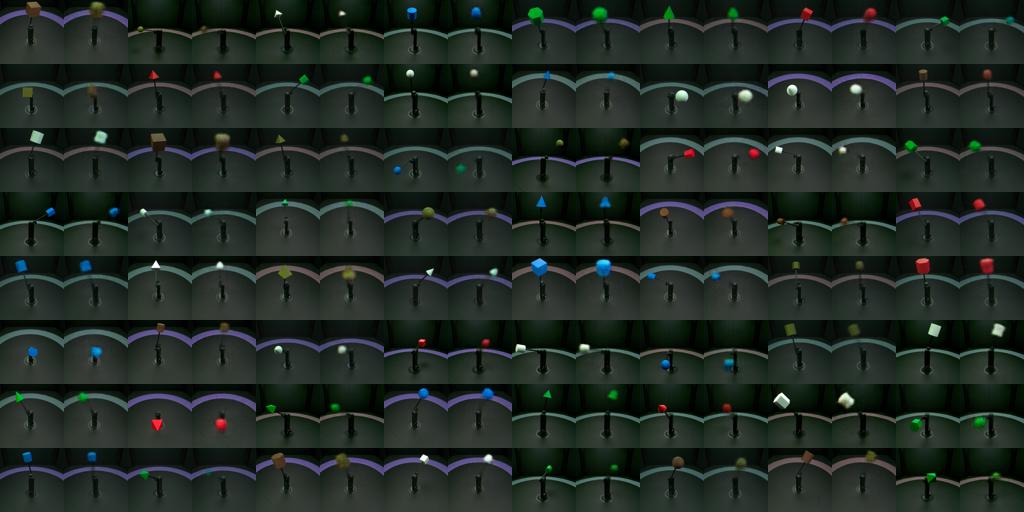}}
    \caption{$\beta$-VAE trained on MPI3D.}
  \end{subfigure}
  \caption{Reconstructions for different data sets and methods. Odd columns show real samples and even columns their reconstruction. As expected, the additional variants of dSprites with continuous noise variables are harder than the original data set.
    On Noisy-dSprites and Color-dSprites the models produce reasonable reconstructions with the noise on Noisy-dSprites being ignored.
    Scream-dSprites is even harder and we observe that the shape information is lost.
    On the other data sets, we observe that reconstructions are blurry but objects are distinguishable. The MPI3D Dataset consists of real images of a robotic arm. \label{figure:reconstructions}}
\end{figure} 

\subsection{Can One Achieve a Good Reconstruction Error Across Data Sets and Models?}
First, we check for each data set that we manage to train models that achieve reasonable reconstructions. 
Therefore, for each data set we sample a random model and show real samples next to their reconstructions.
The results are depicted in Figure~\ref{figure:reconstructions}. 
As expected, the additional variants of dSprites with continuous noise variables are harder than the original data set.
On Noisy-dSprites and Color-dSprites the models produce reasonable reconstructions with the noise on Noisy-dSprites being ignored.
Scream-dSprites is even harder and we observe that the shape information is lost.
On the other data sets, we observe that reconstructions are blurry but objects are distinguishable. Since in MPI3D the objects are small, their shape appear sometime difficult to distinguish. The other factors of variation however are clearly captured.
SmallNORB seems to be the most challenging data set.

\begin{figure}[p]
\centering\includegraphics[width=\textwidth]{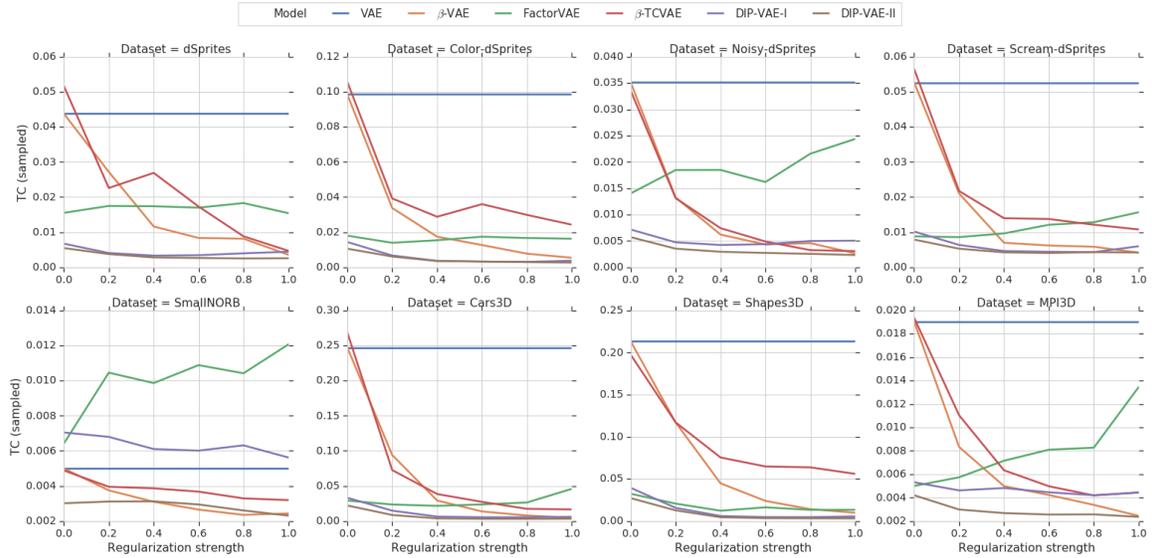}
\caption{Total correlation of sampled representation plotted against regularization strength for different data sets and approaches (except AnnealedVAE). The total correlation of the sampled representation decreases as the regularization strength is increased.}\label{figure:TCsampled}
\end{figure}
\begin{figure}[p]
\centering\includegraphics[width=\textwidth]{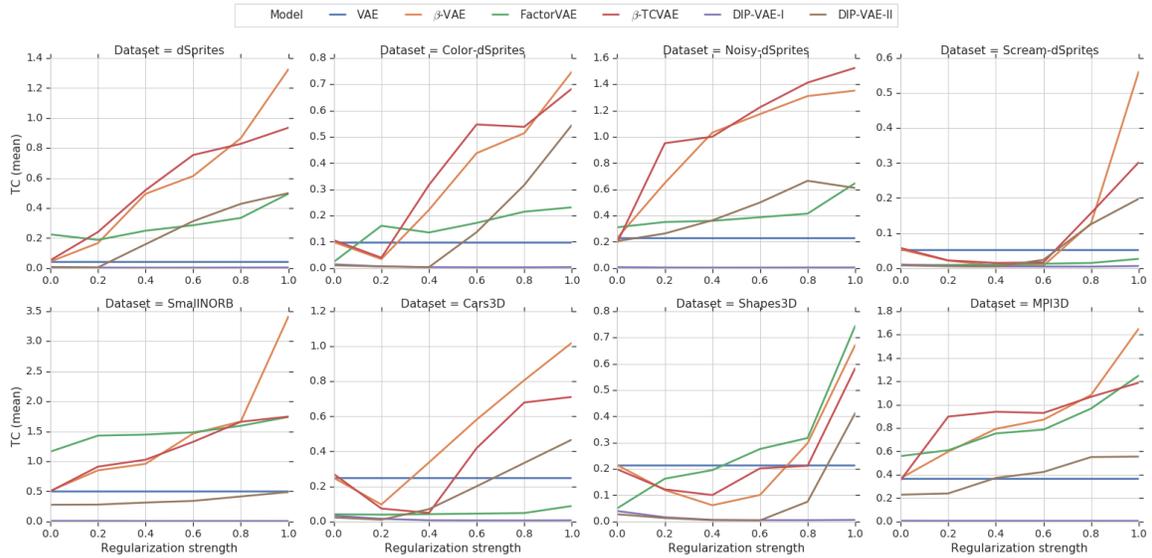}
\caption{Total correlation of mean representation plotted against regularization strength for different data sets and approaches (except AnnealedVAE). The total correlation of the mean representation does not necessarily decrease as the regularization strength is increased.}\label{figure:TCmean}
\end{figure}
\begin{figure}[pt]
\centering\includegraphics[width=\textwidth]{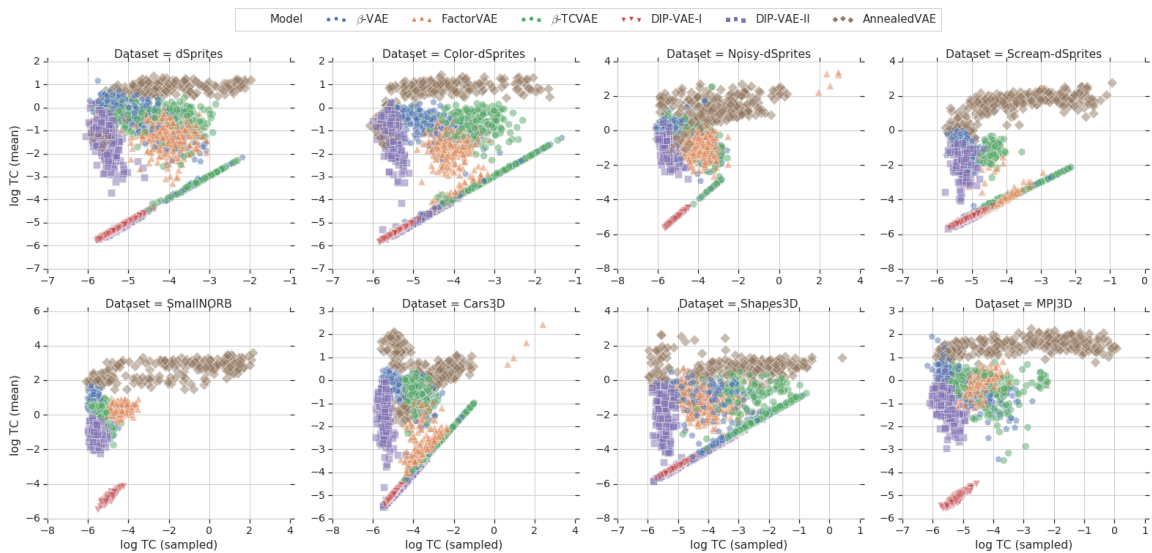}
\caption{Log total correlation of mean vs sampled representations. For a large number of models, the total correlation of the mean representation is higher than that of the sampled representation.}\label{figure:TCmeansampled}
\end{figure}

\subsection{Can Current Methods Enforce a Uncorrelated Aggregated Posterior and Representation?}\label{app:factorizing_q_tc}
We investigate whether the considered unsupervised disentanglement approaches are effective at enforcing a factorizing and thus uncorrelated aggregated posterior.
For each trained model, we sample $\num{10000}$ images and compute a sample from the corresponding approximate posterior. 
We then fit a multivariate Gaussian distribution over these \num{10000} samples by computing the empirical mean and covariance matrix.
Finally, we compute the total correlation of the fitted Gaussian and report the median value for each data set, method and hyperparameter value.

Figure~\ref{figure:TCsampled} shows the total correlation of the sampled representation plotted against the regularization strength for each data set and method except AnnealedVAE.
On all data sets except SmallNORB, we observe that plain vanilla variational autoencoders (the $\beta$-VAE model with $\beta=1$) exhibit the highest total correlation.
For $\beta$-VAE and $\beta$-TCVAE, it can be clearly seen that the total correlation of the sampled representation decreases on all data sets as the regularization strength (in the form of $\beta$) is increased.
The two variants of DIP-VAE exhibit low total correlation across the data sets except DIP-VAE-I which incurs a slightly higher total correlation on SmallNORB compared to a vanilla VAE.
Increased regularization in the DIP-VAE objective also seems to lead a reduced total correlation, even if the effect is not as pronounced as for $\beta$-VAE and $\beta$-TCVAE.
While FactorVAE achieves a low total correlation on all data sets except on SmallNORB, we observe that the total correlation does not seem to decrease with increasing regularization strength. 
We further observe that AnnealedVAE (shown in Figure~\ref{figure:TCsampledApp} in the Appendix) is much more sensitive to the regularization strength.
However, on all data sets except Scream-dSprites (on which AnnealedVAE performs poorly), the total correlation seems to decrease with increased regularization strength.

While many of the considered methods aim to enforce a factorizing aggregated posterior, they use the mean vector of the Gaussian encoder as the representation and not a sample from the Gaussian encoder.
This may seem like a minor, irrelevant modification; however, it is not clear whether a factorizing aggregated posterior also ensures that the dimensions of the mean representation are uncorrelated.
To test whether this is true, we compute the mean of the Gaussian encoder for the same \num{10000} samples, fit a multivariate Gaussian and compute the total correlation of that fitted Gaussian.
Figure~\ref{figure:TCmean} shows the total correlation of the mean representation plotted against the regularization strength for each data set and method except AnnealedVAE.
We observe that, for $\beta$-VAE and $\beta$-TCVAE, increased regularization leads to a substantially increased total correlation of the mean representations.
This effect can also be observed for for FactorVAE, albeit in a less extreme fashion.
For DIP-VAE-I, we observe that the total correlation of the mean representation is consistently low.
This is not surprising as the DIP-VAE-I objective directly optimizes the covariance matrix of the mean representation to be diagonal which implies that the corresponding total correlation (as we compute it) is low.
The DIP-VAE-II objective which enforces the covariance matrix of the sampled representation to be diagonal seems to lead to a factorized mean representation on some data sets (for example Shapes3D), but also seems to fail on others (dSprites, MPI3D).
For AnnealedVAE (shown in Figure~\ref{figure:TCmeanApp} in the Appendix), we overall observe mean representations with a very high total correlation.
In Figure~\ref{figure:TCmeansampled}, we further plot the log total correlations of the sampled representations versus the mean representations for each of the trained models.
It can be clearly seen that for a large number of models, the total correlation of the mean representations is much higher than that of the sampled representations. The same trend can be seen computing the average discrete mutual information of the representation. In this case, the DIP-VAE-I exhibit increasing mutual information in both the mean and sampled representation. This is to be expected as DIP-VAE-I enforces a variance of one for the mean representation. We remark that as the regularization terms and hyperparameter values are different for different losses, one should not draw conclusions from comparing different models at nominally the same regularization strength. From these plots one can only compare the effect of increasing the regularization in the different models.

\subsubsection{Implications}
Overall, these results lead us to conclude with minor exceptions that the considered methods are effective at enforcing an aggregated posterior whose individual dimensions are not correlated but that this does not seem to imply that the dimensions of the mean representation (usually used for representation) are uncorrelated.

\begin{figure}[p]
\centering\includegraphics[width=\textwidth]{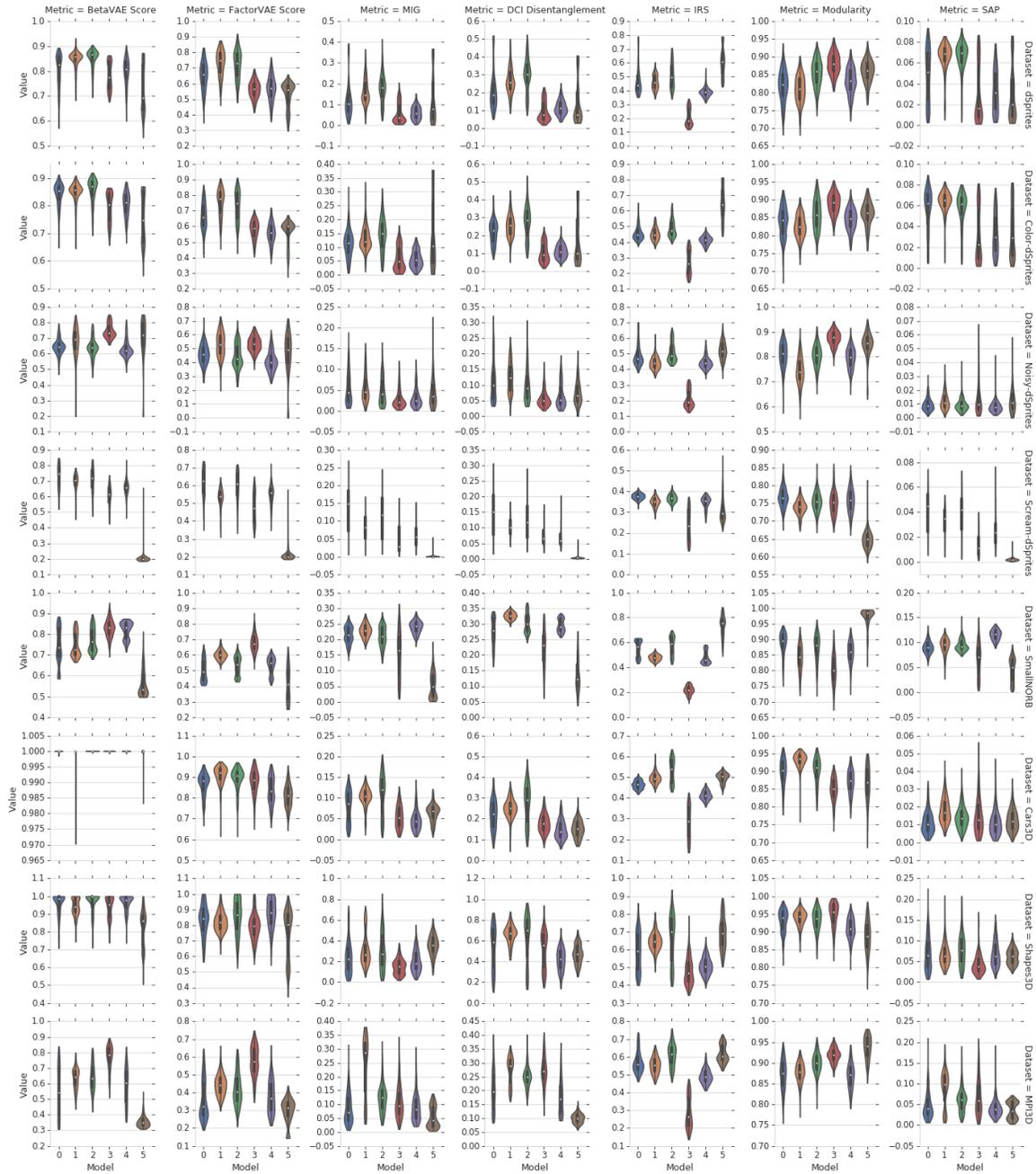}
\caption{Score for each method for each score (column) and data set (row) with different hyperparameters and random seed. Models are abbreviated (0=$\beta$-VAE, 1=FactorVAE, 2=$\beta$-TCVAE, 3=DIP-VAE-I, 4=DIP-VAE-II, 5=AnnealedVAE). The scores are heavily overlapping and we do not observe a consistent pattern. We conclude that hyperparameters and random seed matter more than the model choice.}\label{figure:score_vs_method}
\end{figure}
\begin{figure}[p]
\centering\includegraphics[width=\textwidth]{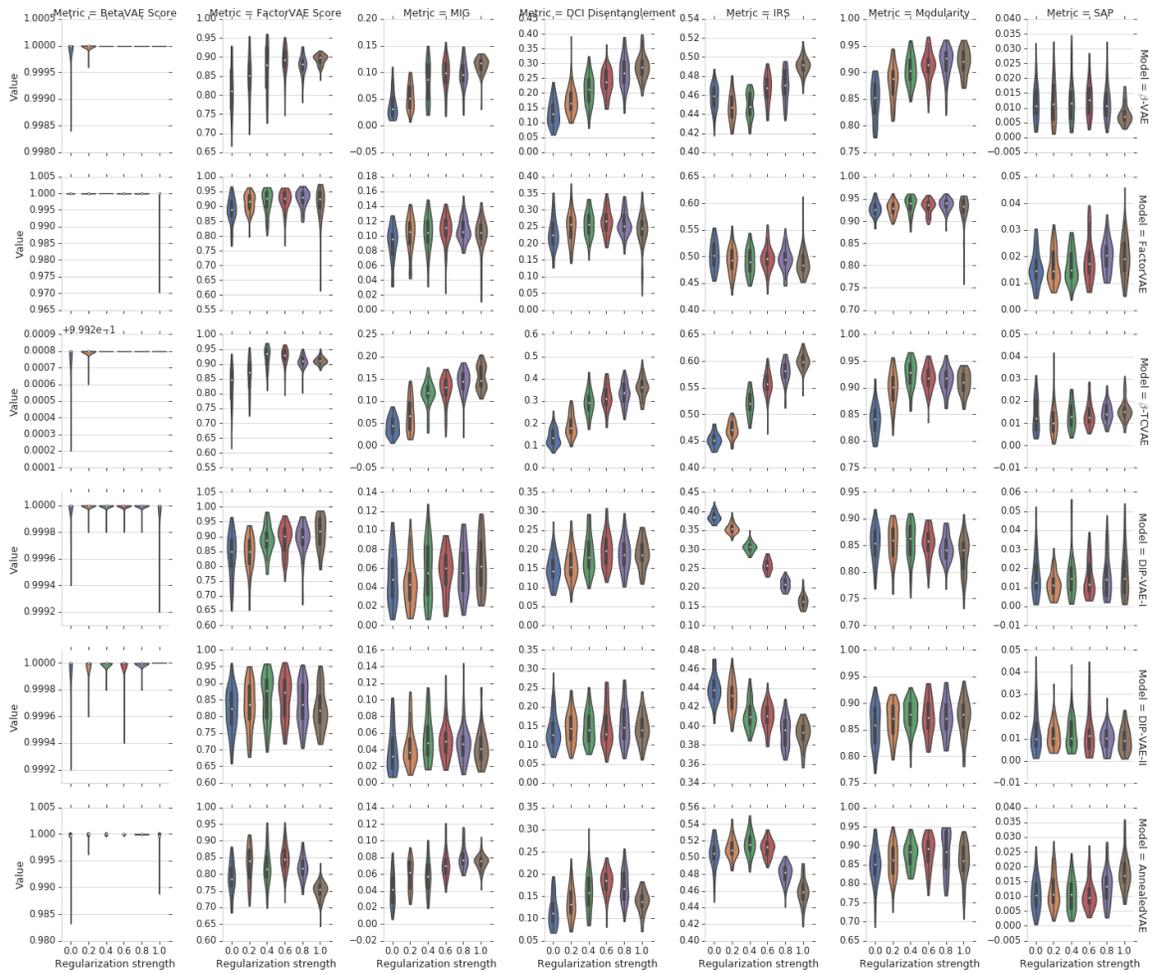}
\caption{Distribution of scores for different models, hyperparameters and regularization strengths on Cars3D. We clearly see that randomness (in the form of different random seeds) has a substantial impact on the attained result and that a good run with a bad hyperparameter can beat a bad run with a good hyperparameter in many cases. IRS seem to be an exception on some data sets.}\label{figure:random_seed_effect_Cars3D}
\end{figure}

\subsection{How Important Are Different Models and Hyperparameters for Disentanglement?}\label{app:hyper_importance}
The primary motivation behind the considered methods is that they should lead to improved disentanglement scores.
This raises the question how disentanglement is affected by the model choice, the hyperparameter selection and randomness (in the form of different random seeds).
To investigate this, we compute all the considered disentanglement metrics for each of our trained models.
In Figure~\ref{figure:score_vs_method}, we show the range of attainable disentanglement scores for each method on each data set varying the regularization strenght and the random seed.
We observe that these ranges are heavily overlapping for different models leading us to (qualitatively) conclude that the choice of hyperparameters and the random seed seems to be substantially more important than the choice of objective function. 
While certain models seem to attain better maximum scores on specific data sets and disentanglement metrics, we do not observe any consistent pattern that one model is consistently better than the other. DIP-VAE-I consistently gets lower IRS score, but is comparable to the other methods with all the other scores.
Furthermore, we note that in our study we have fixed the range of hyperparameters a priori to six different values for each model and did not explore additional hyperparameters based on the results (as that would bias our study).
However, this also means that specific models may have performed better than in Figure~\ref{figure:score_vs_method} if we had chosen a different set of hyperparameters.

In Figure~\ref{figure:random_seed_effect_Cars3D}, we further show the impact of randomness in the form of random seeds on the disentanglement scores.
Each violin plot shows the distribution of the disentanglement metric across all 50 trained models for each model and hyperparameter setting on Cars3D.
We clearly see that randomness (in the form of different random seeds) has a substantial impact on the attained result and that a good run with a bad hyperparameter can beat a bad run with a good hyperparameter in many cases. We note that IRS seem to exhibit a clear trend on some data sets.

Finally, we perform a variance analysis by trying to predict the different disentanglement scores using ordinary least squares for each data set:
If we allow the score to depend only on the objective function (categorical variable), we are only able to explain $37\%$ of the variance of the scores on average. 
Similarly, if the score depends on the Cartesian product of objective function and regularization strength (again categorical), we are able to explain $59\%$ of the variance while the rest is due to the random seed. In Table~\ref{table:variance_explained} in the Appendix, we report the percentage of variance explained for the different metrics in each data set considering the regularization strength or not. 

\subsubsection{Implications} 
The disentanglement scores of unsupervised models are heavily influenced by randomness (in the form of the random seed) and the choice of the hyperparameter (in the form of the regularization strength). The objective function appears to have less impact.

\subsection{Are There Reliable Recipes for Model Selection?}\label{app:hyper_selection}
In this section, we investigate how good hyperparameters can be chosen and how we can distinguish between good and bad training runs. 
In this paper, we advocate that model selection \emph{should not} depend on the considered disentanglement score for the following reasons:
The point of unsupervised learning of disentangled representation is that there is no access to the labels as otherwise we could incorporate them and would have to compare to semi-supervised and fully supervised methods.
All the disentanglement metrics considered in this paper require a substantial amount of ground-truth labels or the full generative model (for example for the BetaVAE and the FactorVAE metric).
Hence, one may substantially bias the results of a study by tuning hyperparameters based on (supervised) disentanglement metrics. 
Furthermore, we argue that it is not sufficient to fix a set of hyperparameters \emph{a priori} and then show that one of those hyperparameters and a specific random seed achieves a good disentanglement score as it amounts to showing the existence of a good model, but does not guide the practitioner in finding it.
Finally, in many practical settings, we might not even have access to adequate labels as it may be hard to identify the true underlying factor of variations, in particular, if we consider data modalities that are less suitable to human interpretation than images.

In the remainder of this section, we hence investigate and assess different ways how hyperparameters and good model runs could be chosen.
In this study, we focus on choosing the learning model and the regularization strength corresponding to that loss function.
However, we note that in practice this problem is likely even harder as a practitioner might also want to tune other modeling choices such architecture or optimizer.
 
\subsubsection{General Recipes for Hyperparameter Selection}
\begin{figure}[p]
\centering\includegraphics[width=\textwidth]{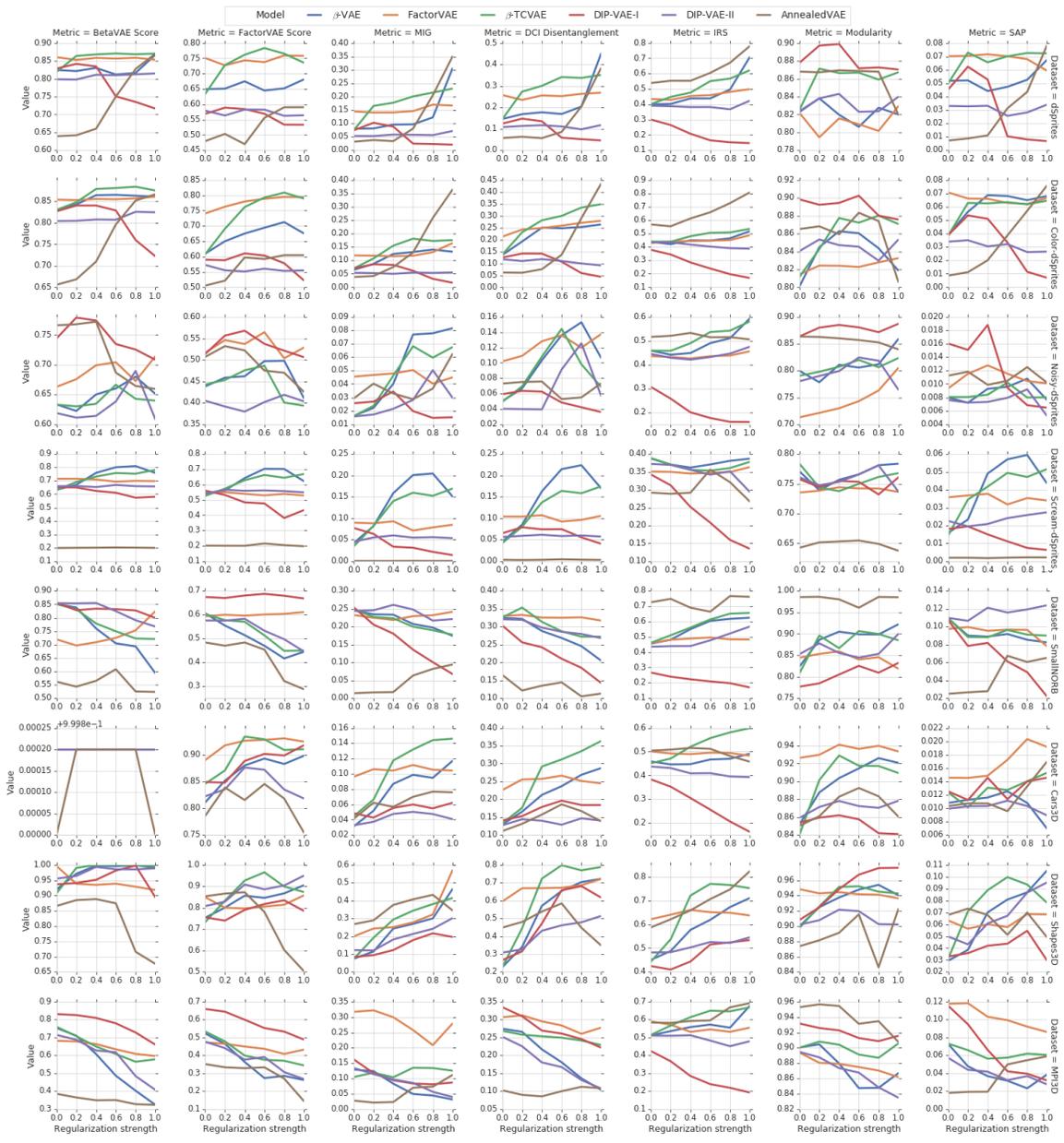}
\caption{Score vs hyperparameters for each score (column) and data set (row). There seems to be no model dominating all the others and for each model there does not seem to be a consistent strategy in choosing the regularization strength.}\label{figure:score_vs_hyp}
\end{figure}

We first investigate whether we may find generally applicable ``rules of thumb'' for choosing the hyperparameters.
For this, we plot in Figure~\ref{figure:score_vs_hyp} different disentanglement metrics against different regularization strengths for each model and each data set.
The values correspond to the median obtained values across 50 random seeds for each model, hyperparameter and data set.
There seems to be no model dominating all the others and for each model there does not seem to be a consistent strategy in choosing the regularization strength to maximize disentanglement scores.
Furthermore, even if we could identify a good objective function and corresponding hyperparameter value, we still could not distinguish between a good and a bad training run.

\subsubsection{Model Selection Based on Unsupervised Scores}
\begin{figure}[p]
\centering\includegraphics[width=\textwidth]{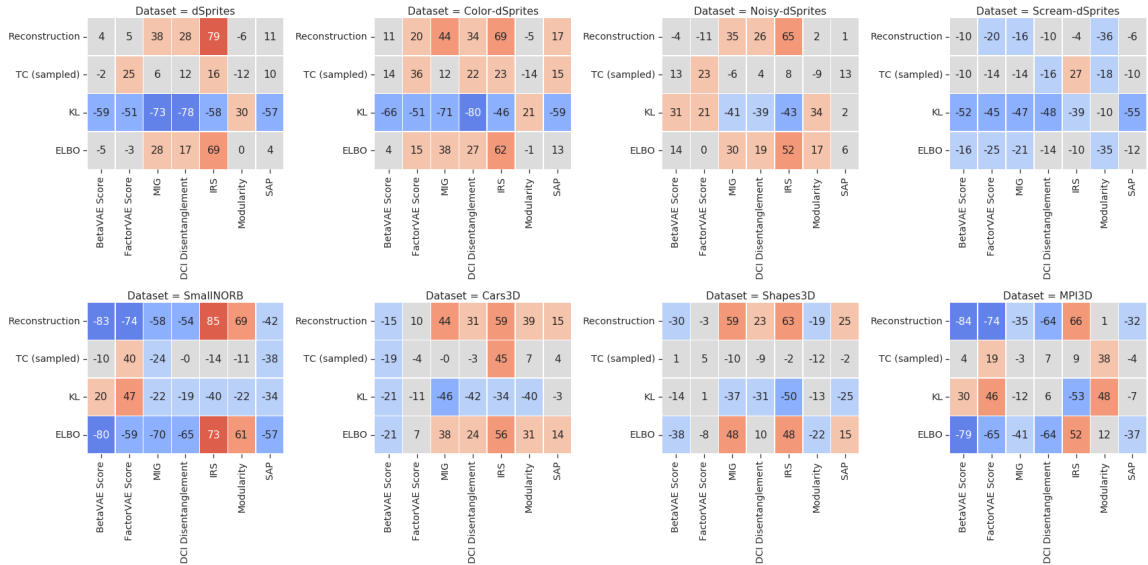}
\caption{Rank correlation between unsupervised scores and supervised disentanglement metrics. 
The unsupervised scores we consider do not seem to be useful for model selection.}\label{figure:unsupervised_metrics}
\end{figure}

Another approach could be to select hyperparameters based on unsupervised scores such as the reconstruction error, the KL divergence between the prior and the approximate posterior, the Evidence Lower Bound or the estimated total correlation of the sampled representation.
This would have the advantage that we could select specific trained models and not just good hyperparameter settings whose median trained model would perform well.
To test whether such an approach is fruitful, we compute the rank correlation between these unsupervised metrics and the disentanglement metrics and present it in Figure~\ref{figure:unsupervised_metrics}.
While we do observe some correlations, no clear pattern emerges which leads us to conclude that this approach is unlikely to be successful in practice.
\begin{table}[t]
    \centering
    \vspace{2mm}
\begin{tabular}{lrr}
\toprule
{} &  Random different data set &  Same data set \\
\midrule
Random different metric &                      52.7\% &          62.1\% \\
Same metric             &                      59.6\% &          81.9\% \\
\bottomrule
\end{tabular}
\caption{Probability of outperforming random model selection on a different random seed. A random disentanglement metric and data set is sampled and used for model selection. That model is then compared to a randomly selected model: (i) on the same metric and data set, (ii) on the same metric and a random different data set, (iii) on a random different metric and the same data set, and (iv) on a random different metric and a random different data set. The results are averaged across $\num{10000}$ random draws.}\label{table:app_random_model}
\end{table}

\begin{figure}[p]
\centering\includegraphics[width=\textwidth]{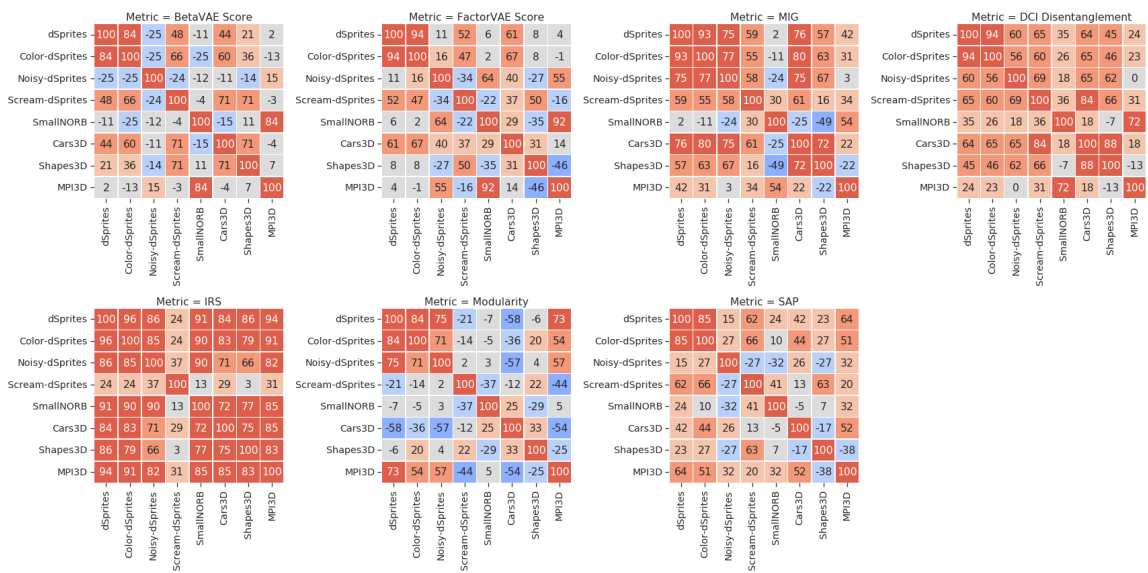}
\caption{Rank-correlation of different disentanglement metrics across different data sets. Good hyperparameters only seem to transfer between dSprites and Color-dSprites but not in between the other data sets.}\label{figure:dataset_vs_dataset}
\end{figure}

\begin{figure}[pt]
\centering\includegraphics[width=\textwidth]{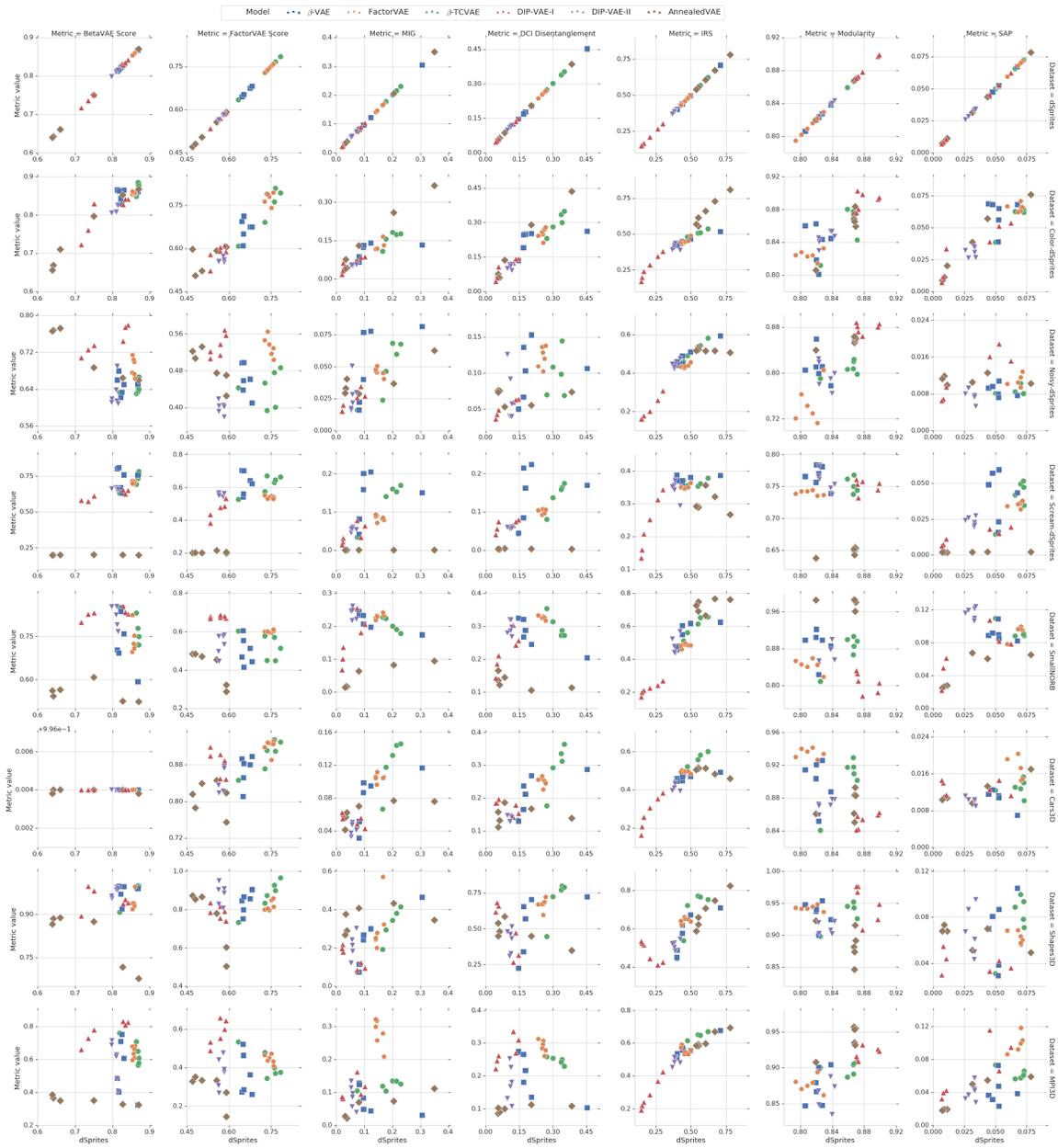}
\caption{Disentanglement scores on dSprites vs other data sets. Good hyperparameters only seem to transfer consistently from dSprites to Color-dSprites.}\label{figure:TransferPairs}
\end{figure}

\subsubsection{Hyperparameter Selection Based on Transfer}
The final strategy for hyperparameter selection that we consider is based on transferring good settings across data sets.
The key idea is that good hyperparameter settings may be inferred on data sets where we have labels available (such as dSprites) and then applied to novel data sets.
To test this idea, we plot in Figure~\ref{figure:TransferPairs} the different disentanglement scores obtained on dSprites against the scores obtained on other data sets.
To ensure robustness of the results, we again consider the median across all 50 runs for each model, regularization strength, and data set.
We observe that the scores on Color-dSprites seem to be strongly correlated with the scores obtained on the regular version of dSprites.
Figure~\ref{figure:dataset_vs_dataset} further shows the rank correlations obtained between different data sets for each disentanglement scores.
This confirms the strong and consistent correlation between dSprites and Color-dSprites. 
While these result suggest that some transfer of hyperparameters is possible, it does not allow us to distinguish between good and bad random seeds on the target data set.

To illustrate this, we compare such a transfer based approach to hyperparameter selection to random model selection as follows: 
We first randomly sample one of our 50 random seeds and consider the set of trained models with that random seed. 
First, we sample one of our 50 random seeds, a random  disentanglement metric and a data set and use them to select the hyperparameter setting with the highest attained score.
Then, we compare that selected hyperparameter setting to a randomly selected model on either the same or a random different data set, based on either the same or a random different metric and for a randomly sampled seed.
Finally, we report the percentage of trials in which this transfer strategy outperforms or performs equally well as random model selection across $\num{10000}$ trials in Table~\ref{table:app_random_model}.
If we choose the same metric and the same data set (but a different random seed), we obtain a score of $81.9\%$.
If we aim to transfer for the same metric across data sets, we achieve around $59.6\%$.
Finally, if we transfer both across metrics and data sets, our performance drops to $52.7\%$. The drop in performance transferring hyperparameters across different metrics may be interpreted in light of the results of Section~\ref{sec:eval_same_concept}.

\subsubsection{Implications} 
Unsupervised model selection remains an unsolved problem. 
Transfer of good hyperparameters between metrics and data sets does not seem to work as there appears to be no unsupervised way to distinguish between good and bad random seeds on the target task. Recent work~\citep{duan2019heuristic} may be used to select stable hyperparameter configurations. The IRS score seem to be more correlated with the unsupervised training metrics on most data set and generally transfer the hyperparameters better. However, as we shall see in Section~\ref{sec:metrics_agreement}, IRS is not very correlated with the other disentanglement metrics.

\section{What Are the Differences Between the Disentanglement Metrics?}\label{sec:evaluating}

The disentanglement of a learned representation can be seen as a certain structural property of the statistical relations between the latent space of the VAE with that of the ground truth factors. Therefore, when evaluating disentangled representations several metrics typically estimate these statistical dependencies first and then compute how well this structure encodes the desired properties. 
As quantifying statistical dependencies through independence testing is a challenging task~\citep{shah2018hardness} several approaches have been proposed.
We identify two prevalent settings: using interventional~\citep{higgins2016beta,kim2018disentangling,suter2018interventional} and observational data~\citep{chen2018isolating,ridgeway2018learning,eastwood2018framework}. 

\begin{figure}
    \begin{center}
    \begin{subfigure}{0.4\textwidth}
    \includegraphics[width=\textwidth]{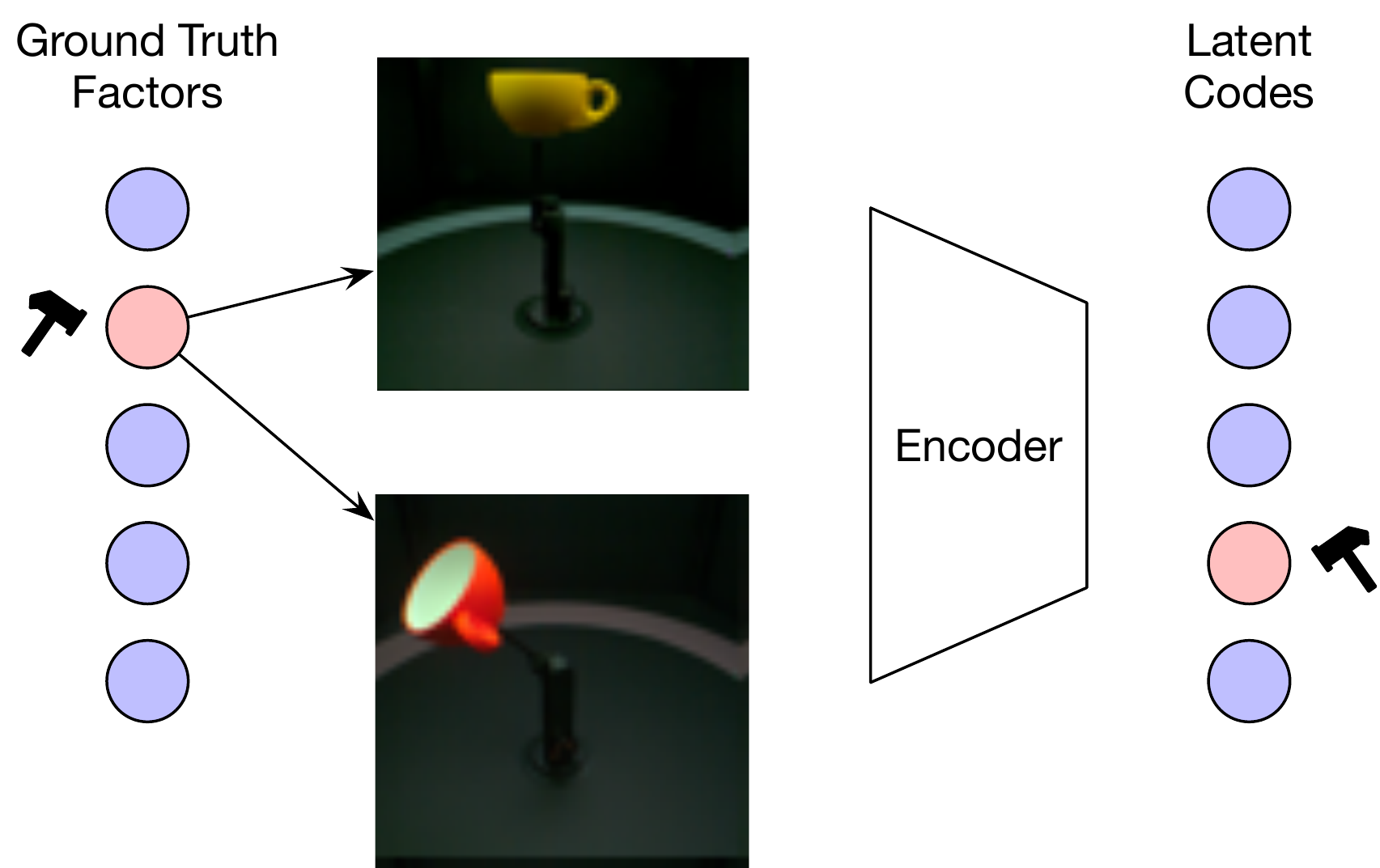}
    \caption{Example of encoder \emph{consistency}~\citep{shu2019weakly} for one factor of variation: intervening (\myhammer{}) on a ground-truth factor (or subset of factors) by \emph{fixing} its value corresponds to \emph{fixing} a dimension (or subset of dimensions) in the representation. In this example the object shape is constant and everything else is changing.}\label{fig:disentanaglement_notions_int_fix}
    \end{subfigure}%
    \hspace{4em}
    \begin{subfigure}{0.4\textwidth}
    \includegraphics[width=\textwidth]{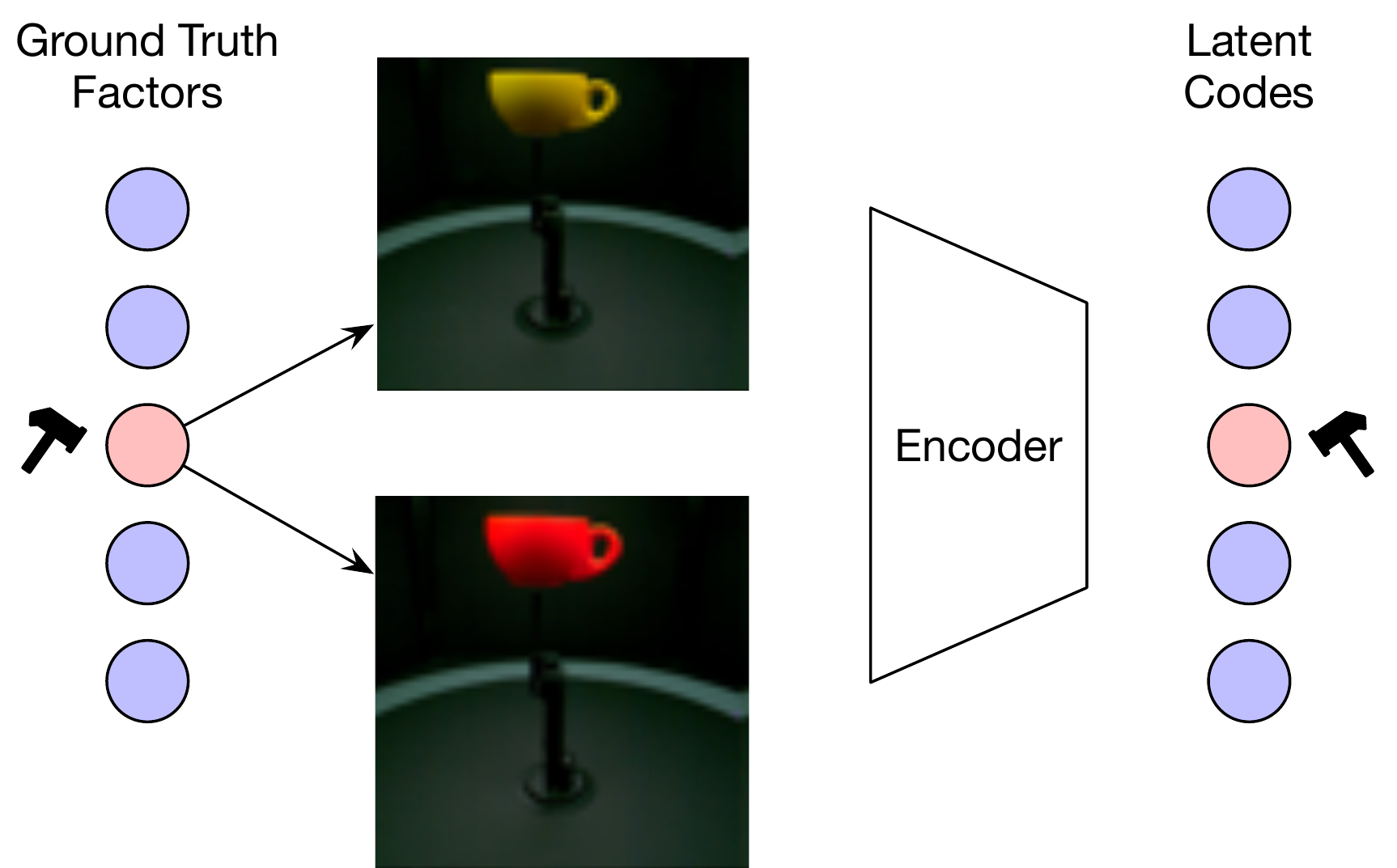}
    \caption{Example of encoder \emph{restrictiveness}~\citep{shu2019weakly}  for one factor of variation: intervening (\myhammer{}) on a ground-truth factor (or subset of factors) by \emph{changing} its value corresponds to \emph{changing} a dimension (or subset of dimensions) in the representation. In this example only the color is changing.}\label{fig:disentanaglement_notions_int_change}
    \end{subfigure}
    \vspace{2em}
    \begin{subfigure}{0.4\textwidth}
    \includegraphics[width=\textwidth]{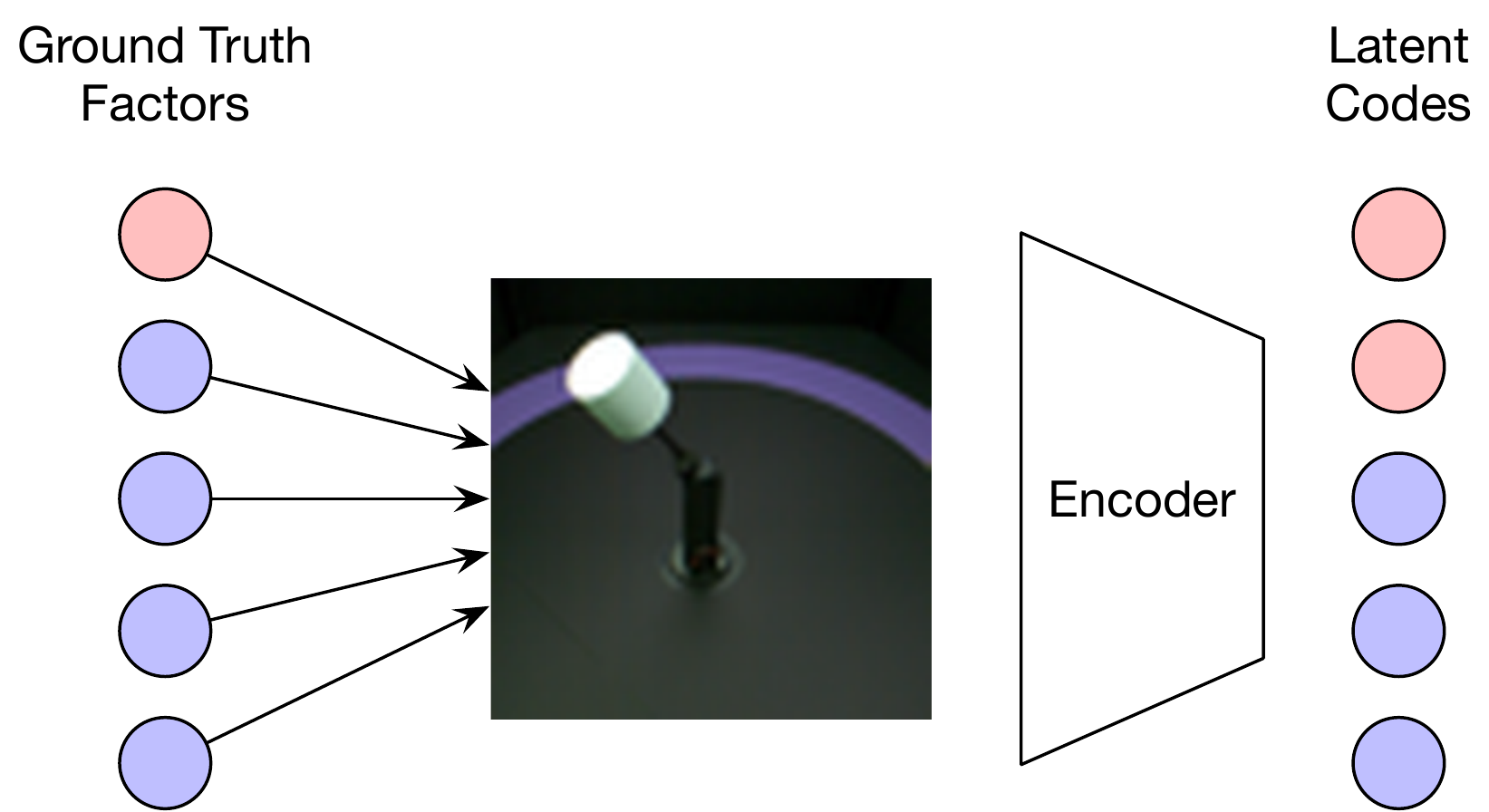}
    \caption{Example of \emph{disentangled} encoder for one factor of variation in the sense of~\citep{eastwood2018framework}: a few dimensions are capturing a single factor. }\label{fig:disentanaglement_notions_disent}
    \end{subfigure}%
    \hspace{4em}
    \begin{subfigure}{0.4\textwidth}
    \vspace{2em}
    \includegraphics[width=\textwidth]{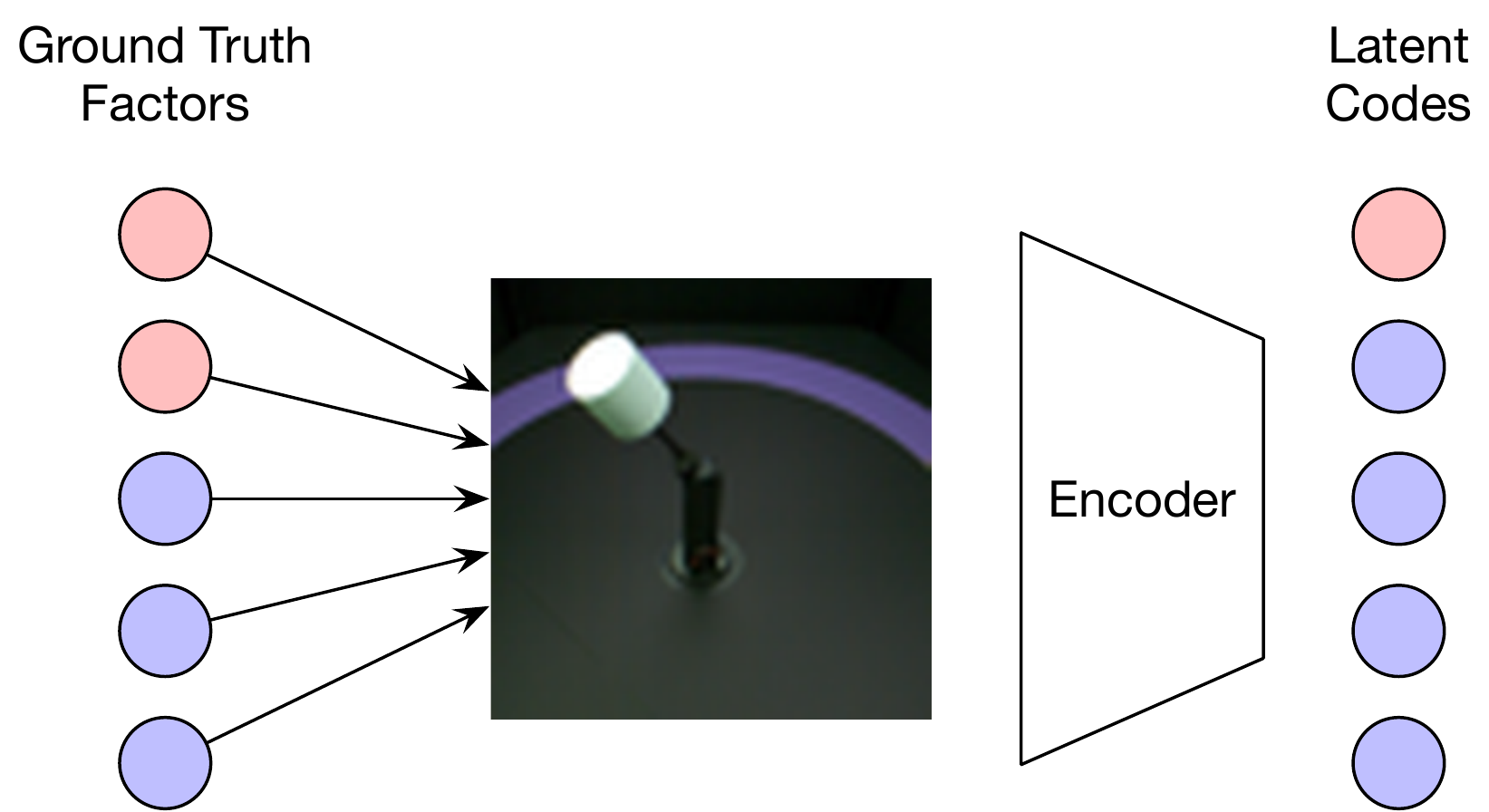}
    \caption{Example of encoder \emph{compactness} for one factor of variation in the sense of~\citep{eastwood2018framework}: a factor of variation should be captured in a single dimension. However multiple factors can still be encoded in the same dimension.}\label{fig:disentanaglement_notions_compact}
    \end{subfigure}
    \end{center}
    \caption{Examples of different notions of disentanglement being captured by the scores. Further, different scores \emph{measure} the same notion in different ways, which can introduce systematic differences in the evaluation. For the encoder to be consistent (a), restrictive (b), disentangled (c), or compact (d) the property highlighted in the each example should hold for each factor.}
    \label{fig:disentanaglement_notions}
\end{figure}

For interventional data, the two main properties a disentangled representation should have are \emph{consistency} and \emph{restrictiveness}~\citep{shu2019weakly}. Examples can be seen in Figures~\ref{fig:disentanaglement_notions_int_fix}~and~\ref{fig:disentanaglement_notions_int_change}. Both can be interpreted in the context of independent mechanisms~\citep{peters2017elements}: interventions on a ground-truth factor should manifest in a localized way in the representation. For example, fixing a certain factor of variation and sampling twice all others should result in a subset of dimensions being constant in the representation of the two points (consistency). This notion is used in the metrics of~\citet{higgins2016beta,kim2018disentangling}. On the other hand, changing the value of a factor of variation while keeping the others constant should result in a single change in the representation. This fact was used in the evaluation metric proposed by~\citet{suter2018interventional}. While~\citep{shu2019weakly} argue that both aspects are necessary for disentangled representations, when the ground-truth factors are independent and unconfounded the two definitions are equivalent.

On the observational data, which is arguably the most practical case, there are several ways of estimating the relationship between factors and codes. For example, \citet{chen2018isolating,ridgeway2018learning} use the mutual information while \citet{eastwood2018framework,kumar2017variational} rely on predictability with a random forest classifier and a SVM respectively. The practical impact of these low-level and seemingly minor differences is not yet understood.

Once the relation between the factors and the codes is known for a given model, we need to evaluate the properties of the structure in order to measure its ``disentanglement''. Since a generally accepted formal definition for disentanglement is missing~\citep{eastwood2018framework, higgins2018towards,ridgeway2018learning}, the desired structure of the latent space compared to the ground truth factors is a topic of debate. 
~\citet{eastwood2018framework} (and in part~\citet{ridgeway2018learning}) proposed three properties of representations: \emph{disentanglement}, \emph{compactness}, and \emph{informativeness}. A representation is disentangled if each dimension only captures a single factor of variation and compact if each factor is encoded in a single dimension, see Figures~\ref{fig:disentanaglement_notions_disent}~and~\ref{fig:disentanaglement_notions_compact}. Note that disentangled representations do not need to be compact nor compact representations need to be disentangled. Combining the two implies that a representation implements a one-to-one mapping between factors of variation and latent codes. Informativeness measures how well the information about the factors of variation is accessible in the latent representations with linear models. The degree of informativeness captured by any of the disentanglement metrics is unclear. In particular, as discussed in Section~\ref{sec:downstream}, it is not clear whether the correlation between disentanglement metrics and downstream performance is an artifact of the linear model used to estimate the relations between factors and code~\citep{eastwood2018framework,kumar2017variational}. 
Maintaining the terminology, the disentanglement scores in~\citep{higgins2016beta,kim2018disentangling,ridgeway2018learning,eastwood2018framework,suter2018interventional} focus on disentanglement in the sense of~\citep{eastwood2018framework} and~\citep{chen2018isolating,kumar2017variational} on compactness. Note that all these scores implement their own ``notion of disentanglement''.
Theoretically, we can characterize existing metrics in these two groups. On the other hand, observing the latent traversal of top performing models, it is not clear what the differences between the scores are and whether compactness and disentanglement are essentially equivalent on representations learned by VAEs (a compact representation is also disentangled and vice-versa).

As a motivating example for this section consider the two models in Figure~\ref{figure:counterexample_traversals}. While visually we may say that they are similarly disentangled, they achieve significantly different MIG scores, making the first model twice as good as the second one. Artifacts like this clearly impact the conclusions one may draw from a quantitative evaluation. Further, the structure of the representation may influence its usefulness downstream, and different properties may be useful for different tasks. For example, the applications in fairness~\citep{locatello2019fairness}, abstract reasoning~\citep{van2019disentangled} and strong generalization~\citep{locatello2020weakly} all conceptually rely on the disentanglement notion of~\citep{eastwood2018framework}.

In this section, we first question how much the metrics agree with each other in terms of how the models are ranked. Second, we focus on the metrics that can be estimated from observational data, as we anticipate they will be more generally applicable in practice. There, we question the impact of different choices in the estimation of the factor-code matrices as well as in the aggregation. This latter step encodes which notion of disentanglement is measured.
Finally, we investigate the sample efficiency of the different metrics in order to provide practical insights on which scores may be used in practical settings where labelled data is scarce.

\begin{figure}[p]
\centering\includegraphics[width=\textwidth]{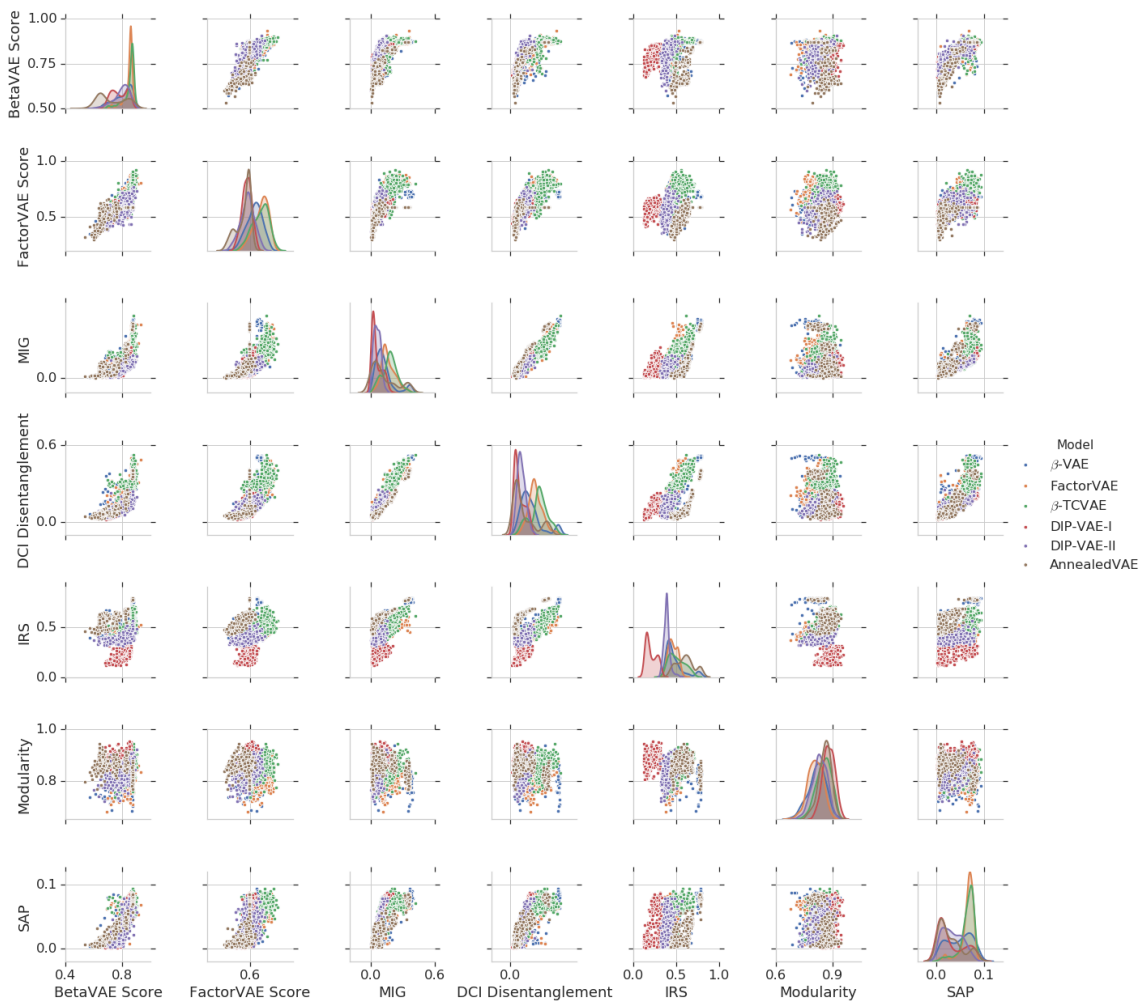}
\caption{Pairwise scatter plots of different disentanglement metrics on dSprites. All the metrics except Modularity appear to be correlated. The strongest correlation seems to be between MIG and DCI Disentanglement.}\label{figure:dsprites_metrics_scatter}
\end{figure}

\subsection{How Much Do Existing Disentanglement Metrics Agree?}
\label{sec:metrics_agreement}
\begin{figure}[pt]
\centering\includegraphics[width=\textwidth]{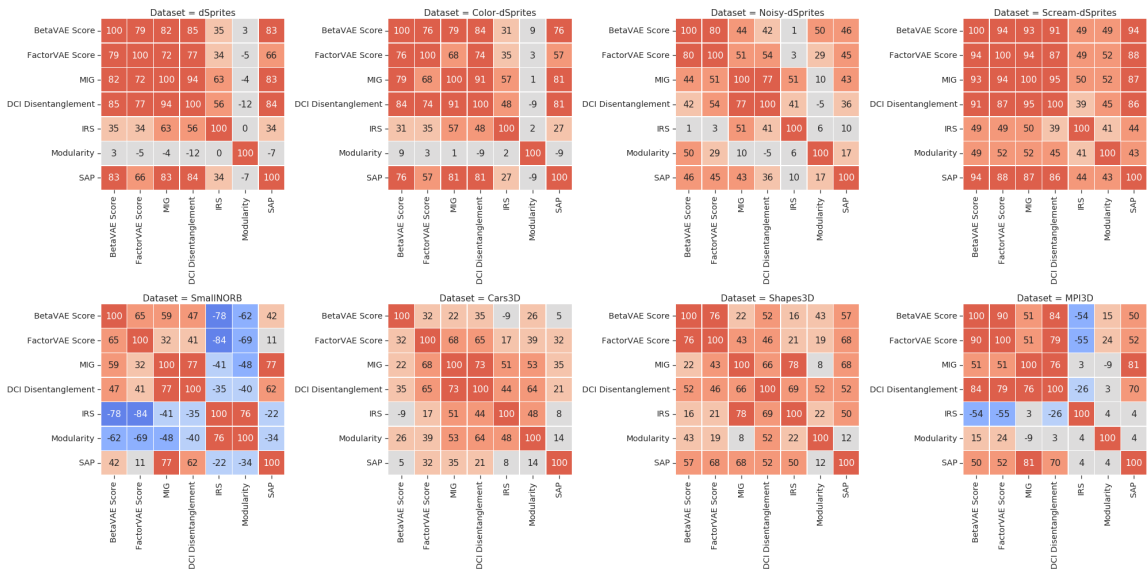}
\caption{Rank correlation of different metrics on different data sets. 
Overall, we observe that all metrics except Modularity seem to be strongly correlated on the data sets dSprites, Color-dSprites and Scream-dSprites and mildly on the other data sets.
 There appear to be two pairs among these metrics that capture particularly similar notions: the BetaVAE and the FactorVAE score as well as the Mutual Information Gap and DCI Disentanglement.
}\label{figure:metrics_rank_correlation}
\end{figure}

As there exists no single, commonly accepted definition of disentanglement, an interesting question is to see how much the different metrics agree.
Figure~\ref{figure:dsprites_metrics_scatter} shows pairwise scatter plots of the different considered metrics on dSprites where each point corresponds to a trained model, while Figure~\ref{figure:metrics_rank_correlation} shows the Spearman rank correlation between different disentanglement metrics on different data sets.
Overall, we observe that all metrics except Modularity and, in part, IRS seem to be correlated strongly on the data sets dSprites, Color-dSprites and Scream-dSprites and mildly on the other data sets.
There appear to be two pairs among these metrics that correlate well: the BetaVAE and the FactorVAE scores as well as the Mutual Information Gap and DCI Disentanglement. Note that this positive correlation does not necessarily imply that these metrics are measuring the same notion of disentanglement.

\begin{figure}[pt]
\begin{center}
{\adjincludegraphics[scale=0.4, trim={0 {0.91\height} 0 0}, clip]{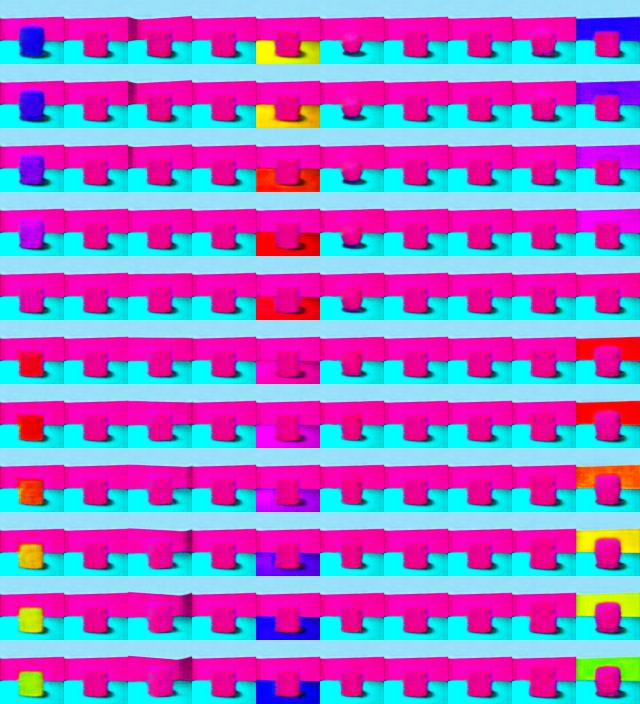}}\vspace{-0.3mm}
{\adjincludegraphics[scale=0.4, trim={0 {0.545\height} 0 {0.37\height}}, clip]{figures/counterexample.jpg}}\vspace{-0.3mm}
{\adjincludegraphics[scale=0.4, trim={0 0 0 {.91\height}}, clip]{figures/counterexample.jpg}}\vspace{2mm}
\end{center}
\begin{center}
{\adjincludegraphics[scale=0.4, trim={0 {0.91\height} 0 0}, clip]{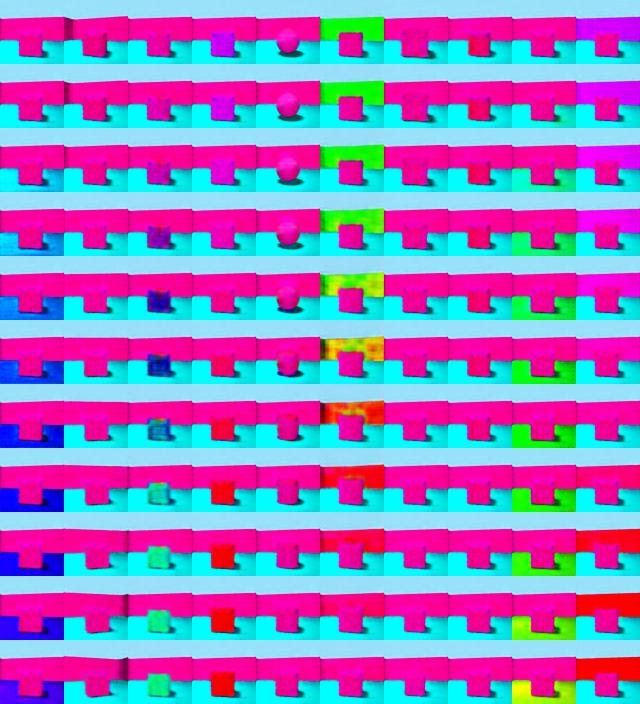}}\vspace{-0.3mm}
{\adjincludegraphics[scale=0.4, trim={0 {0.545\height} 0 {0.37\height}}, clip]{figures/traversals0.jpg}}\vspace{-0.3mm}
{\adjincludegraphics[scale=0.4, trim={0 0 0 {.91\height}}, clip]{figures/traversals0.jpg}}\vspace{2mm}
\end{center}
\caption{Latent traversal of a FactorVAE model (top) and a DIP-VAE-I (bottom) trained on Shapes3D. Despite dimensions 0, 5, and 8 not being perfectly disentangled (see Figure~\ref{figure:counterexample}), the model at the top achieves a MIG of 0.66 while the model at the bottom 0.33. Each column corresponds to a latent dimension. }\label{figure:counterexample_traversals}
\end{figure}

Indeed, we visualize in Figure~\ref{figure:counterexample_traversals} the latent traversals of two models that visually achieve similar disentanglement. Arguably, the model on the bottom may even be more disentangled that the one on the top (the shape in dimension 0 of the top model is not perfectly constant). However, the top model received a MIG of 0.66 while the model at the bottom just 0.33.  We remark that similar examples can be found for other disentanglement metrics as well by looking for models with large disagreement between the scores. The two models in Figure~\ref{figure:counterexample_traversals} have DCI Disentanglement of 0.77 and 0.94 respectively.

The scores that require interventions and measure disentanglement computing consistency versus restrictiveness are not strongly correlated although they should be theoretically equivalent. On the other hand, we notice that the IRS is not very correlated with the other scores either, indicating that the difference may arise from how the IRS is computed.

We now investigate the differences on the scores that are computed from purely observational data: DCI Disentanglement, MIG, Modularity and SAP Score. These scores are composed of two stages. First they estimate a matrix relating factors of variation and latent codes. DCI Disentanglement considers the feature importance of a GBT predicting each factor of variation from the latent codes. MIG and Modularity compute the pairwise mutual information matrix between factors and codes. The SAP Score computes the predictability of each factor of variation from each latent code using a SVM. Second, they aggregate this matrix into a score measuring some of its structural properties. This is typically implemented as a normalized gap between largest and second largest entries in the factor-code matrix either row or column wise. We argue that this second step is the one that most encodes the ``notion of disentanglement'' being measured by the score. However, the correlation between the scores may also be influenced by how the matrix is estimated. In the remainder of this section, we put under scrutiny these two steps, systematically analyzing their similarities, robustness, and biases.

\begin{figure}[pt]
\centering\includegraphics[width=\textwidth]{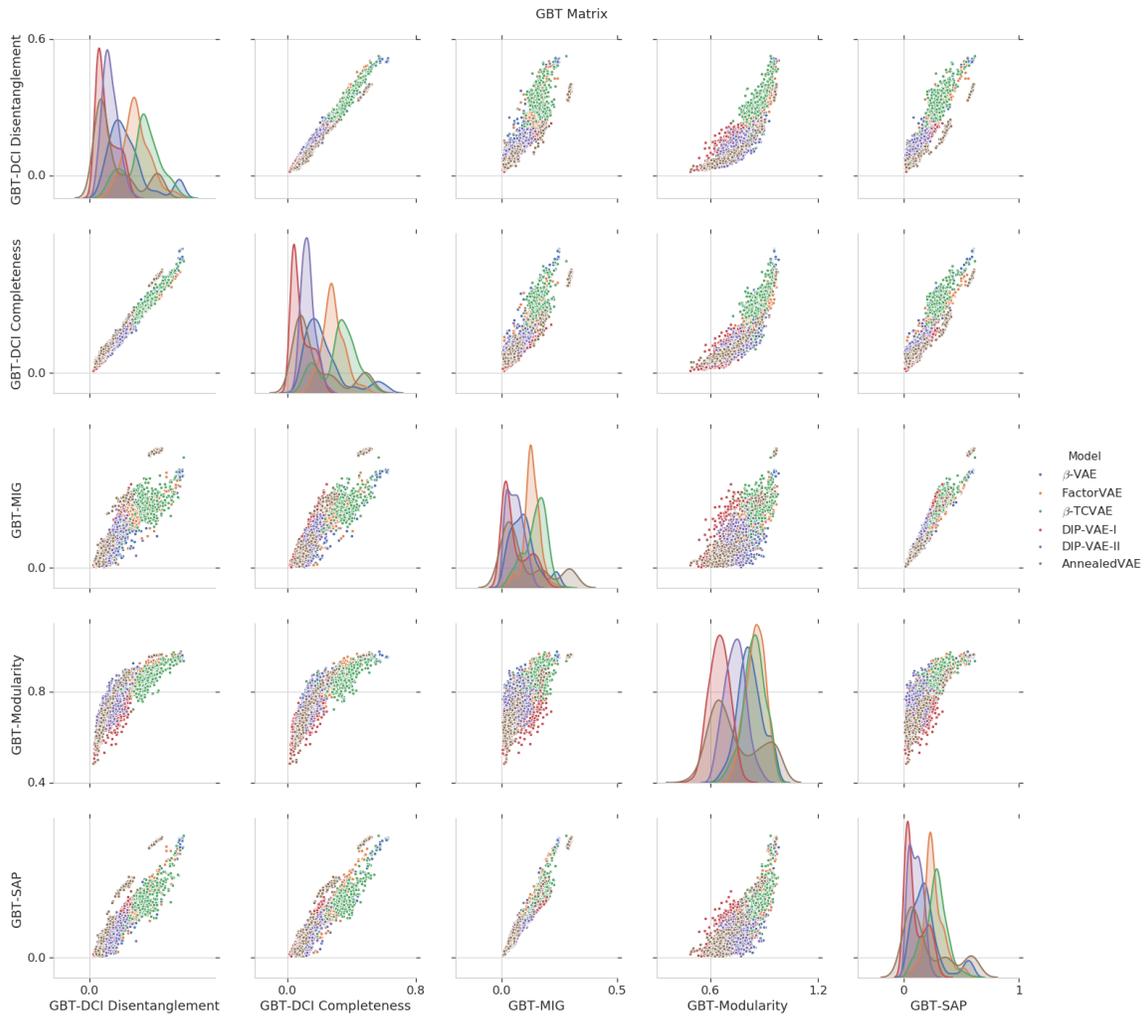}
\caption{Aggregations computed on the same matrix (GBT feature importance) correlate well on dSprites.
}\label{figure:dsprites_metrics_scattergbt_matrix}
\end{figure}
\begin{figure}[pt]
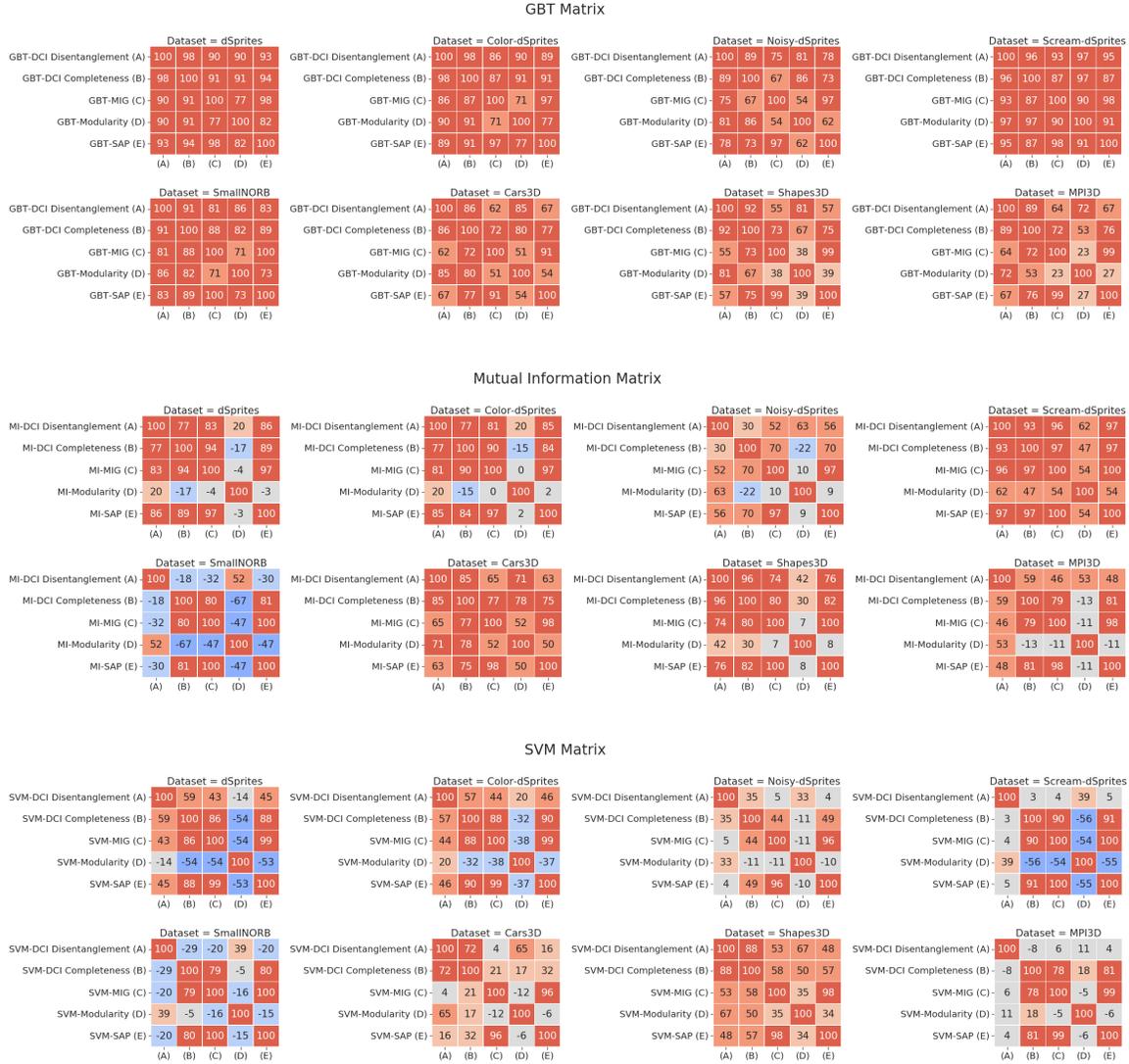

\centering\includegraphics[width=\textwidth]{autofigures/metrics_rank_gbt_matrix}\vspace{6mm}
\centering\includegraphics[width=\textwidth]{autofigures/metrics_rank_mi_matrix}\vspace{6mm}
\centering\includegraphics[width=\textwidth]{autofigures/metrics_rank_svm_matrix}\vspace{6mm}
\caption{Rank correlation of the different aggregations computed on the same matrix (GBT feature importance, mutual information, and predictability with a SVM). When the matrix is the same, MIG, SAP and DCI Completeness are significantly more correlated while the correlation with DCI Disentanglement decreases, highlighting the difference between completeness and disentanglement~\citep{eastwood2018framework}.
}\label{figure:metrics_rank_matrix}
\end{figure}

\subsubsection{What is the Difference Between the Aggregations? Is Compactness Equivalent to Disentanglement in Practice? }\label{sec:eval_same_concept}
In this section, we focus on the metrics that can be computed from observational data.
We question the ``notion of disentanglement'' which is implemented by the second step of DCI Disentanglement, MIG, Modularity and SAP Score and look for differences between disentanglement and compactness in practice. These aggregations measure some structural properties of the statistical relation between factors and codes. In order to empirically understand similarities and differences of these aggregations, we compare their result when evaluating the same input matrix in Figure~\ref{figure:dsprites_metrics_scattergbt_matrix} for dSprites and the GBT feature importance matrix. We observe that the different aggregations seem to correlate well but we note that this correlation is not always consistent across different matrices and data sets as can be seen in Figure~\ref{figure:metrics_rank_matrix}. We note that MIG, SAP and DCI Completeness are  always strongly correlated with each other when the matrix is the same. On the contrary, MIG/SAP and DCI Disentanglement are consistently less correlated on the same matrix. The correlation between Modularity and the other scores varies dramatically depending on the matrix. This is not in contrast with Figure~\ref{figure:metrics_rank_correlation} where we observed MIG being more correlated with DCI Disentanglement rather than SAP Score. Indeed, the dissimilarity between MIG and SAP depends on differences in the estimation of the matrix as we show in Section~\ref{sec:matrix}. 

These results may not be surprising given the insights presented by~\citet{eastwood2018framework}. MIG and SAP computes the gap between the entries of the matrix per factor and therefore penalize compactness rather than disentanglement. In other words, they penalize whether a factor of variation is embedded in multiple codes but do not penalize the same code capturing multiple factors. DCI Disentanglement instead penalizes whether a code is related to multiple factors. Observing these differences in a large pool of trained models is challenging. First, the representations are not evenly distributed across the possible configurations (one-to-one, one-to-many, many-to-one and many-to-many) and for some of these relations (such as one-to-one and many-to-many) the scores behave similarly. Second, when comparing aggregations computed on different matrices it is typically unclear where the difference is coming from. However, we believe it is important to understand these practical differences as enforcing different notions of disentanglement may not result in the same benefits downstream. 

We conclude that the similarity between the scores in Section~\ref{sec:metrics_agreement} is confounded by how the statistical relations are computed. Further, we note that one-to-one or many-to-many mappings are preferred to one-to-many in the models we train, partially supporting the insights from~\citet{rolinek2018variational}.

\begin{figure}[pt]
\begin{subfigure}{0.3\textwidth}
\centering\includegraphics[width=\textwidth]{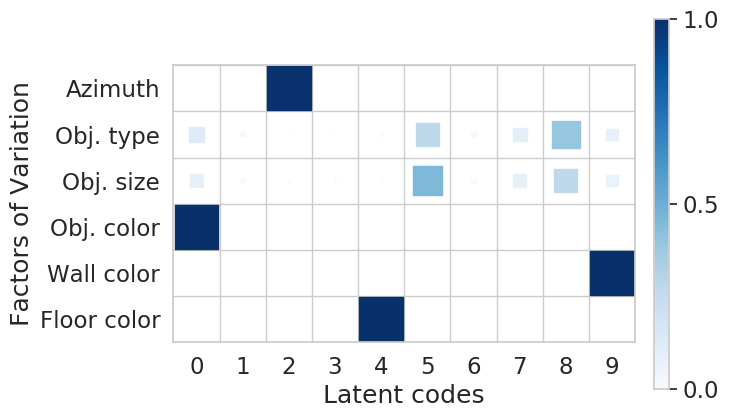}
\end{subfigure}%
\begin{subfigure}{0.3\textwidth}
\centering\includegraphics[width=\textwidth]{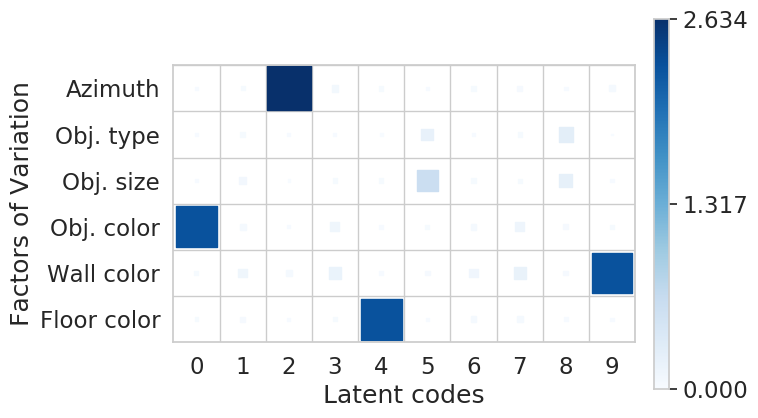}
\end{subfigure}
\begin{subfigure}{0.3\textwidth}%
\centering\includegraphics[width=\textwidth]{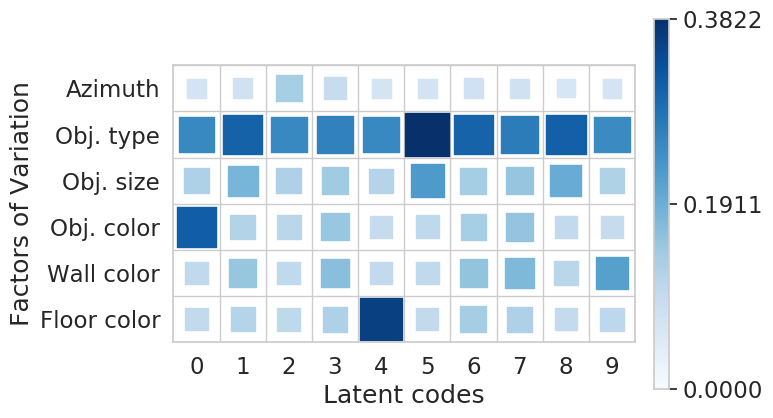}
\end{subfigure}

\begin{subfigure}{0.3\textwidth}
\centering\includegraphics[width=\textwidth]{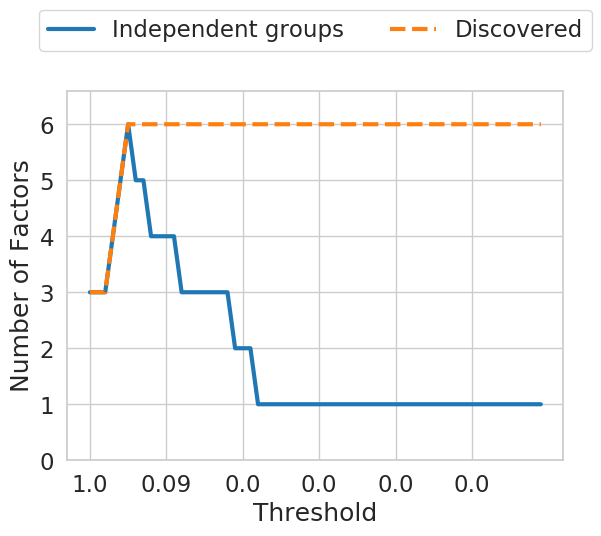}
\end{subfigure}%
\begin{subfigure}{0.3\textwidth}
\centering\includegraphics[width=\textwidth]{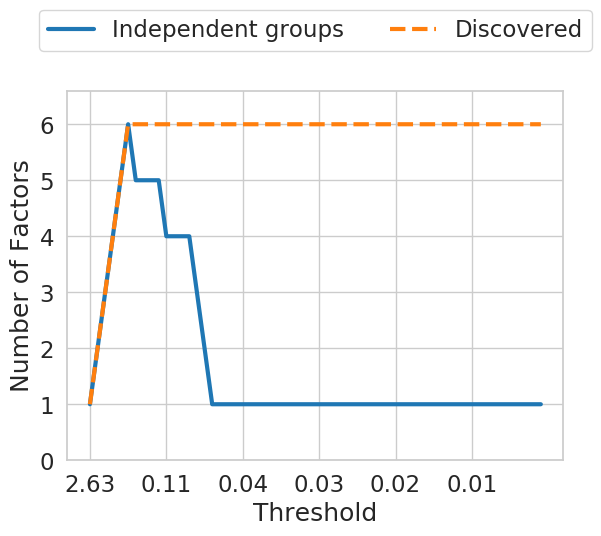}
\end{subfigure}
\begin{subfigure}{0.3\textwidth}%
\centering\includegraphics[width=\textwidth]{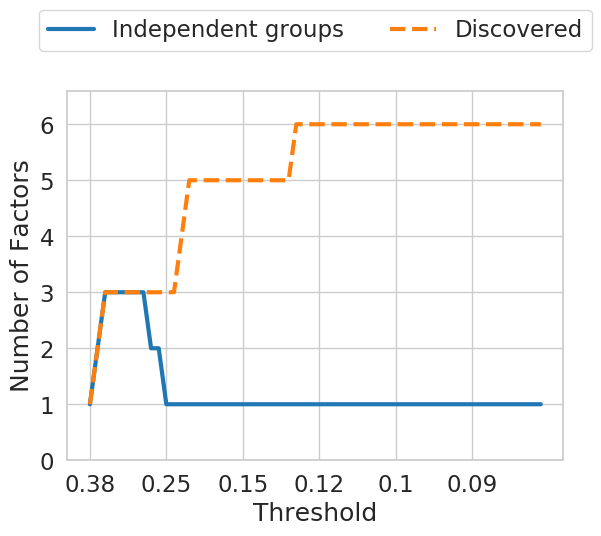}
\end{subfigure}

\begin{subfigure}{0.3\textwidth}
\centering\includegraphics[width=\textwidth]{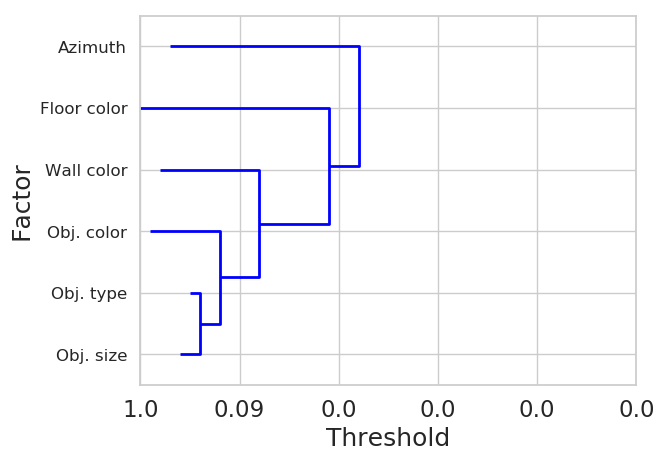}
\end{subfigure}%
\begin{subfigure}{0.3\textwidth}
\centering\includegraphics[width=\textwidth]{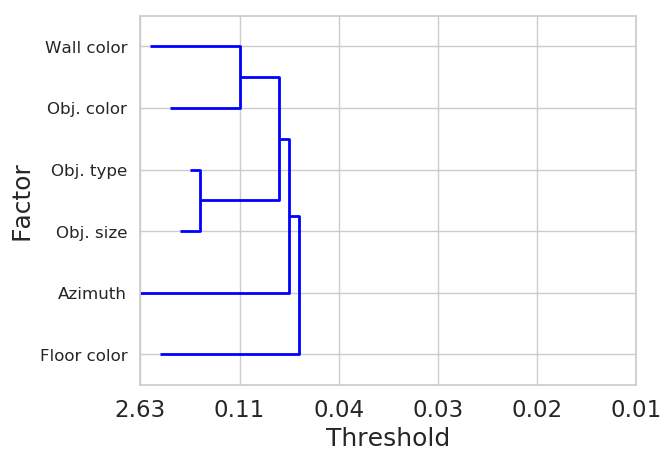}
\end{subfigure}
\begin{subfigure}{0.3\textwidth}%
\centering\includegraphics[width=\textwidth]{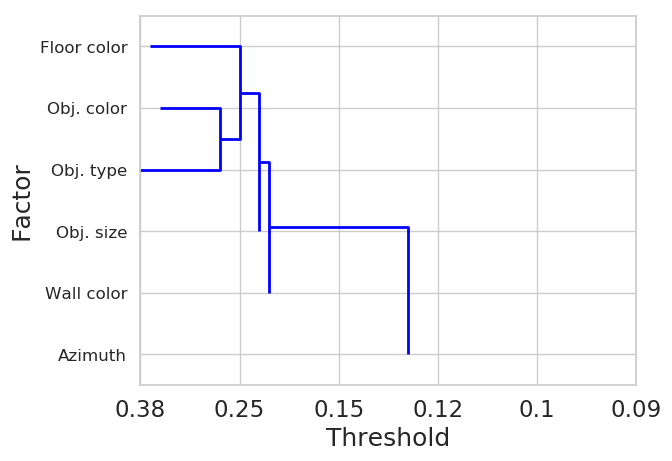}
\end{subfigure}
\caption{ Visualization of the relation between factors of variations and latent codes using for the model at the top of Figure~\ref{figure:counterexample_traversals}: (left) GBT feature importance as in the DCI Disentanglement score, (center) the mutual information as computed in the MIG and Modularity and, (right) SVM predictability as computed by the SAP Score. Top row: factor-code matrix. Middle row: independent-groups curve recording how many connected components of size larger than one there are in the factor-code bipartite graph defined by the matrix at a given threshold. Bottom row: dendrogram plot recording which factors are merged at which threshold. The long tail of the SVM importance matrix explains the weaker correlation between MIG and SAP Score in Figure~\ref{figure:metrics_rank_correlation} even though the scores are measuring a similar concept. The dendrogram plots computed from the independent-groups curve can be used to systematically analyze which factors are merged at which threshold by the different estimation techniques (e.g. SVM, GBT feature importance and mutual information).}\label{figure:counterexample}
\end{figure}

\subsubsection{Does the Estimation Factor-Code Matrices Impact the Evaluation?}~\label{sec:matrix}
In this section, we continue to investigate the metrics that can be computed from observational data and focus on the different matrices estimating the statistical relations between factors of variation and latent codes. First, we build new visualization tools that allow us to understand both what a model has learned and how its been evaluated by the factor-code matrices.
\begin{figure}[t]
\begin{center}
\begin{subfigure}{0.3\textwidth}
\centering\includegraphics[width=\textwidth]{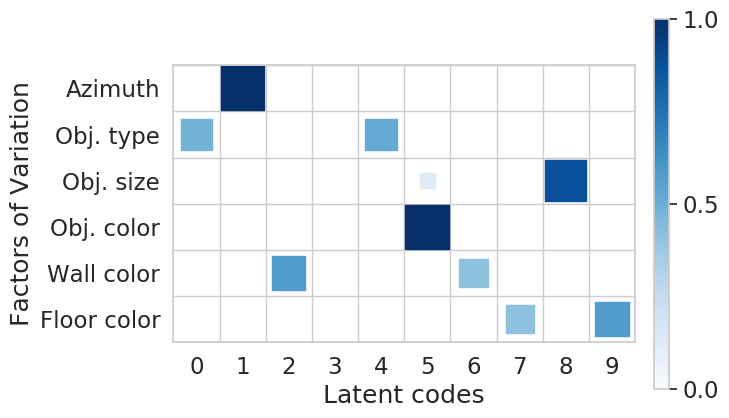}
\centering\includegraphics[width=\textwidth]{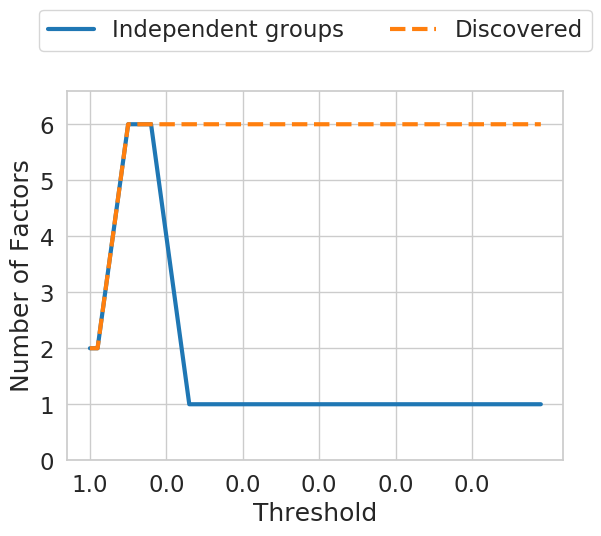}
\centering\includegraphics[width=\textwidth]{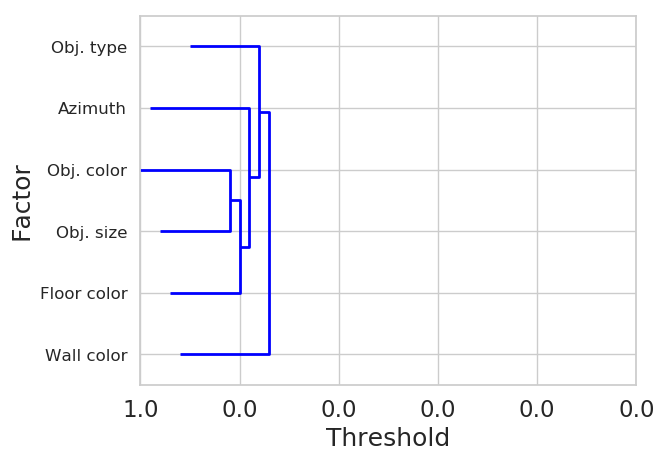}
\end{subfigure}%
\begin{subfigure}{0.3\textwidth}
\centering\includegraphics[width=\textwidth]{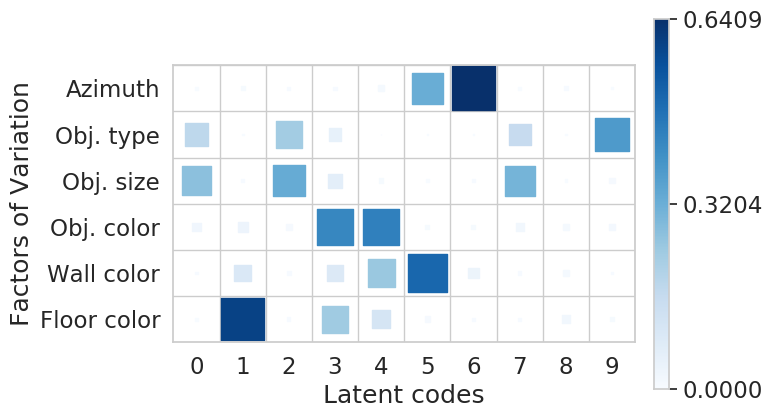}
\centering\includegraphics[width=\textwidth]{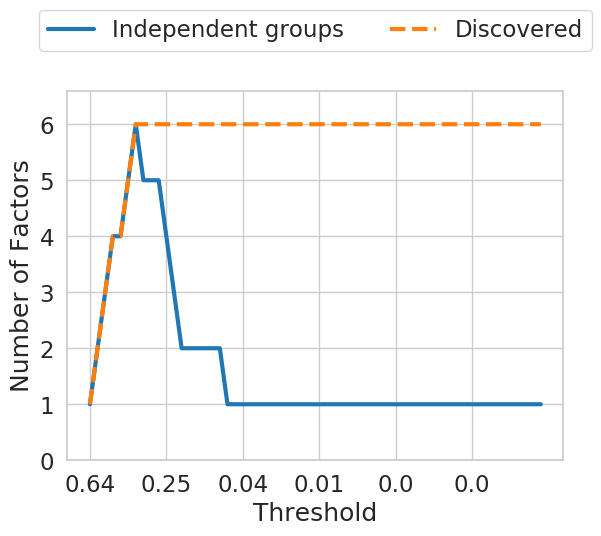}
\centering\includegraphics[width=\textwidth]{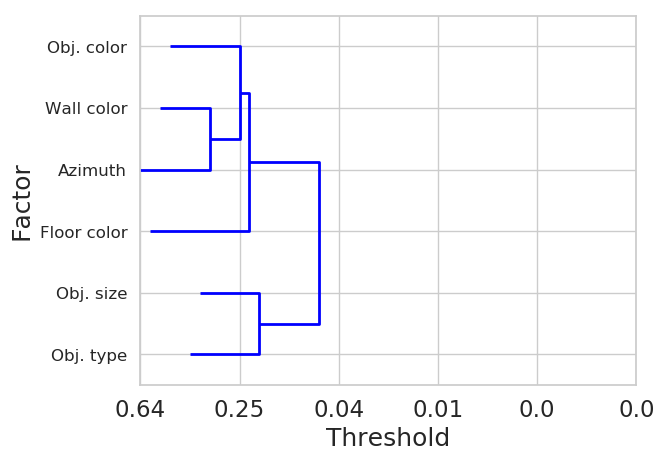}
\end{subfigure}%
\begin{subfigure}{0.3\textwidth}
\centering\includegraphics[width=\textwidth]{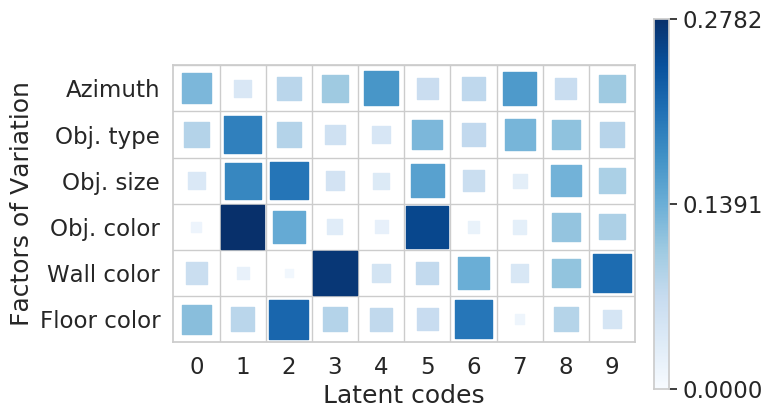}
\centering\includegraphics[width=\textwidth]{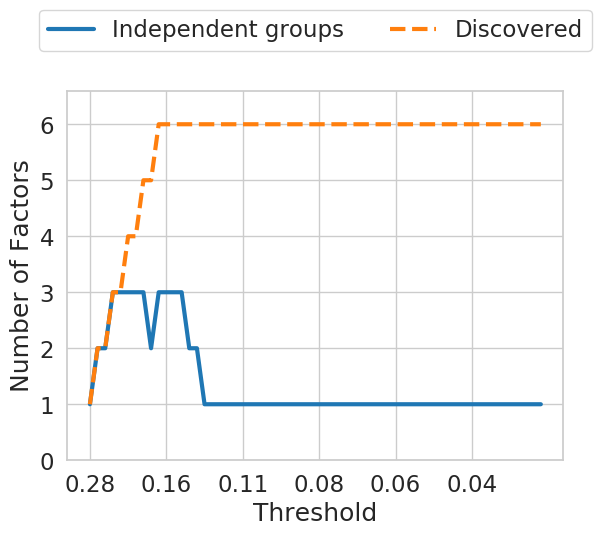}
\centering\includegraphics[width=\textwidth]{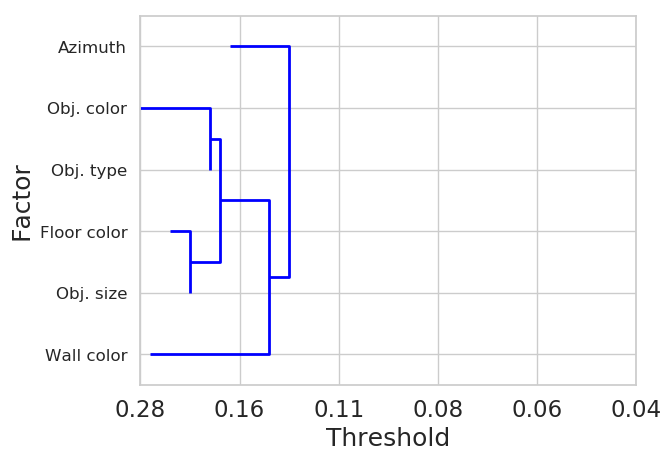}
\end{subfigure}
\end{center}
\caption{(top row) Visualization of the GBT importance matrix used in the DCI Disentanglement score for models with top (left), average (center), and worse (right) DCI Disentanglement on Shapes3D. (middle row) Independent-groups curves of the GBT importance matrix. (bottom row) Dendrogram plot recording when factors are merged. Comparing these plots with the ones in Figure~\ref{figure:precision_curves_mi}, we note that there are differences in the factor-code matrices. In particular, they disagree on which factors are most entangled. }~\label{figure:precision_curves_gbt}
\end{figure}

\begin{figure}[t]
\begin{center}
\begin{subfigure}{0.3\textwidth}
\centering\includegraphics[width=\textwidth]{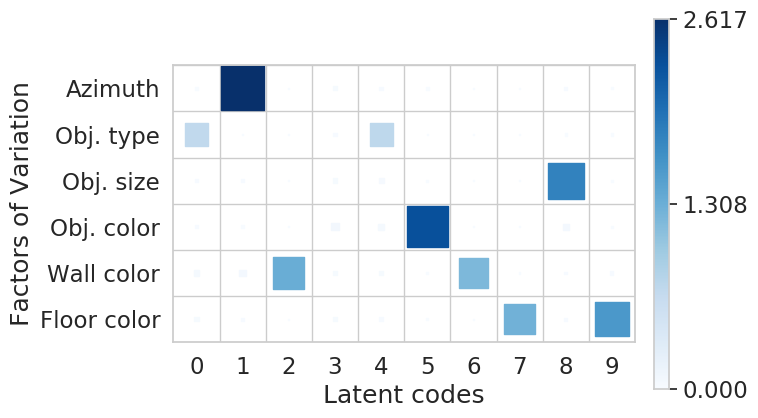}
\centering\includegraphics[width=\textwidth]{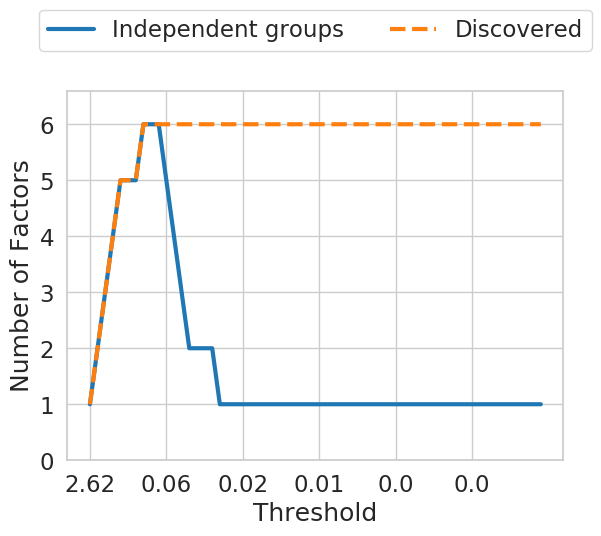}
\centering\includegraphics[width=\textwidth]{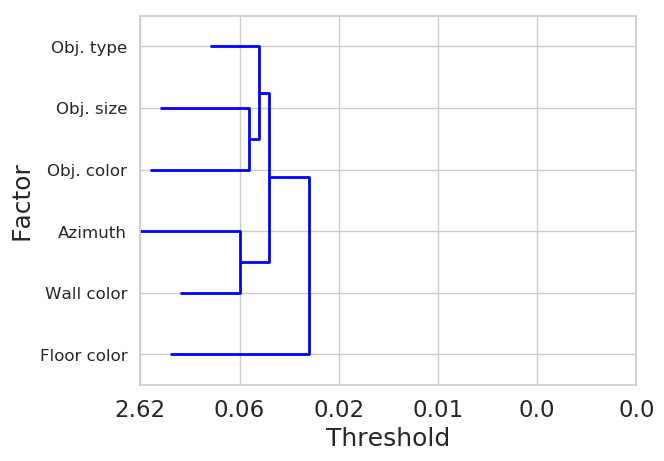}
\end{subfigure}%
\begin{subfigure}{0.3\textwidth}
\centering\includegraphics[width=\textwidth]{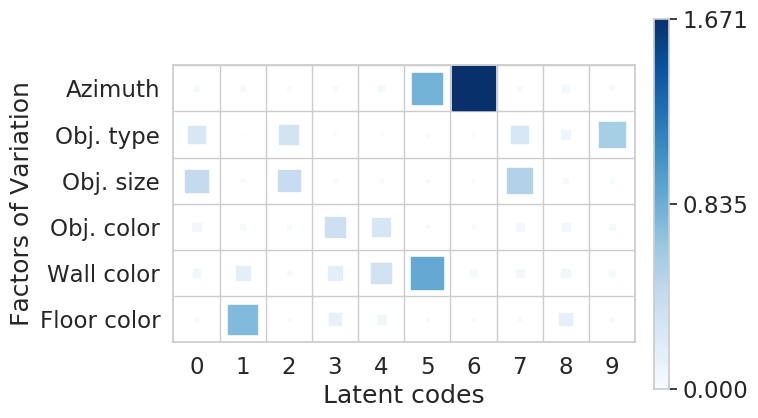}
\centering\includegraphics[width=\textwidth]{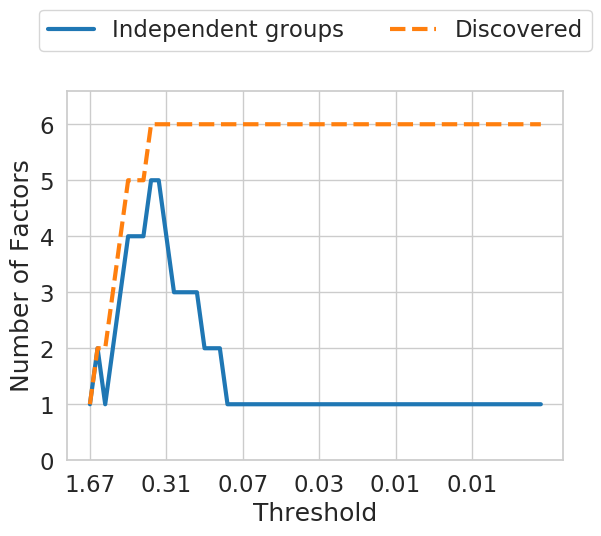}
\centering\includegraphics[width=\textwidth]{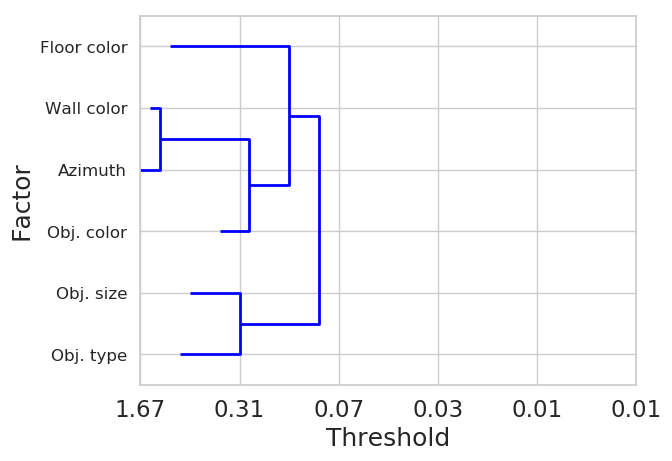}
\end{subfigure}%
\begin{subfigure}{0.3\textwidth}
\centering\includegraphics[width=\textwidth]{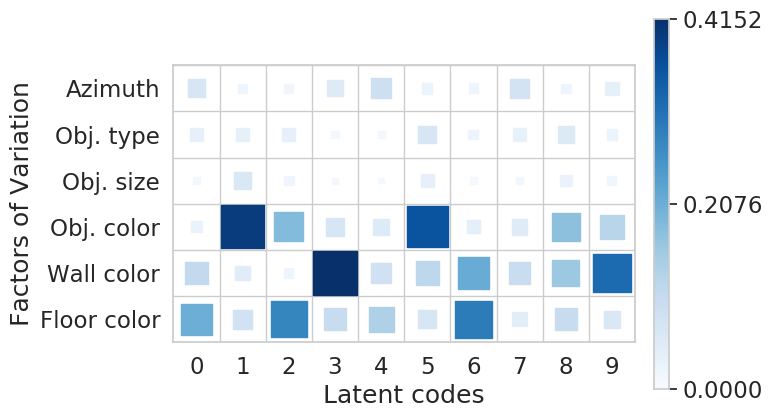}
\centering\includegraphics[width=\textwidth]{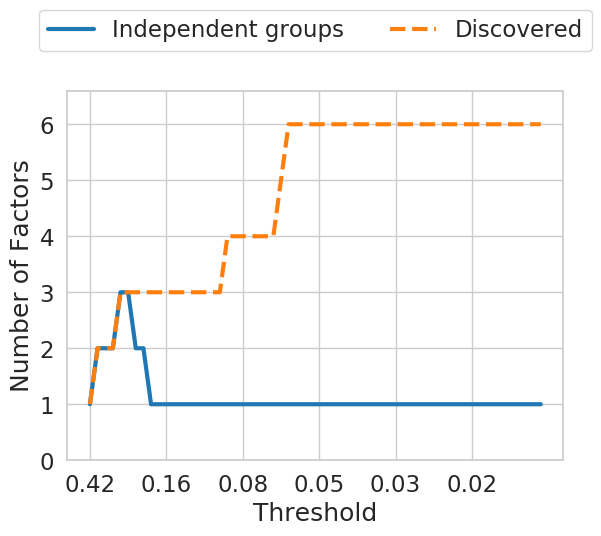}
\centering\includegraphics[width=\textwidth]{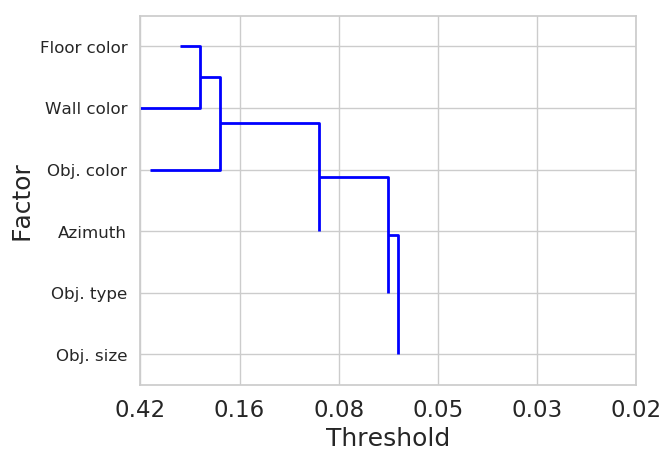}
\end{subfigure}%
\caption{
(top row) Visualization of the mutual information matrix used in the MIG and Modularity scores for the same models of Figure~\ref{figure:precision_curves_gbt}. (middle row) Independent-groups curves of the mutual information matrix. (bottom row) Dendrogram plot recording when factors are merged. Comparing these plots with the ones in Figure~\ref{figure:precision_curves_gbt}, we note that there are differences in the factor-code matrices. In particular, they disagree on which factors are most entangled. }~\label{figure:precision_curves_mi}
\end{center}
\end{figure}

In Figure \ref{figure:counterexample} we visualize the model at the bottom of Figure~\ref{figure:counterexample_traversals}. On the first row, we plot the factor-codes matrices as learned by GBT feature importance, pairwise mutual information and SVM predictability respectively. We observe that for the GBT features and the mutual information matrix the largest entries are the same but the latter underestimates the effect of some dependencies, for example object size and type in dimensions number five and eight. The SVM feature importance, also agrees on some of the large values but exhibit a longer tail compared to the other matrices.

In order to further analyze the differences between the matrices we view them as weights on the edges of a bipartite graph encoding the statistical relation between each factor of variation and code. We can now delete all edges with weight smaller than some threshold and count (i) how many factors of variation are connected with at least a latent code and (ii) the number of connected components with size larger than one. In Figure~\ref{figure:counterexample} (middle row), we plot these two curves computed on the respective matrices, and, in Figure~\ref{figure:counterexample} (bottom row), we record which factors are merged at which threshold.  Factors that are merged at lower threshold are more entangled in the sense that are more statistically related to a shared latent dimension.

The long tail of the SVM importance matrix explains why we observed a weaker correlation between MIG and SAP Score in Figure~\ref{figure:metrics_rank_correlation} even though the scores are measuring a similar concept. Indeed, we can observe in the middle row of Figure~\ref{figure:counterexample} that the largest entries of the three matrices are distributed differently, in particular for the SVM predictability. Similarly, we can read in the dendrogram plot that the factors are merged in a significantly different order for the SVM predictability compared to the other two matrices. We hypothesize that the long tail of the SVM predictability is a consequence of spurious correlations and optimization issues that arise from how the score is computed (fitting a threshold separately on each code predicting each factor).

In Figures~\ref{figure:precision_curves_gbt} and~\ref{figure:precision_curves_mi} we compare the factor-code matrices, independent-groups curves, and dendrograms for the best, average and worse model in terms of DCI Disentanglement. Figure~\ref{figure:precision_curves_gbt} shows the plots for the GBT (Gradient Boosted Trees) feature importance matrix used by the DCI Disentanglement score and Figure~\ref{figure:precision_curves_mi} the mutual information matrix of MIG and Modularity. By comparing these plots, we can clearly distinguish which model is the most disentangled but we again note differences in how the factors of variation are captured by the different matrices. In particular, we again observe that the two matrices may disagree on which factors are most entangled in the same model. For example, the GBT features computed on the model on the left suggest that object color and size are more entangled while the mutual information matrix suggest azimuth and wall color.
These differences appear to be systematic. From the dendrogram plot of each model and estimation matrix, we can compute at which threshold each pair of factors is merged on average. This allow us to systematically analyze the differences in terms of which factors are found more entangled by the different matrices.
In Figure~\ref{figure:confused_factors_dSprites} in the Appendix, we can see that on dSprites and Color-dSprites some factors of variation are consistently entangled across different data set and estimation matrices indicating that they are hardest to disentangle. On the other variants, the different matrices significantly disagree and similar results can be observed in Figure~\ref{figure:confused_factors_other} in the Appendix for the other data sets. This indicates systematic differences in the structure found by different estimation techniques which may impact the final computation of the scores.

\begin{figure}[pt]
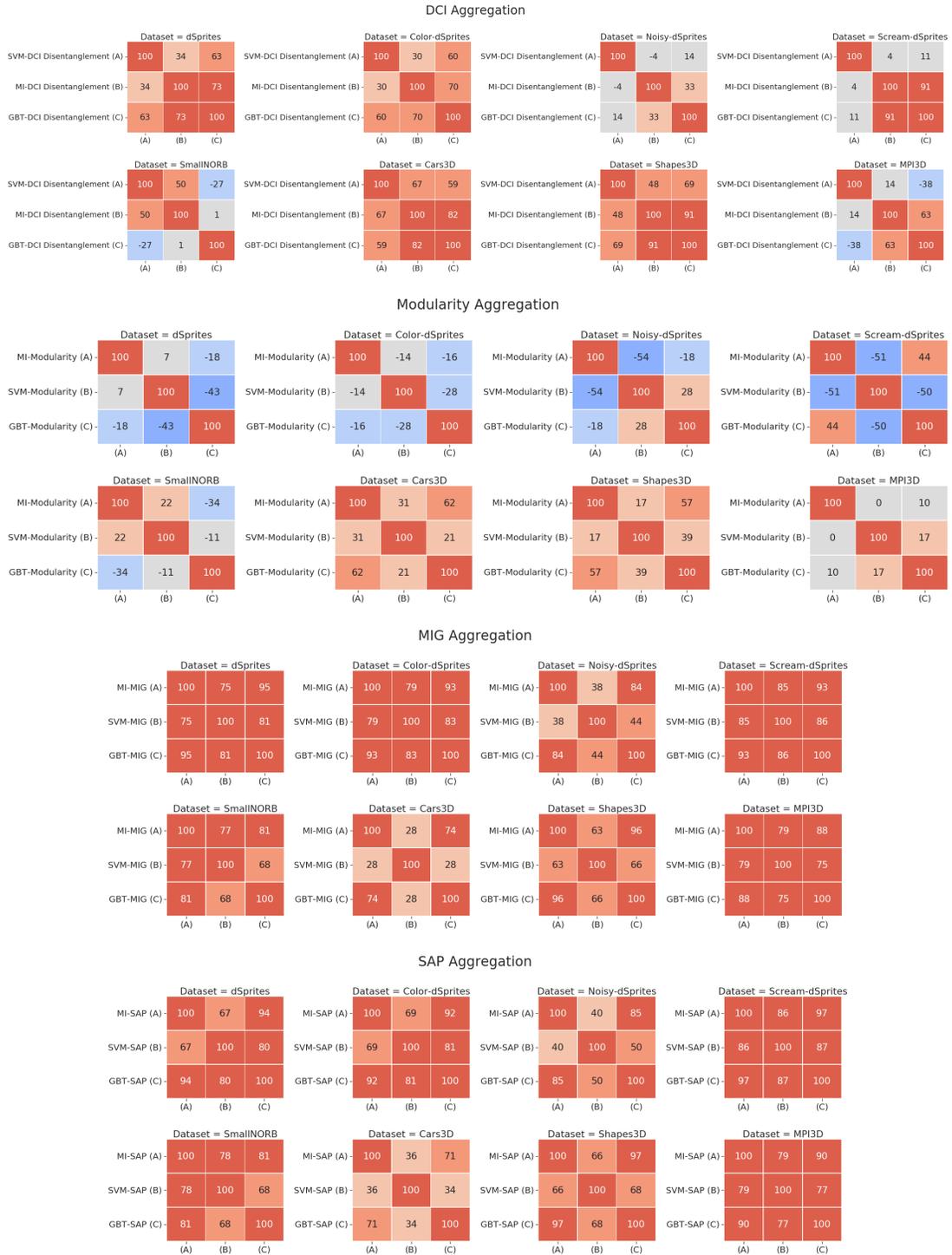

\centering\includegraphics[width=0.95\textwidth]{autofigures/metrics_rank_dci_aggregation}\vspace{3mm}
\centering\includegraphics[width=0.95\textwidth]{autofigures/metrics_rank_modularity_aggregation}\vspace{3mm}
\centering\includegraphics[width=0.75\textwidth]{autofigures/metrics_rank_mig_aggregation}\vspace{3mm}
\centering\includegraphics[width=0.75\textwidth]{autofigures/metrics_rank_sap_aggregation}
\caption{Rank correlation of DCI Disentanglement, Modularity, SAP Score and MIG aggregations on different matrices. The ranking seem to be generally different and data set dependant indicating that systematic differences in the estimation matrix may impact the evaluation of disentanglement. MIG and SAP aggregations appear to be more robust to changes in the estimation matrix.
}\label{figure:metrics_rank_sap_mig_aggregation}
\end{figure}

Finally, we test whether the differences in the factor-code matrix impact the computation of the disentanglement scores. To do so, we compare the ranking produced by each aggregation computed on the different matrices. If the different matrices encode the same statistical relations, the ranking should also be similar.
We observe in Figure~\ref{figure:metrics_rank_sap_mig_aggregation} that the ranking seem to be generally different and the level of correlation appears to depend on the data set. Overall, the aggregation of SAP Score and MIG seem to be more robust to changes in the estimation matrix compared to Modularity and DCI Disentanglement. 

Based on this result, we conclude that systematic differences in the estimation matrix may indeed impact the evaluation of disentanglement. It seem important for the evaluation that the statistical relations between factors and codes are robustly and consistently estimated. We observed that changing the estimation technique may produce different rankings of the models. It appears therefore important to not bias the evaluation by considering a single estimation technique, unless reliability guarantees are also given.

\subsubsection{Implications} 

We conclude that the different disentanglement scores are not measuring the same concept: they measure different notions of disentanglement (compactness versus disentanglement) that are generally correlated in practice but not equivalent. 

In particular, MIG and SAP Score intend disentanglement differently than DCI Disentanglement as they are rather measure completeness: they do not penalize multiple factors of variation being captured by a single latent dimension. Modularity seem to be more dependent on the estimation matrix as its correlation with the other scores changes significantly with different matrices. Furthermore, there are systematic differences between the different techniques to estimate the relation between factors of variation and latent codes that influence the correlation of the scores: the ranking of the models is different depending on the chosen estimation technique.

We argue that future works advancing the state-of-the-art in disentanglement, with or without any form of supervision, should reflect upon which notion of disentanglement they consider and how it is measured in the chosen evaluation protocol. 

\looseness=-1Not all the properties that are generally associated with the term ``disentanglement'' are necessarily related to all the scores considered in this paper and specific downstream tasks may require specific notions~\citep{locatello2019fairness,van2019disentangled,locatello2020weakly}. Further, separating the estimation of the statistical dependencies between factors of variation and codes from what the score is measuring may help clarify the properties that are being evaluated. As robustly capturing these statistical dependencies is a crucial step of the evaluation metrics that do not rely on interventions, we argue that future work on disentanglement scores should specifically highlight (i) how this estimation is performed precisely, (ii) its sample complexity/variance and (iii) biases (for example do they work well with coarse grained as opposed to fine grained factors of variation). Future research is necessary to understand both how estimation metrics overestimate or underestimate the amount of disentanglement and how to robustly aggregate this information into a score. Among the scores tested in this paper, we recommend to use the DCI aggregation, either with the GBT feature importance or the mutual information matrix, ideally both. 

\begin{figure}[pt]
\centering\includegraphics[width=\textwidth]{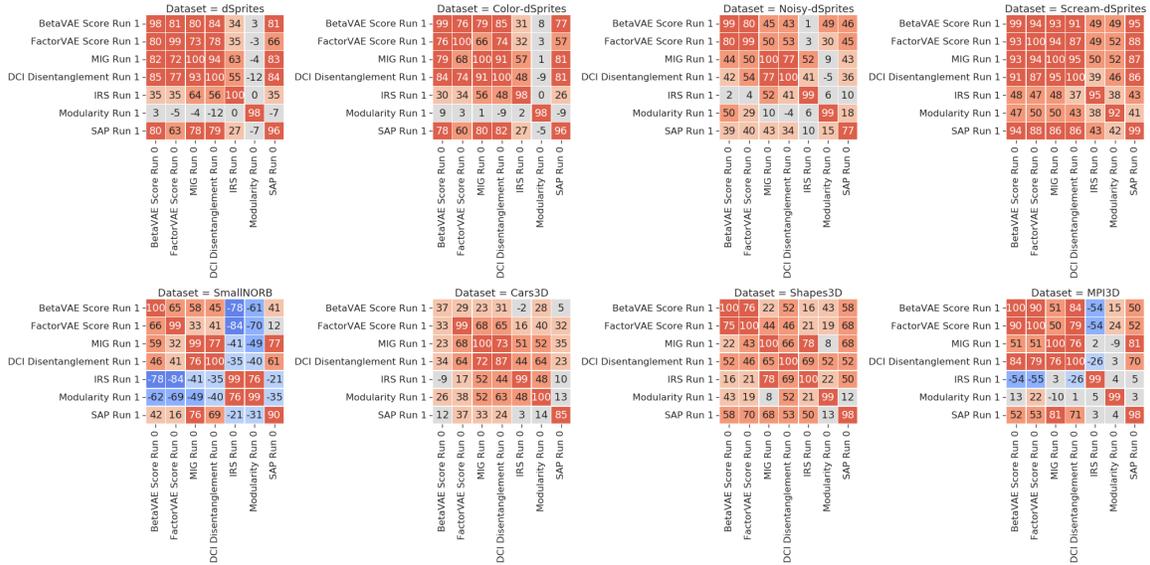}
\caption{Rank correlation of different metrics on different data sets across two runs. 
Overall, we observe that the disentanglement scores computed with \num{10000} examples are relatively stable.
}\label{figure:metrics_rank_correlation_10000}
\end{figure}
\begin{figure}[pt]
\centering\includegraphics[width=\textwidth]{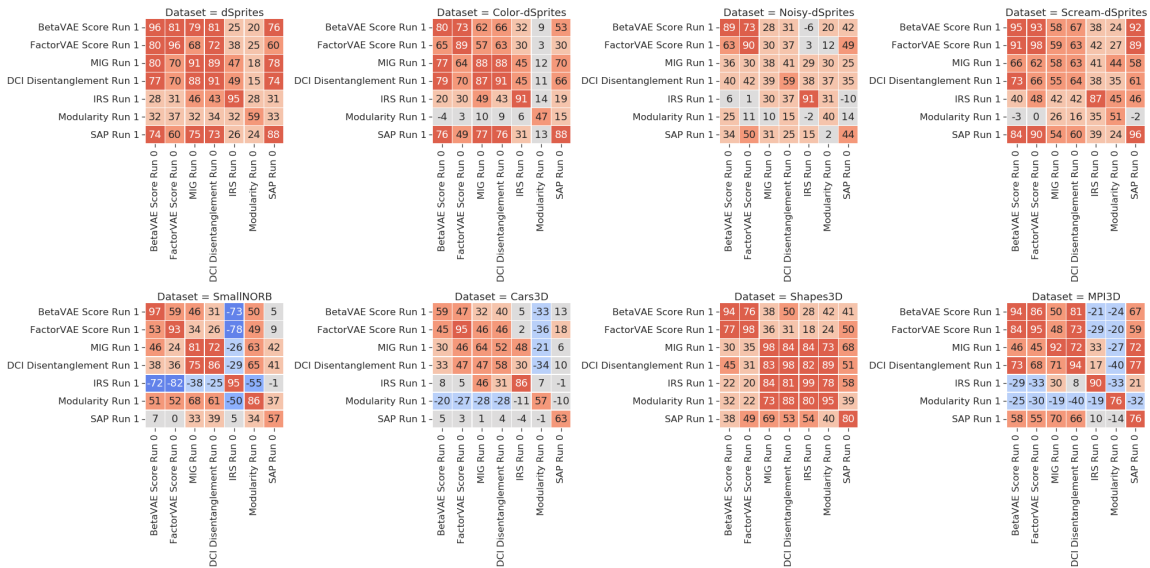}
\caption{Rank correlation of different metrics computed using \num{100} examples on different data sets across two runs. 
Overall, we observe that with fewer examples the disentanglement scores are significantly less stable.
}\label{figure:metrics_rank_correlation_100}
\end{figure}

\subsection{Is the Computation of the Disentanglement Scores Reliable?}
The computation of the disentanglement scores require supervision and having access to a large number of observations of $\rvz$ may be unreasonable. 
\looseness=-1On the other hand, for the purpose of this study we are interested in a stable and reproducible experimental setup. In Figure~\ref{figure:metrics_rank_correlation_10000}, we observe that running the disentanglement scores twice yields comparable results with \num{10000} examples. Using just \num{100} examples may be feasible in practice as suggested by~\cite{locatello2019disentangling} but has less stable results as depicted in Figure~\ref{figure:metrics_rank_correlation_100}. We observe that not every score is equally sample efficient. The FactorVAE scores and the IRS seem to be the most efficient ones, followed by DCI Disentanglement and MIG.

\subsubsection{Implications}
Computing the disentanglement scores on these data sets with \num{10000} examples yields stable results and is appropriate for the purpose of this study. Finding sample efficient disentanglement scores is an important research direction for practical semi-supervised disentanglement~\citep{locatello2019disentangling}.

\begin{figure}[pt]
\centering\includegraphics[width=\textwidth]{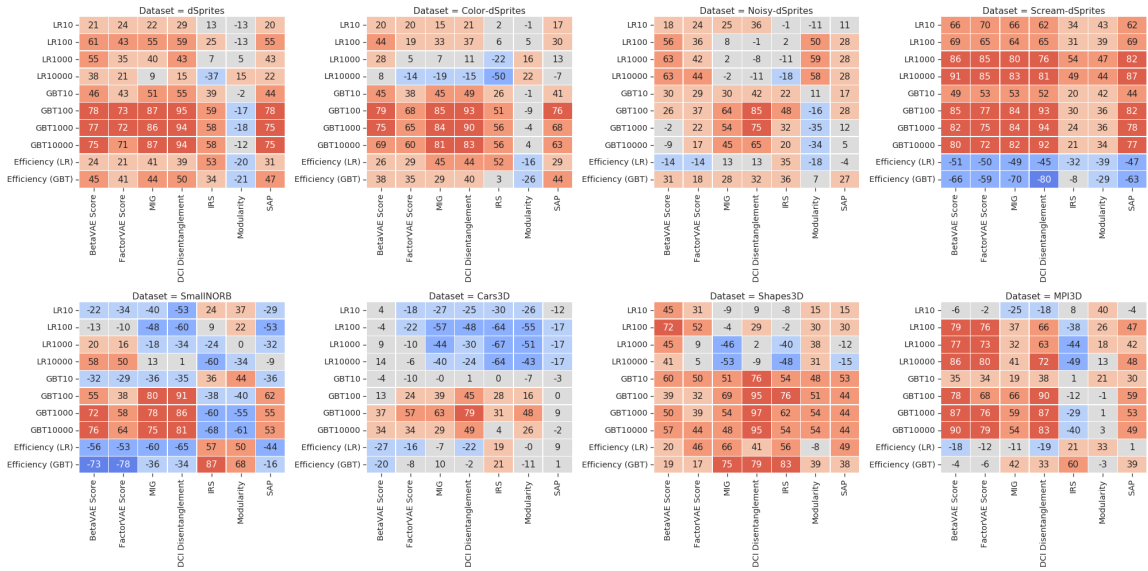}
\caption{Rank-correlation between the metrics and the performance on downstream task on different data sets. We observe some correlation between most disentanglement metrics and downstream performance. However, the correlation varies across data sets.
}\label{figure:downstream_tasks}
\end{figure}
\begin{figure}[pt]
\centering\includegraphics[width=\textwidth]{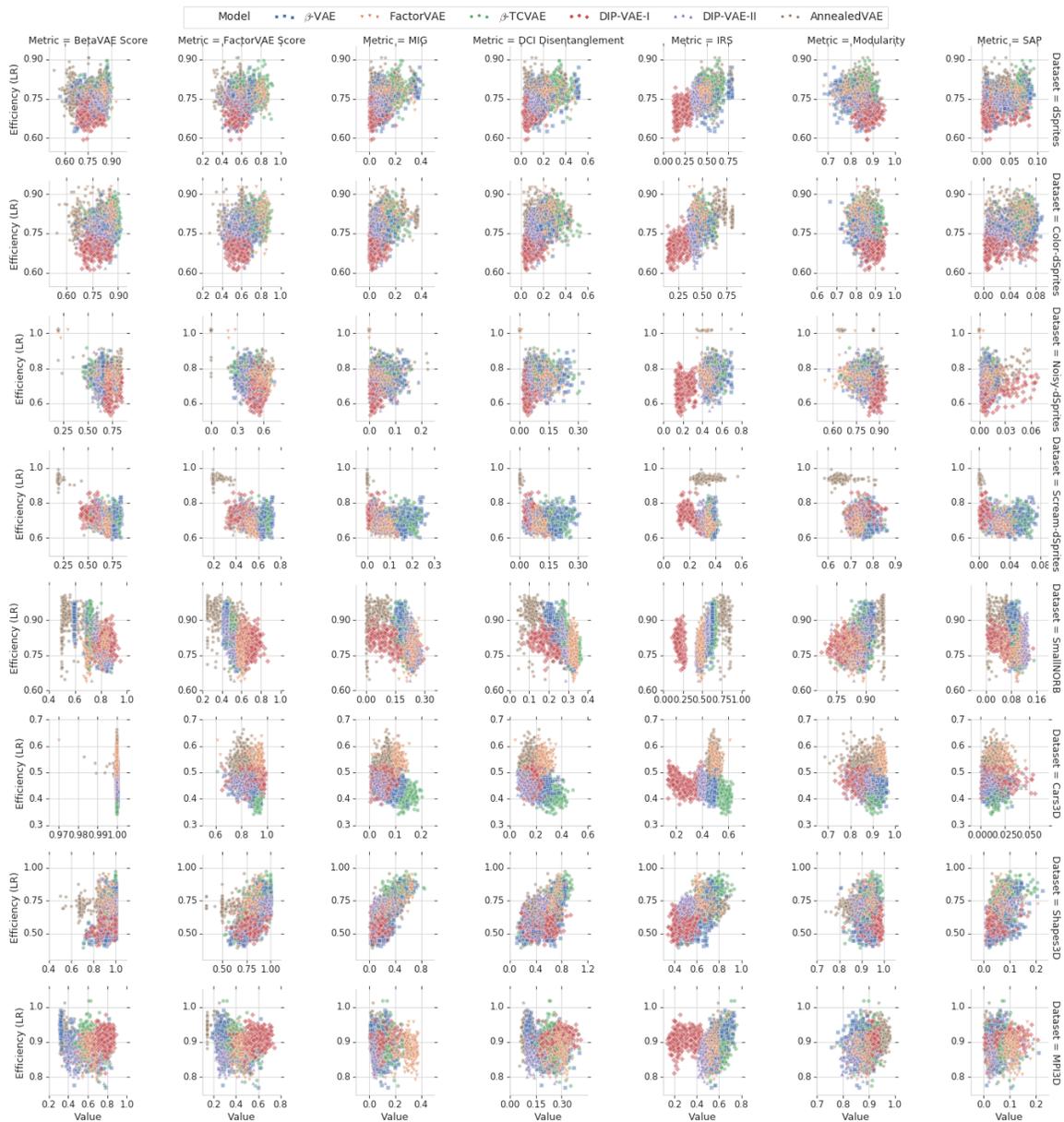}
\caption{Statistical efficiency (accuracy with $\num{100}$ samples $\div$ accuracy with $\num{10000}$ samples) based on a logistic regression versus disentanglement metrics for different models and data sets. We do not observe that higher disentanglement scores lead to higher statistical efficiency.}\label{figure:stat_efficiency_lr}
\end{figure}
\begin{figure}[pt]
\centering\includegraphics[width=\textwidth]{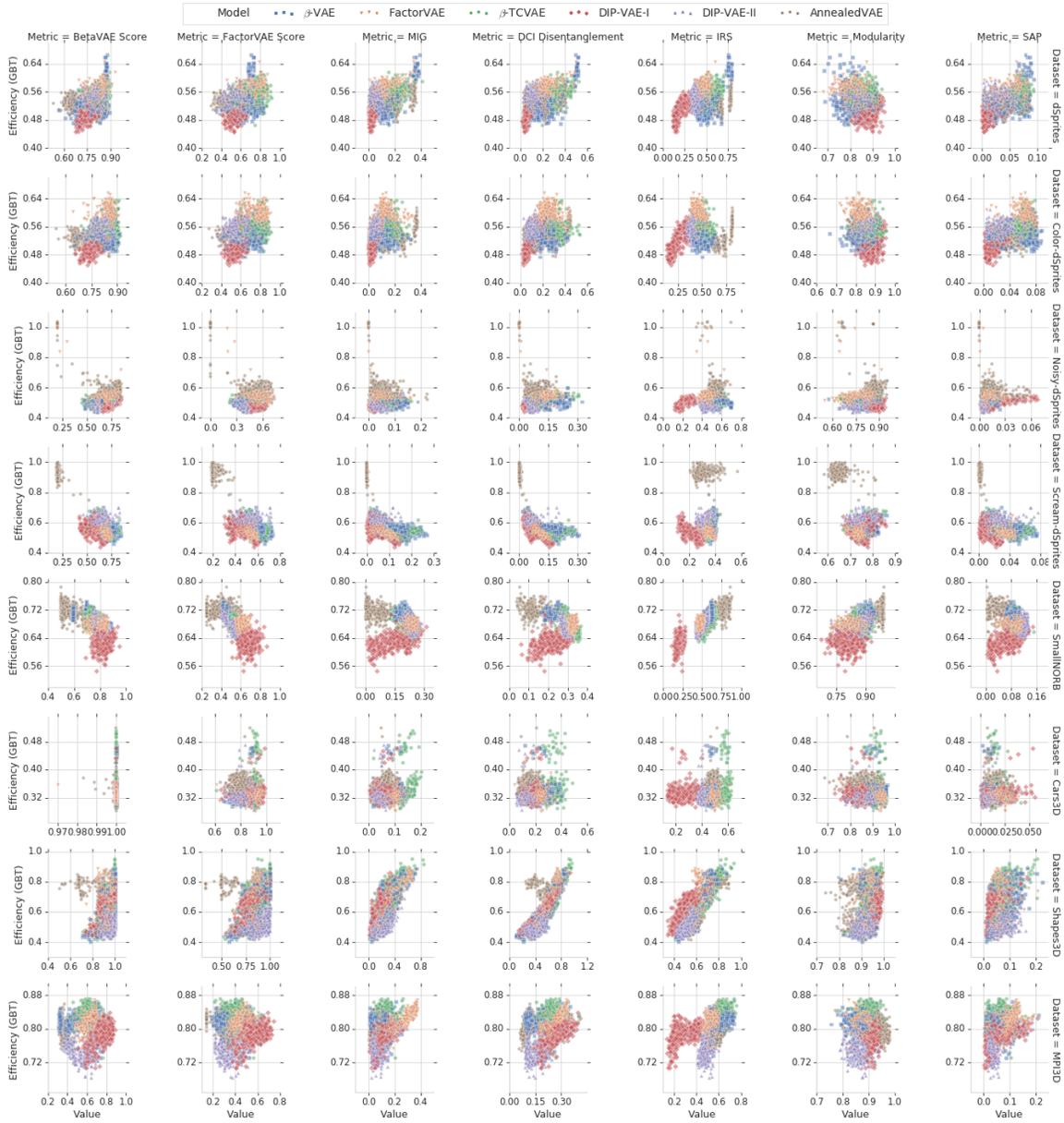}
\caption{Statistical efficiency (accuracy with $\num{100}$ samples $\div$ accuracy with $\num{10000}$ samples) based on gradient boosted trees versus disentanglement metrics for different models and data sets. We do not observe that higher disentanglement scores lead to higher statistical efficiency (except for DCI Disentanglement and Mutual Information Gap on Shapes3D and to some extend in Cars3D).}\label{figure:stat_efficiency_gbt}
\end{figure}
\begin{figure}[t]
\centering\includegraphics[width=\textwidth]{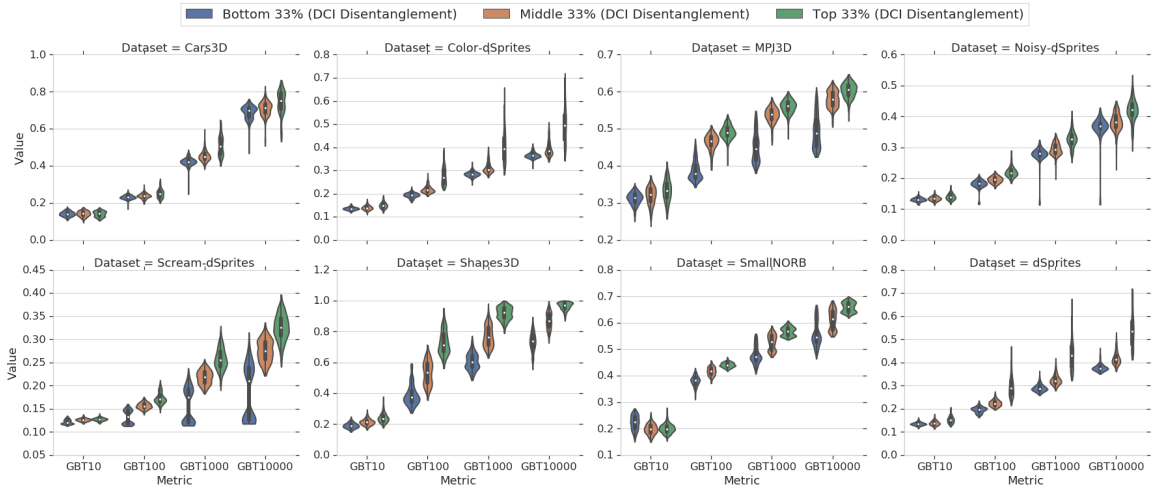}
\caption{Downstream performance for three groups with increasing DCI Disentanglement scores.}\label{figure:dci_downstream_gbt_analysis}
\end{figure}
\begin{figure}[t]
\centering\includegraphics[width=\textwidth]{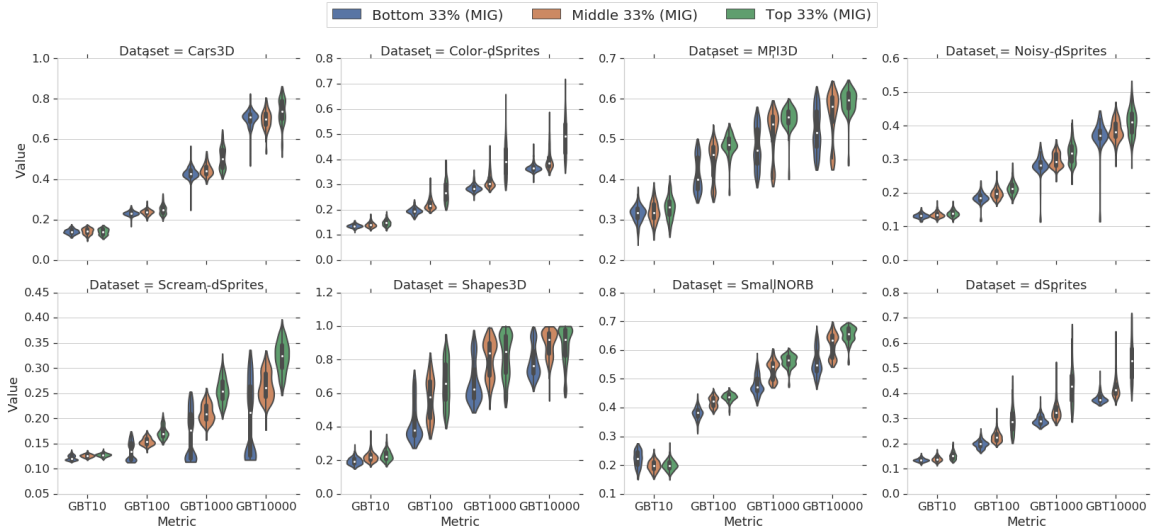}
\caption{Downstream performance for three groups with increasing MIG scores.}\label{figure:mig_downstream_gbt_analysis}
\end{figure}
\begin{figure}[t]
\centering\includegraphics[width=\textwidth]{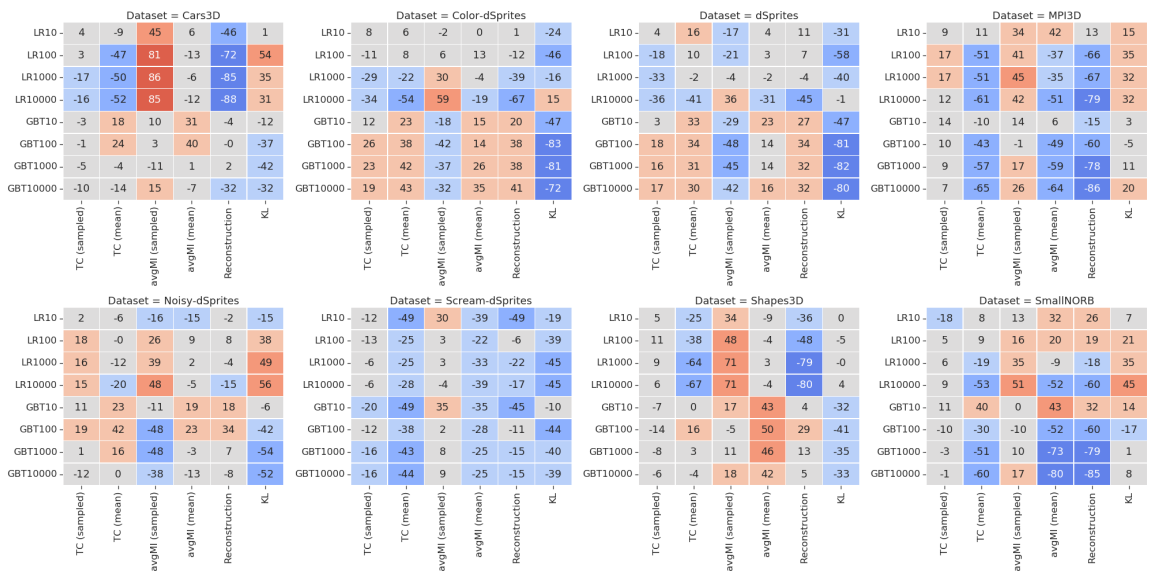}
\caption{Rank correlation between unsupervised scores and downstream performance.}\label{figure:unsup_vs_downstream}
\end{figure}

\section{Are These Disentangled Representations Useful for Downstream Tasks in Terms of the Sample Complexity of Learning?}\label{sec:downstream}
One of the key motivations behind disentangled representations is that they are assumed to be useful for later downstream tasks.
In particular, it is argued that disentanglement should lead to a better sample complexity of learning~\citep{bengio2013representation,scholkopf2012causal,peters2017elements}.
In this section, we consider the simplest downstream classification task where the goal is to recover the true factors of variations from the learned representation using either multi-class logistic regression (LR) or gradient boosted trees (GBT).
Our goal is to investigate the relationship between disentanglement and the average classification accuracy on these downstream tasks as well as whether better disentanglement leads to a decreased sample complexity of learning.

To compute the classification accuracy for each trained model, we sample true factors of variations and observations from our ground truth generative models.
We then feed the observations into our trained model and take the mean of the Gaussian encoder as the representations.
Finally, we predict each of the ground-truth factors based on the representations with a separate learning algorithm.
We consider both a 5-fold cross-validated multi-class logistic regression as well as gradient boosted trees of the Scikit-learn package.
For each of these methods, we train on $\num{10}$, $\num{100}$, $\num{1000}$ and $\num{10000}$ samples.
We compute the average accuracy across all factors of variation using an additional set $\num{10000}$ randomly drawn samples.

Figure~\ref{figure:downstream_tasks} shows the rank correlations between the disentanglement metrics and the downstream performance for all considered data sets.
We observe that all metrics except Modularity seem to be correlated with increased downstream performance on the different variations of dSprites and to some degree on Shapes3D. 
However, it is not clear whether this is due to the fact that disentangled representations perform better or whether some of these scores actually also (partially) capture the informativeness of the evaluated representation. Furthermore, the correlation is weaker or inexistent on other data sets (for example, Cars3D). Finally, we report in Figure~\ref{figure:unsup_vs_downstream} the rank correlation between unsupervised scores computed after training on the mean and sampled representation and downstream performance.
Depending on the data set, the rank correlation ranges from from mildly negative, to mildly positive.
In particular, we do not observe enough evidence supporting the claim that decreased total correlation of the aggregate posterior proves beneficial for downstream task performance. 

To assess the sample complexity argument we compute for each trained model a statistical efficiency score which we define as the average accuracy based on $\num{100}$ samples divided by the average accuracy based on $\num{10000}$ samples for either the logistic regression or the gradient boosted trees.
The key idea is that if disentangled representations lead to sample efficiency, then they should also exhibit a higher statistical efficiency score. We remark that this score differs from the definition of sample complexity commonly used in statistical learning theory.
The corresponding results are shown in Figures~\ref{figure:stat_efficiency_lr} and~\ref{figure:stat_efficiency_gbt} where we plot the statistical efficiency versus different disentanglement metrics for different data sets and models and in Figure~\ref{figure:downstream_tasks} where we show rank correlations.
Overall, we do not observe conclusive evidence that models with higher disentanglement scores also lead to higher statistical efficiency.
We note that some AnnealedVAE models seem to exhibit a high statistical efficiency on Scream-dSprites and to some degree on Noisy-dSprites.
This can be explained by the fact that these models have low downstream performance and that hence the accuracy with $\num{100}$ samples is similar to the accuracy with $\num{10000}$ samples.
We further observe that DCI Disentanglement and MIG seem to be lead to a better statistical efficiency on the the data set Shapes3D for gradient boosted trees.
Figures~\ref{figure:dci_downstream_gbt_analysis} and~\ref{figure:mig_downstream_gbt_analysis} show the downstream performance for three groups with increasing levels of disentanglement (measured in DCI Disentanglement and MIG respectively).
We observe that indeed models with higher disentanglement scores seem to exhibit better performance for gradient boosted trees with 100 samples.
However, considering all data sets, it appears that overall increased disentanglement is rather correlated with better downstream performance (on some data sets) and not statistical efficiency.
We do not observe that higher disentanglement scores reliably lead to a higher sample efficiency.

\subsection{Implications}
While the empirical results in this section are negative, they should also be interpreted with care.
\looseness=-1After all, we have seen in previous sections that the models considered in this study fail to reliably produce disentangled representations. 
Hence, the results in this section might change if one were to consider a different set of models, for example semi-supervised or fully supervised one.
Furthermore, there are many more potential notions of usefulness such as interpretability and fairness that we have not considered in our experimental evaluation. While prior work~\citep{steenbrugge2018improving,laversanne2018curiosity,nair2018visual,higgins2017darla,higgins2018scan} successfully applied disentanglement methods such as $\beta$-VAE on a variety of downstream tasks, it is not clear to us that these approaches and trained models performed well \emph{because of disentanglement}. Finally, we remark that disentanglement is mostly about \textit{how} the information is stored in the representation. Tasks that explicitly rely on this structure are likely to benefit more from disentanglement rather than the ones considered in this paper. Notable examples are applications in fairness \citep{locatello2019fairness} and abstract visual reasoning \citep{van2019disentangled}. In the former, the authors show that disentanglement can be used to isolate the effect of unobserved sensitive variables to limit their negative impact to the downstream prediction. In the latter, the authors show compelling evidence that disentanglement is useful for abstract visual reasoning tasks in terms of sample complexity. We remark that the benefits~\citet{locatello2019fairness} and~\citet{van2019disentangled} observed are specific to some of the notions of disentanglement considered in this paper, such as DCI Disentanglement and FactorVAE. 

\section{Conclusions}
\label{sec:conclusion}
\looseness=-1In this work we first theoretically show that the unsupervised learning of disentangled representations is fundamentally impossible without inductive biases. We then performed a large-scale empirical study with six state-of-the-art disentanglement methods, seven disentanglement metrics on eight data sets and conclude the following:
(i) A factorizing aggregated posterior (which is sampled) does not seem to necessarily imply that the dimensions in the representation (which is taken to be the mean) are uncorrelated. 
(ii) Random seeds and hyperparameters seem to matter more than the model but tuning seem to require supervision.
(iii) The different evaluation metrics do measure the same notion of disentanglement and have different biases in their estimation.
(iv) We did not observe that increased disentanglement necessarily implies a decreased sample complexity of learning downstream tasks.
Based on these findings, we suggest three main directions for future research:

\subsection{Inductive Biases and Implicit and Explicit Supervision}
\looseness=-1Our theoretical impossibility result in Section~\ref{sec:impossibility} highlights the need of inductive biases while our experimental results indicate that the role of supervision is crucial. As currently there does not seem to exist a reliable strategy to choose hyperparameters in the unsupervised learning of disentangled representations, we argue that future work should make the role of inductive biases and implicit and explicit supervision more explicit. Recent work~\citep{duan2019heuristic} proposed a stability based heuristic for unsupervised model selection while~\citep{locatello2019disentangling} explored the few-labels regime. Further exploring these techniques may help us understand the practical role of inductive biases and implicit/explicit supervision.

On the other hand, we would encourage and motivate future work on disentangled representation learning that deviates from the static, purely unsupervised setting considered in this work.
Promising settings (that have been explored to some degree) seem to be for example (i) disentanglement learning with interactions \citep{thomas2018disentangling}, (ii) when weak forms of supervision like grouping information are available \citep{bouchacourt2017multi,shu2019weakly,gvae2019,locatello2020weakly}, or (iii) when temporal structure is available for the learning problem~\citep{locatello2020weakly}.
The last setting seems to be particularly interesting given recent identifiability results in non-linear ICA~\citep{hyvarinen2016unsupervised} that enable semi-supervised~\cite{sorrenson2020disentanglement,khemakhem2019variational} and weakly-supervised approaches~\cite{bouchacourt2017multi,gvae2019,shu2019weakly,locatello2020weakly}.

\subsection{Concrete Practical Benefits of Disentangled Representations}
\looseness=-1In our experiments we investigated whether higher disentanglement scores lead to increased sample efficiency for downstream tasks and did not find evidence that this is the case.
Note that these results only apply to the setting and downstream task used in our study. However, recent work~\citep{locatello2019fairness,van2019disentangled} shows compelling evidence supporting the usefulness of some notions of disentangled representations. On some tasks, the structure of the representation may indeed play an important role. A clear example is \citep{van2019disentangled}, where the task involves reasoning about the factors of variation in a sequence of images. Interpretability and fairness~\citep{locatello2019fairness} as well as interactive settings seem to be particularly promising candidates. 
One potential approach to include inductive biases, offer interpretability, and generalization is the concept of independent causal mechanisms and the framework of causal inference \citep{pearl2009causality, peters2017elements}. However, as the different scores considered in this paper measure different notions of disentanglement, it appears to be important to understand which benefits each specific notion may bring.

\subsection{Experimental Setup and Diversity of Data Sets.}
\looseness=-1Our study also highlights the need for a sound, robust, and reproducible experimental setup on a diverse set of data sets in order to draw valid conclusions.
We have observed that it is easy to draw spurious conclusions from experimental results if one only considers a subset of methods, metrics and data sets. Hence,
we argue that it is crucial for future work to perform experiments on a wide variety of data sets to see whether conclusions and insights are generally applicable.
This is particularly important in the setting of disentanglement learning as experiments are largely performed on toy-like data sets. Furthermore, as the considered metrics are measuring different notions of disentanglement, it is important for future work to be explicit about the properties of the learned representation and how these properties are being evaluated.
For this reason, we released \texttt{disentanglement\_lib}, the library we created to train and evaluate the different disentanglement methods and metrics on multiple data sets. 
We also released more than \num{10000} trained models to provide a solid baseline for future research.

\acks{The authors thank Irina Higgins, Ilya Tolstikhin, Paul Rubenstein and Josip Djolonga for helpful discussions and comments. This research was partially supported by the Max Planck ETH
Center for Learning Systems, by an ETH core grant
(to Gunnar R\"atsch) and a Google Ph.D. Fellowship to FL. This work was partially done while
FL was at Google Research Zurich and at the Max Planck Institute for Intelligent Systems. }

\newpage

\appendix
\section{Proof of Theorem~\ref{thm:impossibility}}
\label{sec:proof}
\begin{proof}
To show the claim, we explicitly construct a family of functions $f$ using a sequence of bijective functions.
Let $d > 1$ be the dimensionality of the latent variable $\rvz$ and consider the function $g:\supp(\rvz)\to[0,1]^d$ defined by
\[
g_i(\vv) = \P(\rz_i\leq v_i) \quad \forall i =1, 2, \dots, d.
\]
Since $\P$ admits a density $p(\rvz)=\prod_ip(\rz_i)$, the function $g$ is bijective and, for almost every $\vv\in\supp(\rvz)$, it holds that $\frac{\partial g_i(\vv)}{\partial v_i}\neq 0$ for all $i$ and $\frac{\partial g_i(\vv)}{\partial v_j} = 0$ for all $i\neq j$.
Furthermore, it is easy to see that, by construction, $g(\rvz)$ is a independent $d$-dimensional uniform distribution.
Similarly, consider the function $h:(0,1]^d \to \R^d$ defined by
\[
h_i(\vv) = \psi^{-1}(v_i)\quad \forall i =1, 2, \dots, d,
\]
where $\psi(\cdot)$ denotes the cumulative density function of a standard normal distribution.
Again, by definition, $h$ is bijective with $\frac{\partial h_i(\vv)}{\partial v_i}\neq 0$ for all $i$ and $\frac{\partial h_i(\vv)}{\partial v_j} = 0$ for all $i \neq j$.
Furthermore, the random variable $h(g(\rvz))$ is a $d$-dimensional standard normal distribution.

Let $\mA\in\R^{d\times d}$ be an arbitrary orthogonal matrix with $A_{ij}\neq0$ for all $i=1,2, \dots, d$ and $j=1, 2, \dots, d$. 
An infinite family of such matrices can be constructed using a Householder transformation:
Choose an arbitrary $\alpha\in(0, 0.5)$ and consider the vector $\vv$ with $v_1=\sqrt{\alpha}$ and $v_i=\sqrt{\frac{1-\alpha}{d-1}}$ for $i=2, 3, \dots, d$. 
By construction, we have $\vv^T\vv = 1$ and both $v_i\neq0$ and $v_i\neq\sqrt{\frac12}$ for all $i=1, 2, \dots, d$.
Define the matrix $\mA = \mI_d - 2 \vv\vv^T$ and note that $A_{ii} = 1 - 2v_i^2 \neq 0$ for all $1,2, \dots, d$ as well as $A_{ij} = -v_iv_j\neq0$ for all $i\neq j$.
Furthermore, $\mA$ is orthogonal since 
\begin{align*}
\mA^T\mA = \left(\mI_d - 2 \vv\vv^T\right)^T\left(\mI_d - 2 \vv\vv^T\right) = \mI_d - 4 \vv\vv^T + 4 \vv(\vv^T\vv)\vv^T = \mI_d.
\end{align*}

Since $\mA$ is orthogonal, it is invertible and thus defines a bijective linear operator.
The random variable $\mA h(g(\rvz))\in\R^d$ is hence an independent, multivariate standard normal distribution since the covariance matrix $\mA^T\mA$ is equal to $\mI_d$.

Since $h$ is bijective, it follows that $h^{-1}(\mA h(g(\rvz)))$ is an independent $d$-dimensional uniform distribution.
Define the function $f: \supp(\rvz)\to\supp(\rvz)$
\[
f(\vu) = g^{-1}(h^{-1}(\mA h(g(\vu))))
\]
and note that by definition $f(\rvz)$ has the same marginal distribution as $\rvz$ under $\P$, \ie, $\P(\rvz \leq \vu) = \P(f(\rvz) \leq \vu)$ for all $\vu$.
Finally, for almost every $\vu\in\supp(\rvz)$, it holds that
\begin{equation*}
\frac{\partial f_i(\vu)}{\partial u_j} 
= 
\frac{A_{ij} \cdot\frac{\partial h_j(g(\vu))}{\partial v_j}\cdot \frac{\partial g_j(\vu)}{\partial u_j}}
{\frac{\partial h_i(h_i^{-1}(\mA h(g(\vu)))}{\partial v_i} \cdot \frac{\partial g_i(g^{-1}(h^{-1}(\mA h(g(\vu)))))}{\partial v_i}}
\neq 0,
\end{equation*}
as claimed.
Since the choice of the matrix $\mA$ was arbitrary, there exists an infinite family of such functions $f$.
\end{proof}

 \section{Experimental Conditions and Guiding Principles.}~\label{app:guiding_principles}
In our study, we seek controlled, fair and reproducible experimental conditions. 
We consider the case in which we can sample from a well defined and known ground-truth generative model by first sampling the factors of variations from a distribution $P(\rvz)$ and then sampling an observation from $P(\rvx | \rvz)$. 
Our experimental protocol works as follows:
During training, we only observe the samples of $\rvx$ obtained by marginalizing $P(\rvx | \rvz)$ over $P(\rvz)$. After training, we obtain a representation $r(\rvx)$ by either taking a sample from the probabilistic encoder $Q(\rvz|\rvx)$ or by taking its mean. 
Typically, disentanglement metrics consider the latter as the representation $r(\rvx)$. 
During the evaluation, we assume to have access to the whole generative model: we can draw samples from both $P(\rvz)$ and $P(\rvx | \rvz)$. 
In this way, we can perform interventions on the latent factors as required by certain evaluation metrics. 
We explicitly note that we effectively consider the statistical learning problem where we optimize the loss and the metrics on the known data generating distribution. 
As a result, we do not use separate train and test sets but always take i.i.d. samples from the known ground-truth distribution.
This is justified as the statistical problem is well defined and it allows us to remove the additional complexity of dealing with overfitting and empirical risk minimization.
\section{Limitations of Our Study.}\label{app:limitations}
While we aim to provide a useful and fair experimental study, there are clear limitations to the conclusions that can be drawn from it due to design choices that we have taken.
In all these choices, we have aimed to capture what is considered the state-of-the-art inductive bias in the community.

On the data set side, we only consider images with a heavy focus on synthetic images. 
We do not explore other modalities and we only consider the toy scenario in which we have access to a data generative process with uniformly distributed factors of variations. 
Furthermore, all our data sets have a small number of independent discrete factors of variations without any confounding variables.

For the methods, we only consider the inductive bias of convolutional architectures. 
We do not test fully connected architectures or additional techniques such as skip connections. 
Furthermore, we do not explore different activation functions, reconstruction losses or different number of layers. 
We also do not vary any other hyperparameters other than the regularization weight. 
In particular, we do not evaluate the role of different latent space sizes, optimizers and batch sizes. 
We do not test the sample efficiency of the metrics but simply set the size of the train and test set to large values.

Implementing the different disentanglement methods and metrics has proven to be a difficult endeavour.
Few ``official'' open source implementations are available and there are many small details to consider.
We take a best-effort approach to these implementations and implemented all the methods and metrics from scratch as any sound machine learning practitioner might do based on the original papers.
When taking different implementation choices than the original papers, we explicitly state and motivate them.
\section{Differences with Previous Implementations.}\label{app:differences}
As described above, we use a single choice of architecture, batch size and optimizer for all the methods which might deviate from the settings considered in the original papers.
However, we argue that unification of these choices is the only way to guarantee a fair comparison among the different methods such that valid conclusions may be drawn in between methods.
The largest change is that for DIP-VAE and for $\beta$-TCVAE we used a batch size of 64 instead of 400 and 2048 respectively.
However, \citet{chen2018isolating} shows in Section H.2 of the Appendix that the bias in the mini-batch estimation of the total correlation does not  significantly affect the performances of their model even with small batch sizes.
For DIP-VAE-II, we did not implement the additional regularizer on the third order central moments since no implementation details are provided and since this regularizer is only used on specific data sets.

Our implementations of the disentanglement metrics deviate from the implementations in the original papers as follows:
First, we strictly enforce that all factors of variations are treated as discrete variables as this corresponds to the assumed ground-truth model in all our data sets.
Hence, we used classification instead of regression for the SAP score and the disentanglement score of~\citep{eastwood2018framework}.
This is important as it does not make sense to use regression on true factors of variations that are discrete (for example on shape on dSprites).
Second, wherever possible, we resorted to using the default, well-tested Scikit-learn~\citep{scikitlearn} implementations instead of using custom implementations with potentially hard to set hyperparameters.
Third, for the Mutual Information Gap~\citep{chen2018isolating}, we estimate the \textit{discrete} mutual information (as opposed to continuous) on the \textit{mean} representation (as opposed to sampled) on a \textit{subset} of the samples (as opposed to the whole data set).
We argue that this is the correct choice as the mean is usually taken to be the representation.
Hence, it would be wrong to consider the full Gaussian encoder or samples thereof as that would correspond to a different representation.
Finally, we fix the number of sampled train and test points across all metrics to a large value to ensure robustness. 
\section{Main Experiment Hyperparameters}\label{app:hyperparameters}
In our study, we fix all hyperparameters except one per each model. 
Model specific hyperparameters can be found in Table~\ref{table:sweep_main}. 
The common architecture is depicted in Table~\ref{table:architecture_main} along with the other fixed hyperparameters in Table~\ref{table:fixed_param_main}. 
For the discriminator in FactorVAE we use the architecture in Table~\ref{table:discriminator_main} with hyperparameters in Table~\ref{table:param_discriminator_main}. 
All the hyperparameters for which we report single values were not varied and are selected based on the literature. 
\begin{table*}
\centering
\caption{Encoder and Decoder architecture for the main experiment.}
\vspace{2mm}
\begin{tabular}{l  l}
\toprule
\textbf{Encoder} & \textbf{Decoder}\\
\midrule 
Input: $64\times 64 \times$ number of channels & Input: $\R^{10}$\\
$4\times 4$ conv, 32 ReLU, stride 2 & FC, 256 ReLU\\
$4\times 4$ conv, 32 ReLU, stride 2 & FC, $4\times 4\times 64$ ReLU\\
$4\times 4$ conv, 64 ReLU, stride 2 & $4\times 4$ upconv, 64 ReLU, stride 2\\
$4\times 4$ conv, 64 ReLU, stride 2 & $4\times 4$ upconv, 32 ReLU, stride 2\\
FC 256, F2 $2\times 10$ & $4\times 4$ upconv, 32 ReLU, stride 2\\
& $4\times 4$ upconv, number of channels, stride 2\\
\bottomrule
\end{tabular}
\label{table:architecture_main}
\end{table*}

\begin{table*}
\centering
\caption{Model's hyperparameters. We allow a sweep over a single hyperparameter for each model.}
\vspace{2mm}
\begin{tabular}{l l  l}
\toprule
\textbf{Model} & \textbf{Parameter} & \textbf{Values}\\
\midrule 
$\beta$-VAE & $\beta$ & $[1,\ 2,\ 4,\ 6,\ 8,\ 16]$\\
AnnealedVAE & $c_{max}$ & $[5,\ 10,\ 25,\ 50,\ 75,\ 100]$\\
& iteration threshold & $100000$\\
& $\gamma$ & $1000$\\
FactorVAE & $\gamma$ & $[10,\ 20,\ 30,\ 40,\ 50,\ 100]$\\
DIP-VAE-I & $\lambda_{od}$ & $[1,\ 2,\ 5,\ 10,\ 20,\ 50]$\\
&$\lambda_{d}$ & $10\lambda_{od}$\\
DIP-VAE-II & $\lambda_{od}$ & $[1,\ 2,\ 5,\ 10,\ 20,\ 50]$\\
&$\lambda_{d}$ & $\lambda_{od}$\\
$\beta$-TCVAE & $\beta$ & $[1,\ 2,\ 4,\ 6,\ 8,\ 10]$\\
\bottomrule
\end{tabular}
\label{table:sweep_main}
\end{table*}

\begin{table}[t]\caption{Other fixed hyperparameters.}
\centering
  \begin{subtable}[t]{0.4\linewidth}
    \centering
    \small
    \begin{tabular}{l  l}
      \toprule
      \textbf{Parameter} & \textbf{Values}\\
      \midrule 
      Batch size & $64$\\
      Latent space dimension & $10$\\
      Optimizer & Adam\\
      Adam: beta1 & 0.9\\
      Adam: beta2 & 0.999\\
      Adam: epsilon & 1e-8\\
      Adam: learning rate & 0.0001\\
      Decoder type & Bernoulli\\
      Training steps & 300000\\
      \bottomrule
    \end{tabular}
    \caption{Hyperparameters common to each of the considered methods.}\label{table:fixed_param_main}
  \end{subtable}%
  \hspace{5mm}
  \begin{subtable}[t]{0.4\linewidth}
    \centering
    \begin{tabular}{l  }
      \toprule
      \textbf{Discriminator} \\
      \midrule 
      FC, 1000 leaky ReLU\\
      FC, 1000 leaky ReLU\\
      FC, 1000 leaky ReLU\\
      FC, 1000 leaky ReLU\\
      FC, 1000 leaky ReLU\\
      FC, 1000 leaky ReLU\\
      FC, 2 \\
      \bottomrule
    \end{tabular}
    \caption{Architecture for the discriminator in FactorVAE.}\label{table:discriminator_main}
  \end{subtable}
  \hspace{5mm}
  \begin{subtable}[t]{.4\linewidth}
    \centering
    \begin{tabular}{l  l}
      \toprule
      \textbf{Parameter} & \textbf{Values}\\
      \midrule 
      Batch size & $64$\\
      Optimizer & Adam\\
      Adam: beta1 & 0.5\\
      Adam: beta2 & 0.9\\
      Adam: epsilon & 1e-8\\
      Adam: learning rate & 0.0001\\
      \bottomrule
    \end{tabular}
    \caption{Parameters for the discriminator in FactorVAE.}\label{table:param_discriminator_main}
  \end{subtable}
\end{table}

\section{Data Sets and Preprocessing}\label{app:datasets}
All the data sets contains images with pixels between \num{0} and \num{1}. 
\textbf{Color-dSprites:} Every time we sample a point, we also sample a random scaling for each channel uniformly between \num{0.5} and \num{1}. 
\textbf{Noisy-dSprites:} Every time we sample a point, we fill the background with uniform noise. 
\textbf{Scream-dSprites:} Every time we sample a point, we sample a random $64\times 64$ patch of \textit{The Scream} painting. We then change the color distribution by adding a random uniform number to each channel and divide the result by two. Then, we embed the dSprites shape by inverting the colors of each of its pixels.

\section{Additional Figures}\label{app:additional_figures}
\begin{figure}[h]
\centering\includegraphics[width=\textwidth]{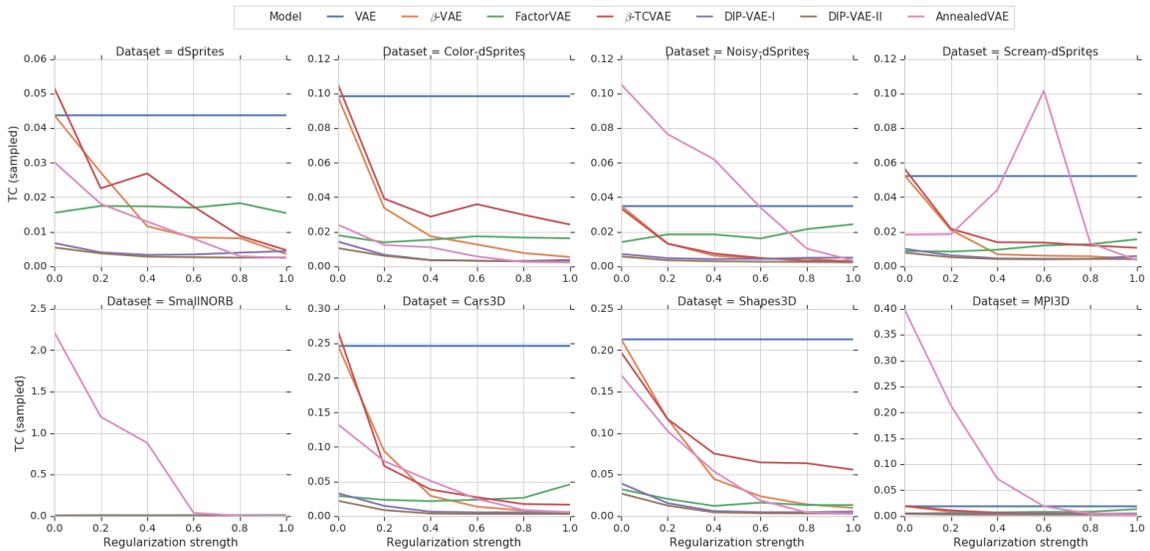}
\caption{Total correlation of sampled representation plotted against regularization strength for different data sets and approaches (including AnnealedVAE).}\label{figure:TCsampledApp}
\vspace{1.5cm}
\end{figure}

\begin{figure}[h!]
\centering\includegraphics[width=\textwidth]{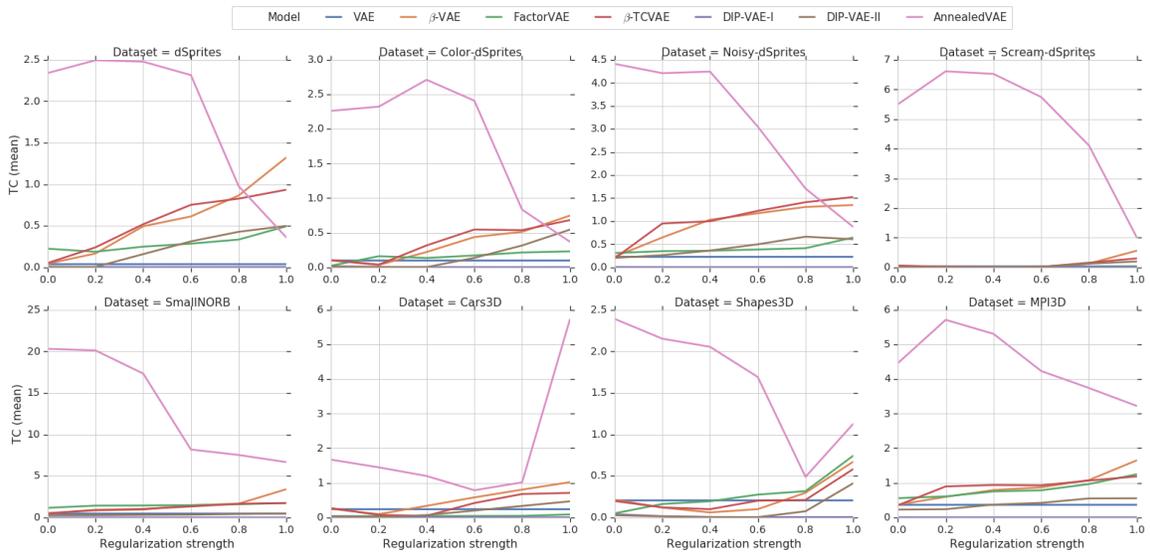}
\caption{Total correlation of mean representation plotted against regularization strength for different data sets and approaches (including AnnealedVAE).}\label{figure:TCmeanApp}
\end{figure}

In this section, we report additional figures complementing the experiments in the main text. In Figures~\ref{figure:TCsampledApp} and~\ref{figure:TCmeanApp}, we report the same plot of Figures~\ref{figure:TCsampled} and~\ref{figure:TCmean} including the AnnealedVAE method.

\begin{figure}[p]
\centering\includegraphics[width=\textwidth]{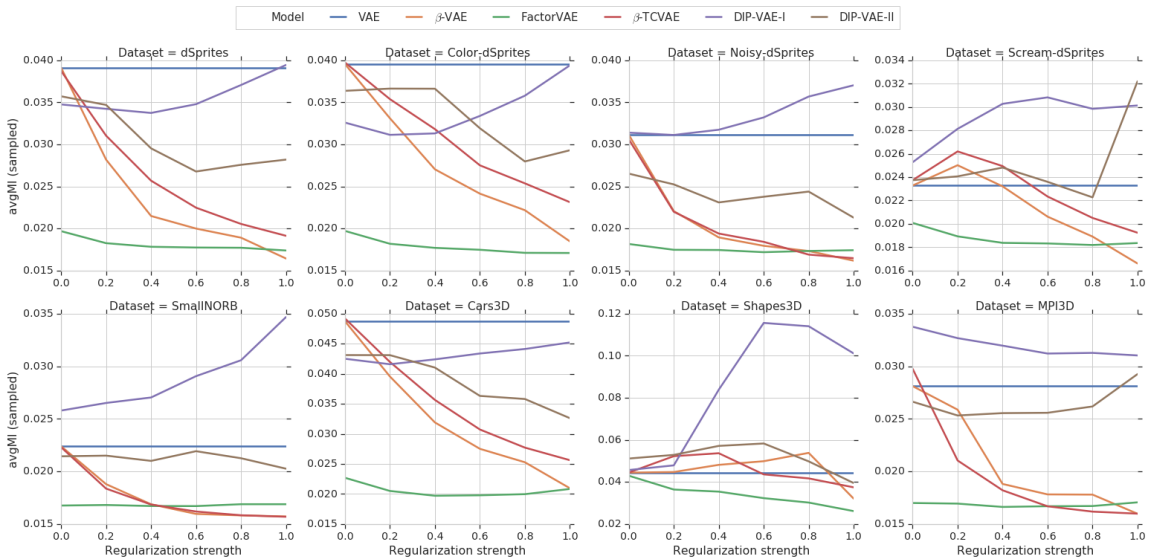}
\caption{The average mutual information of the dimensions of the sampled representation generally decrease except for DIP-VAE-I.  }\label{figure:avgMIsampled}
\end{figure}

\begin{figure}[p]
\centering\includegraphics[width=\textwidth]{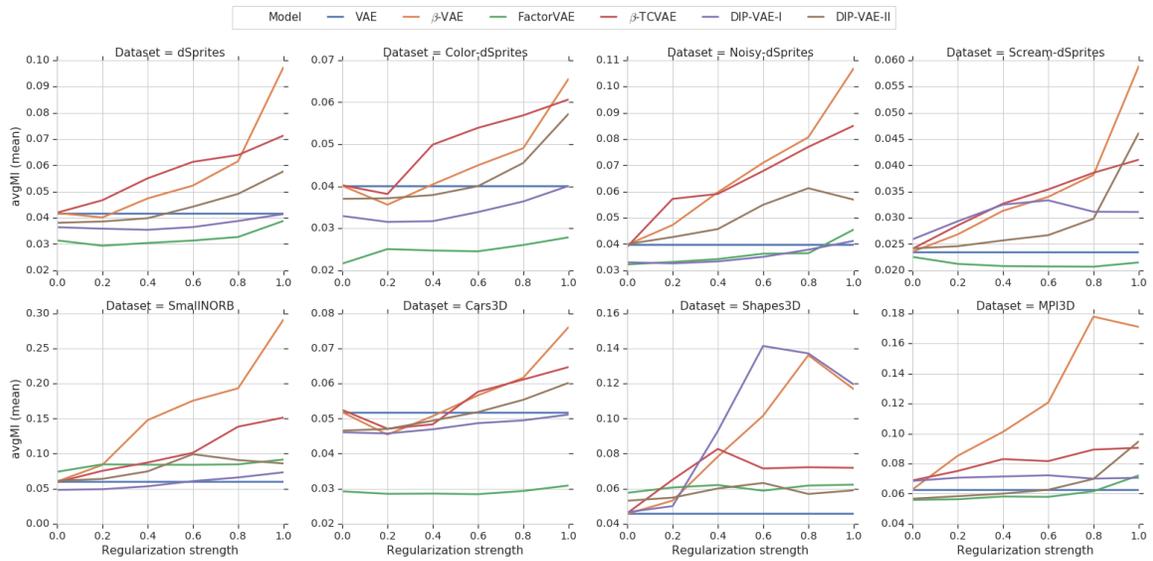}
\caption{The average mutual information of the dimensions of the mean representation generally increase.}\label{figure:avgMImean}
\end{figure}

In Figures~\ref{figure:avgMIsampled} and~\ref{figure:avgMImean} we observed a trend similar to Figures~\ref{figure:TCsampled} and~\ref{figure:TCmean} if we consider the distance from diagonal of the matrix encoding the pairwise mutual information between factors of variation and codes instead of the total correlation.

In Table~\ref{table:variance_explained}, we report the variance per data set explained by the objective only (a) and both objective and hyperparameters (b).

\begin{figure}[p]
\centering\includegraphics[width=0.9\textwidth]{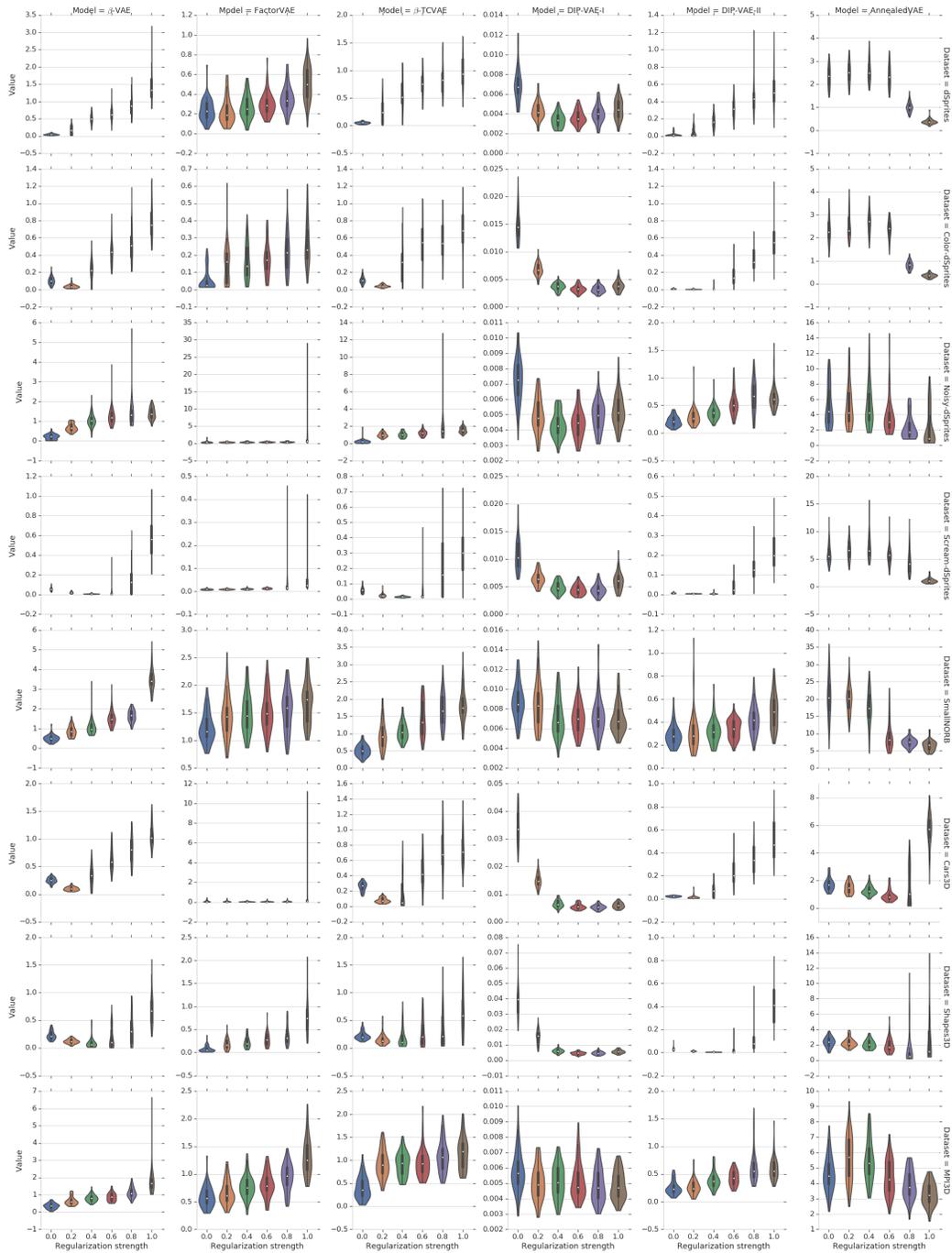}
\caption{The effect of randomness on the total correlation of the mean representation for each method. We observe an overlap between the different hyperparameters settings similar to what we observed in Figure~\ref{figure:random_seed_effect_Cars3D} for the disentanglement metrics on Cars3D. }\label{figure:TC_random_seed_effect_all_datasets}
\end{figure}

In Figure~\ref{figure:TC_random_seed_effect_all_datasets}, we plot the distribution of the total correlation of the mean representation of each method for different regularization strengths on the different data sets. Overall, we note that the different hyperparameters settings produce representations whose total correlation significantly overlaps. This trend is comparable to what we observed in Figure~\ref{figure:random_seed_effect_Cars3D} for the disentanglement scores on Cars3D.


\begin{table}
\vspace{2mm}
\begin{subtable}{\textwidth}
\centering
\begin{tabular}{lrrrrrrr}
\toprule
{} &  A &  B &  C &  D &  E &  F &  G \\
\midrule
Cars3D          &             1\% &                  38\% &              26\% &  78\% &  34\% &         35\% &   8\% \\
Color-dSprites  &            30\% &                  39\% &              50\% &  75\% &  25\% &         24\% &  28\% \\
MPI3D           &            61\% &                  59\% &              50\% &  78\% &  45\% &         45\% &  21\% \\
Noisy-dSprites  &            17\% &                  21\% &              18\% &  78\% &  10\% &         43\% &   9\% \\
Scream-dSprites &            90\% &                  50\% &              78\% &  54\% &  45\% &         61\% &  55\% \\
Shapes3D        &            33\% &                  21\% &              14\% &  43\% &  21\% &         27\% &  10\% \\
SmallNORB       &            68\% &                  73\% &              60\% &  87\% &  72\% &         62\% &  57\% \\
dSprites        &            31\% &                  43\% &              49\% &  71\% &  27\% &         30\% &  33\% \\
\bottomrule
\end{tabular}
\caption{Percentage of variance explained regressing the disentanglement scores on the different data sets from the objective function only.}
\end{subtable}
\begin{subtable}{\textwidth}
\centering
\vspace{2mm}
\begin{tabular}{lrrrrrrr}
\toprule
{} & A & B & C &  D &  E &  F &  G \\
\midrule
Cars3D          &             5\% &                  67\% &              41\% &  97\% &  60\% &         49\% &  13\% \\
Color-dSprites  &            68\% &                  80\% &              59\% &  92\% &  75\% &         40\% &  56\% \\
MPI3D           &            91\% &                  81\% &              80\% &  94\% &  56\% &         62\% &  44\% \\
Noisy-dSprites  &            27\% &                  42\% &              25\% &  87\% &  29\% &         53\% &  22\% \\
Scream-dSprites &            93\% &                  74\% &              84\% &  83\% &  66\% &         68\% &  75\% \\
Shapes3D        &            61\% &                  79\% &              53\% &  82\% &  57\% &         48\% &  33\% \\
SmallNORB       &            87\% &                  90\% &              82\% &  95\% &  89\% &         73\% &  78\% \\
dSprites        &            64\% &                  77\% &              55\% &  90\% &  71\% &         38\% &  57\% \\
\bottomrule
\end{tabular}
\caption{Percentage of variance explained regressing the disentanglement scores on the different data sets from the Cartesian product of objective function and regularization strength.}
\end{subtable}
\caption{Variance of the disentanglement scores explained by the objective function or its cartesian product with the hyperparameters. The variance explained is computed regressing using ordinary least squares.  Legend: A = BetaVAE Score,  B = DCI Disentanglement, C = FactorVAE Score,  D = IRS,  E = MIG, F = Modularity, G = SAP.}\label{table:variance_explained}
\end{table}

\begin{figure}
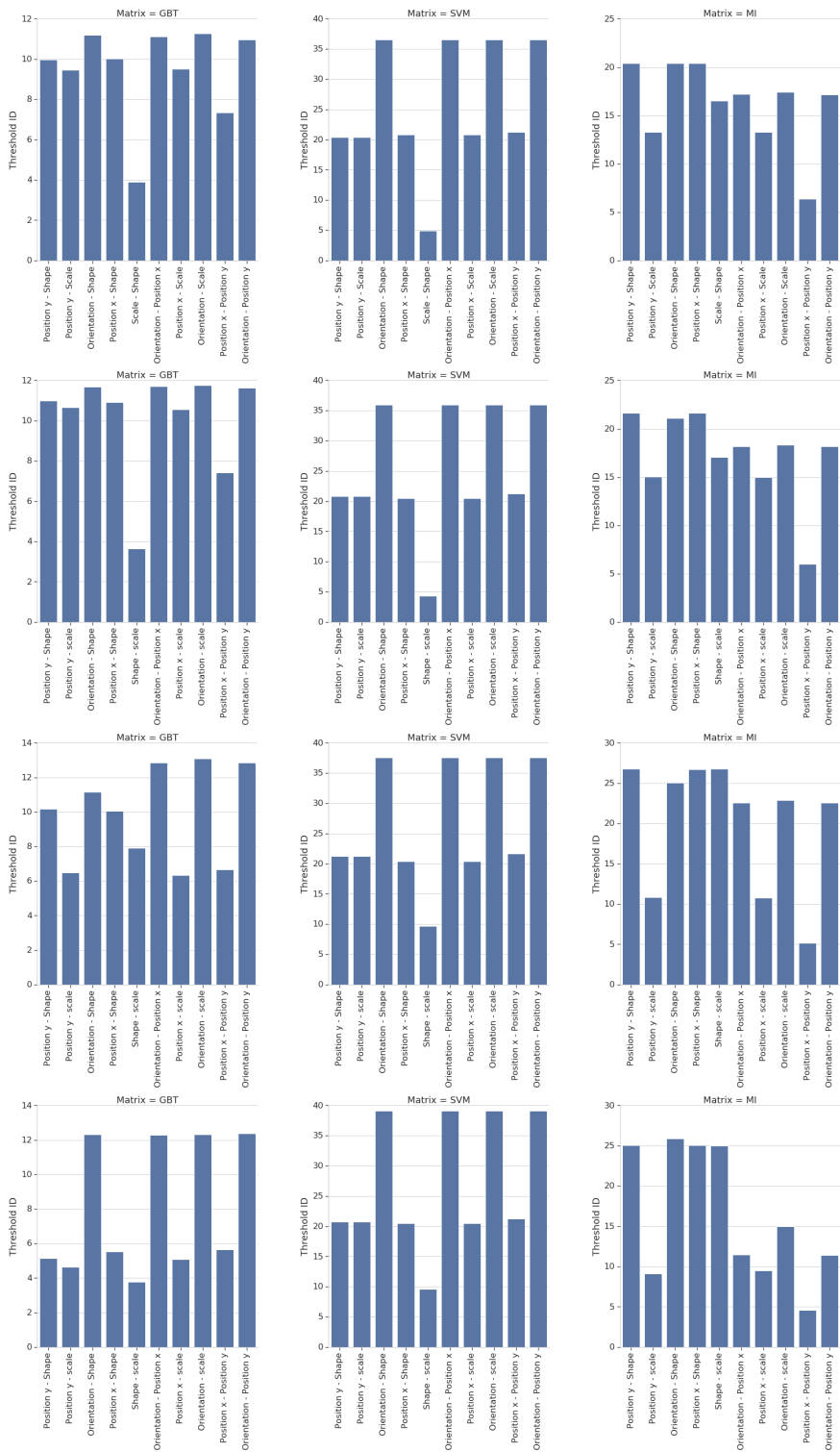

\centering\includegraphics[width=0.76\textwidth]{autofigures/confused_factors_dSprites}
\centering\includegraphics[width=0.76\textwidth]{autofigures/confused_factors_Color-dSprites}
\centering\includegraphics[width=0.76\textwidth]{autofigures/confused_factors_Noisy-dSprites}
\centering\includegraphics[width=0.76\textwidth]{autofigures/confused_factors_Scream-dSprites}
\caption{Threshold ID of confused factors for dSprites, Color-dSprites, Noisy-dSprites and Scream-dSprites. Lower threshold means that the two factors are found more entangled.
}\label{figure:confused_factors_dSprites}
\end{figure}
\begin{figure}
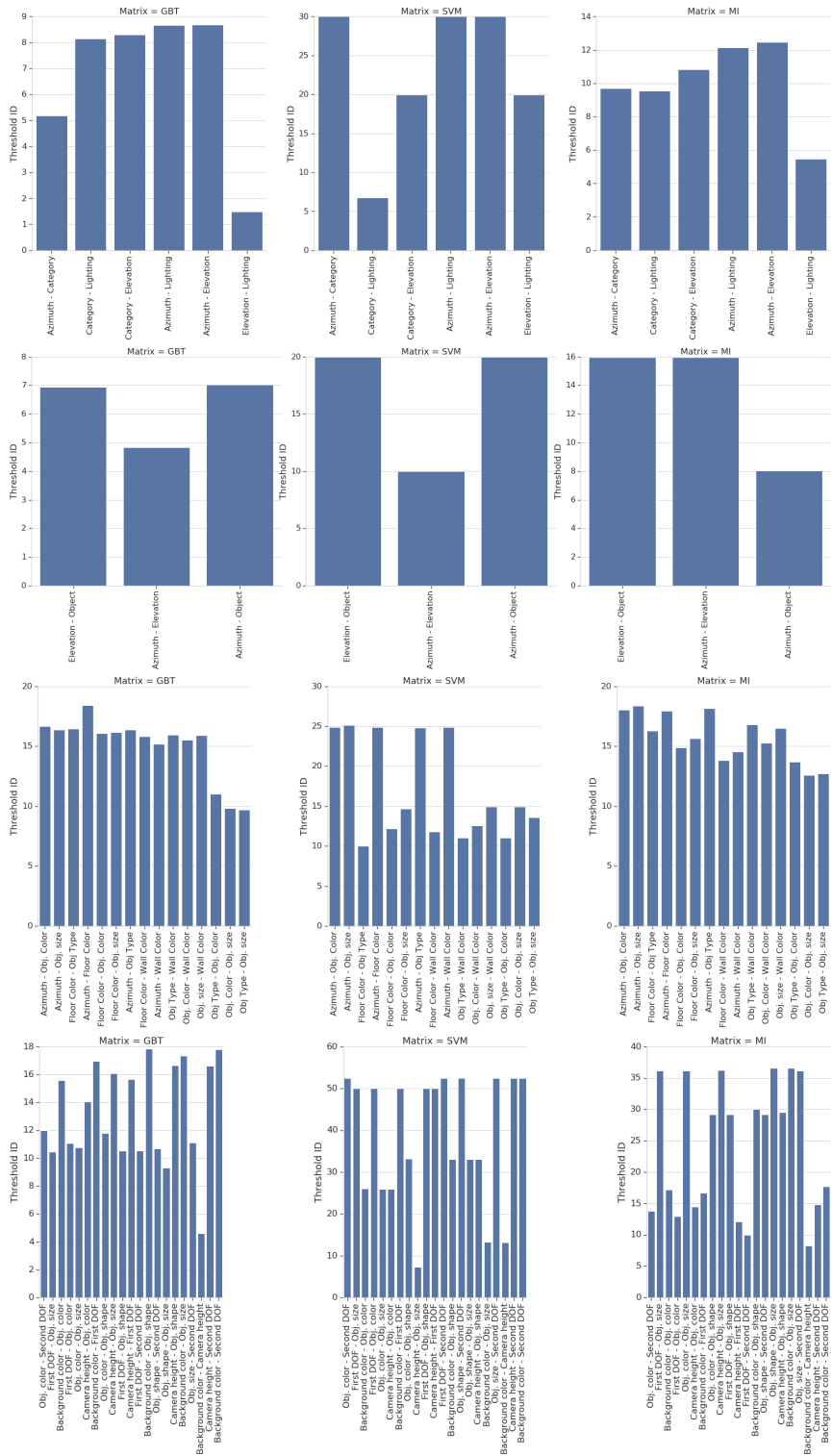

\centering\includegraphics[width=0.75\textwidth]{autofigures/confused_factors_SmallNORB}
\centering\includegraphics[width=0.75\textwidth]{autofigures/confused_factors_Cars3D}
\centering\includegraphics[width=0.75\textwidth]{autofigures/confused_factors_Shapes3D}
\centering\includegraphics[width=0.75\textwidth]{autofigures/confused_factors_MPI3D}
\caption{Threshold ID of confused factors for SmallNORB, Cars3D, Shapes3D and MPI3D. Lower threshold means that the two factors are found more entangled.
}\label{figure:confused_factors_other}
\end{figure}

\newpage
\clearpage
\vskip 0.2in
\bibliography{main}

\end{document}